\documentclass[runningheads]{llncs}
\usepackage[utf8]{inputenc}
\usepackage{accv}
\usepackage{accvabbrv}
\usepackage[accsupp]{axessibility}
\usepackage{graphicx}
\usepackage{float}
\usepackage{booktabs}
\usepackage{tabularx}
\usepackage{array}
\usepackage{siunitx}
\usepackage{multirow}
\usepackage{threeparttable}
\usepackage{xurl}
\usepackage{hyperref}
\usepackage{orcidlink}
\hypersetup{hidelinks,pdftitle={Image-Space Rule Discovery},pdfauthor={Misora Sugiyama, Toya Oyama, and Hirokatsu Kataoka},pdfkeywords={image editing, visual problem solving, worksheet benchmark, WISP}}
\definecolor{taskMarking}{HTML}{9BB9DF}
\definecolor{taskFilling}{HTML}{FBA659}
\definecolor{taskCopying}{HTML}{D0B7DF}
\definecolor{taskCounting}{HTML}{B2D39D}
\definecolor{taskSuppression}{HTML}{D3D3D3}
\DeclareRobustCommand{\Marking}{\textbf{Marking}}
\DeclareRobustCommand{\Filling}{\textbf{Filling}}
\DeclareRobustCommand{\Copying}{\textbf{Copying}}
\DeclareRobustCommand{\Counting}{\textbf{Counting}}
\DeclareRobustCommand{\Suppression}{\textbf{Suppression}}
\graphicspath{{figures/}}
\newcolumntype{Y}{>{\centering\arraybackslash}X}
\begin{document}
\title{Image-Space Rule Discovery}
\titlerunning{Image-Space Rule Discovery}
\author{Misora Sugiyama\inst{1,2}\orcidlink{0009-0006-1245-8619} \and Toya Oyama\inst{3,2}\orcidlink{0009-0007-8108-6529} \and Hirokatsu Kataoka\inst{4,5,2}\orcidlink{0000-0001-8844-165X}}
\authorrunning{M.~Sugiyama et al.}
\institute{New York University \and LIMIT.Lab \and The University of Tokyo \and National Institute of Advanced Industrial Science and Technology (AIST) \and Visual Geometry Group, University of Oxford}
\maketitle
\setcounter{footnote}{0}

\begin{abstract}
Can image editors solve worksheets while preserving unrelated content? WISP (Worksheet Image-Space Problem Solving) defines 11 tasks and eight information conditions. On 2,200 shared items per editor, Nano Banana Pro achieves a 48.7\% strict automatic proxy pass rate (Auto-Strict), versus 0.0--13.4\% for four other editors. Analysis of 13,200 outputs, including a separate diagnostic, distinguishes target, preservation, and format failures. Auto-Strict agrees with human-majority judgments on 95.5\% of 440 outputs (inter-rater Fleiss' $\kappa=0.894$). Excluding the two no-edit controls, Nano Banana Pro achieves 45.1\% Auto-Strict, versus at most 3.2\% for the other editors. Further probes show high symbol-tracking proxy scores but only 49\% exact Wason selection. These results measure problem solving through edits, without establishing open-ended rule induction or transfer.\footnote{Data and code: \url{https://github.com/misora-sugiyama/WISP}.}
\end{abstract}

\begin{figure}[H]
\centering
\includegraphics[width=\linewidth]{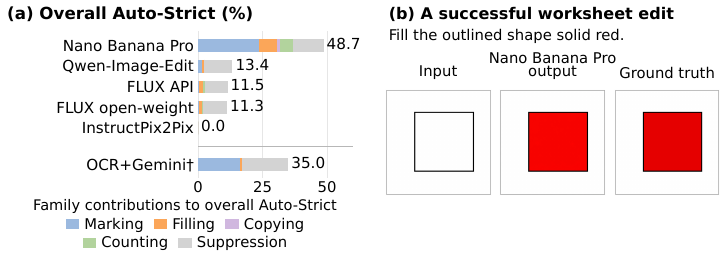}
\caption{Overall Auto-Strict on 2,200 shared items (left; colors show each task family's contribution; $\dagger$: separate OCR+Gemini diagnostic) and a successful Target-Shape Filling edit (right; V0; input/output/GT crops).}
\label{fig:teaser}
\end{figure}

\section{Introduction}
\label{sec:intro}
Recent vision and vision-language systems achieve strong image recognition, visual question answering, and multimodal instruction following~\cite{radford2021clip,alayrac2022flamingo,li2023blip2,liu2024llava}, usually returning text or symbolic labels. Image-editing models pose a complementary question: can a system solve a visual problem by placing its answer at the intended destination in the same image while preserving unrelated content?

\begin{figure}[!t]
\centering
\includegraphics[width=\linewidth]{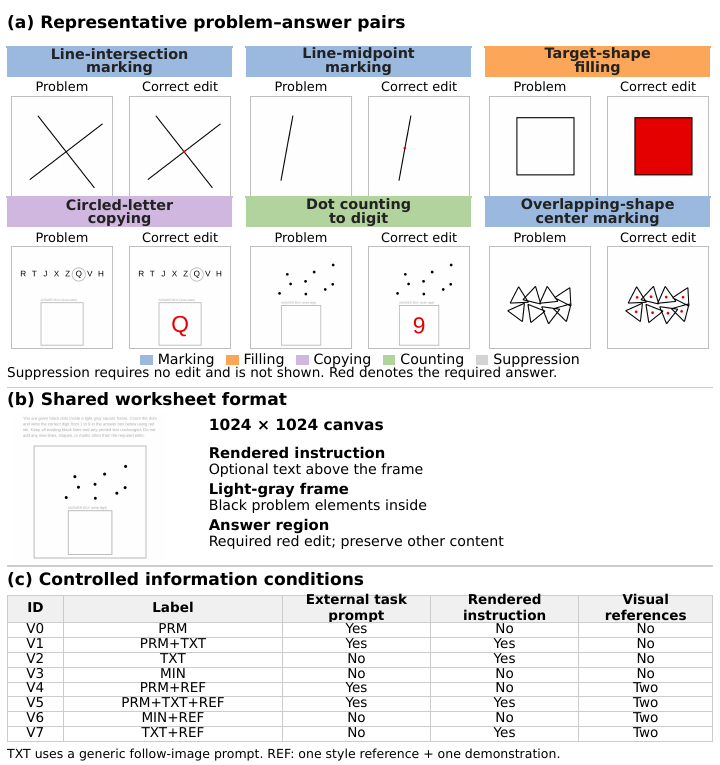}
\caption{WISP task and protocol overview. (a) Representative problem--answer pairs from \Marking{}, \Filling{}, \Copying{}, and \Counting{}. Suppression controls require no edit and are not shown. Red denotes the required output edit. (b) The shared $1024\times1024$ worksheet format. (c) Eight conditions controlling external prompts, rendered instructions, and visual references.}
\label{fig:benchmark_overview}
\end{figure}

WISP (Worksheet Image-Space Problem Solving) evaluates this interface using a fixed worksheet canvas, controlled task information, and task-specific scoring targets (Figure~\ref{fig:benchmark_overview}). Marking a line intersection, for example, requires identifying the operation, localizing its target, and producing the requested number of marks while preserving the geometry. A plausible red dot is insufficient if the model redraws lines, alters the printed instruction, or adds extra answers; doing nothing preserves the input but fails an active editing task.

We evaluate the complete solve-and-edit process. External or rendered instructions explicitly specify many operations, making these tests of instruction-conditioned problem solving rather than discovery of an unknown rule. A minimal task cue alone (MIN), or with visual references (MIN+REF), leaves more of the operation to be inferred; neither setting establishes generalization to unseen rule families, layouts, or symbols. We interpret outputs as observable behavior without attributing computation to particular components of a proprietary model.

We compare five end-to-end editors on 11 controlled worksheet tasks and report OCR+Gemini plan-render as a separate diagnostic. We break down the recorded evaluation decisions and validate the automatic metrics with three raters to distinguish target success from preservation and format failures. Further diagnostics examine instruction access and sensitivity and probe more demanding reasoning tasks. Together, these analyses show where solving a worksheet and executing a faithful edit come apart. Figure~\ref{fig:teaser} summarizes the overall comparison and illustrates a successful worksheet edit.

\section{Related Work}
\label{sec:related}
\noindent\textbf{Instruction-guided image editing.} Diffusion and latent-diffusion models support high-quality image synthesis~\cite{sohldickstein2015nonequilibrium,ho2020ddpm,song2021score,dhariwal2021diffusion,rombach2022ldm}. SDEdit, Prompt-to-Prompt, and Null-text inversion broaden controlled image editing~\cite{meng2022sdedit,hertz2022prompttoprompt,mokady2023nulltext}, while InstructPix2Pix and MagicBrush study natural-language-guided edits~\cite{brooks2023instructpix2pix,zhang2023magicbrush}. Editing and generation evaluations commonly emphasize instruction alignment, perceptual quality, compositionality, or preference~\cite{hu2023tifa,huang2025t2icompbenchplusplus,ghosh2023geneval}. WISP instead specifies an operation at a constrained destination and evaluates whether the system performs it while preserving the worksheet.

\noindent\textbf{Diagnostic visual reasoning.} CLEVR uses synthetic scenes and functional programs~\cite{johnson2017clevr}. MMMU tests knowledge across diverse tasks~\cite{yue2024mmmu}. RAVEN evaluates relational and analogical reasoning~\cite{zhang2019raven}. Fine-grained VLM diagnostics reveal failures on intersections, overlapping shapes, circled letters, grids, and counting~\cite{Rahmanzadehgervi_2024_ACCV,singh2024vlmsreallyblind,vo2025vlmsbiased}. WISP studies related primitives through visual outputs on the original canvas, exposing additional errors in destination binding, edit count, unnecessary changes, and native format. 

\noindent\textbf{Generative vision and pixel-space interfaces.} Veo 3 has been studied as a zero-shot visual learner that receives an input image and an external text instruction~\cite{wiedemer2025video}. Vision Banana instruction-tunes Nano Banana Pro to produce 2D and 3D vision-task outputs as RGB images~\cite{gabeur2026image}. FLUX-family editors adapted with task-specific LoRA and lightweight conditioning can approximate visually encoded numerical and physical mappings through supervised input--solution learning~\cite{mizrahi2026numericalsolvers}. WISP instead evaluates first-shot localized worksheet edits without task-specific training for this benchmark, with preservation requirements and conditions in which task instructions appear only inside the image. ProVisE elicits protocol-constrained pixel answers, parses them into structured predictions, and evaluates them with benchmark metrics; its SpatialGen-Bench evaluates spatial cognition~\cite{wang2026showdonttell}. That work emphasizes metric-compatible adaptation across spatial benchmarks and text-versus-visual answer interfaces. WISP shares the visual-answer motivation while controlling task-information sources and evaluating destination binding, output-count control, unnecessary-edit suppression, and preservation on the original worksheet. Low-level editing failures remain relevant even when outputs look plausible~\cite{zuo2025nanobanana}.

\noindent\textbf{Reasoning-based editing benchmarks.} RISEBench studies temporal, causal, spatial, and logical editing~\cite{zhao2025rise}; KRIS-Bench targets factual, conceptual, and procedural knowledge~\cite{wu2025kris}; UniREditBench covers multi-object real-world and game-world reasoning~\cite{han2025unireditbench}; and GIR-Bench examines understanding--generation consistency and multi-step reasoning~\cite{li2025girbench}. WISP complements these benchmarks with controlled worksheet operations and separate evaluation of target success, preservation, and format requirements.

\section{Worksheet Image-Space Problem Solving}
\label{sec:image-space-rule-discovery}
WISP asks whether a model can obtain the requested answer from the available worksheet and task information, then draw the answer on the same canvas while respecting the editing constraints. We provide frozen inputs, ground-truth edited outputs, task parameters, and scoring metadata. The evaluator derives target masks from the ground-truth images and preservation masks from the inputs. Success requires both the requested edit and preservation of the surrounding content.

\noindent\textbf{Task design and behavioral requirements.} The fixed $1024\times1024$ layout places optional instructions above a light-gray frame and black problem elements and answer regions inside it. Fixing this format reduces visual-style variation when analyzing operational failures. Eleven templates span five families. \Marking{} contains Line-Intersection Marking, Line-Midpoint Marking, Dot-Circling Marking, and Overlapping-Shape Center Marking; \Filling{} contains Target-Shape Filling; \Copying{} covers Circled-Letter Copying, Identical-Letter Copying (All-A Control), and Single-Letter Copying; \Counting{} contains Dot Counting to Digit; and \Suppression{} contains Blank-Answer Suppression and Empty-Circle Suppression. These families probe localized placement, region coverage, answer transfer, counting-to-symbol output, and the ability to withhold an edit. They motivate eight failure-analysis dimensions: instruction access, problem recognition, answer inference, output-count control, destination binding, input preservation, unnecessary-edit suppression, and format preservation. These are descriptive dimensions, not independently calibrated ability scores.

For active tasks, the desired output is the original worksheet plus the required red answer; suppression tasks require withholding that answer. Changes to the printed instruction, black geometry, answer boxes, or frame violate preservation. Family colors are used only for presentation, with red reserved for the answer channel in the task convention.

\noindent\textbf{Controlled information conditions.} The four no-reference conditions are V0 (PRM), V1 (PRM+TXT), V2 (TXT), and V3 (MIN). The reference conditions are V4 (PRM+REF), V5 (PRM+TXT+REF), V6 (MIN+REF), and V7 (TXT+REF). Each reference-conditioned item contains exactly two references: one style reference and one input--output demonstration. PRM and PRM+TXT, including their REF variants, retain task-specific external prompts. TXT and TXT+REF direct the model to follow the instructions written in the image. MIN and MIN+REF use the generic prompt ``Edit the image.'' Nano Banana Pro also receives fixed instructions identifying the input image and the required output format. In the Nano Banana Pro and FLUX API reference experiments, the worksheet and two references are supplied as separate image inputs. We group PRM/TXT-based conditions and their REF variants as instruction-conditioned and MIN/MIN+REF as more induction-oriented. Success under a task-specific prompt does not demonstrate use of image text; TXT more directly tests access to a rendered task specification.

\noindent\textbf{Construction and reproducibility.} The public release contains the fixed shared dataset, ground-truth images, metadata, and all six sets of frozen shared outputs. Masks are derived during evaluation rather than distributed as separate files in the fixed dataset. The repository also provides the evaluator and a development renderer. Newly rendered instances are distinct from the frozen paper data.

\section{Evaluation Protocol}
\label{sec:evaluation}
\noindent\textbf{Models and scale.} We compare Nano Banana Pro, Qwen-Image-Edit, FLUX.2 Klein 4B API and open-weight implementations, and InstructPix2Pix. This set covers proprietary API systems, open-weight editors, and an established instruction-editing baseline; it does not support a controlled attribution of differences to scale or architecture. Each editor is evaluated on 11 tasks, 50 base problems per task, and V0--V3, yielding 2,200 outputs per editor. A larger FLUX.2 Klein 4B API experiment contains $11\times400\times8=35{,}200$ outputs and includes reference conditions~\cite{blackforestlabs2026flux2klein}.

In experiments run between June and July 2026, we access Nano Banana Pro through the Gemini API as \nolinkurl{gemini-3-pro-image-preview}~\cite{google2026gemini3} and use \nolinkurl{Qwen/Qwen-Image-Edit} for Qwen~\cite{wu2025qwenimage}. Open-weight FLUX runs through \texttt{Flux2\discretionary{}{}{}Klein\discretionary{}{}{}Pipeline} with checkpoint \nolinkurl{black-forest-labs/FLUX.2-klein-4B}. For InstructPix2Pix, we use the \texttt{StableDiffusion\discretionary{}{}{}InstructPix2Pix\discretionary{}{}{}Pipeline} implementation and checkpoint \nolinkurl{timbrooks/instruct-pix2pix}.

\noindent\textbf{Structured-plan diagnostic.} We evaluate OCR+Gemini plan-render on the same 2,200-item subset, using \nolinkurl{gemini-3.1-pro-preview}. It first performs OCR and extracts visual candidates with OpenCV. It then predicts an edit plan using candidate IDs. A task-specific deterministic Python renderer draws the answer. Plans are checked using input-derived validity rules and, when flagged, revised by Gemini at most once. Deterministic automatic repair is disabled. This diagnostic controls rendering but is not an end-to-end editor; its validation-and-revision procedure is distinct from the first-shot policy of the five editors.

\noindent\textbf{Automatic operational proxies.} For the final shared comparison, Auto-Strict requires task-specific target coverage or localization, component or hit counts where applicable, preservation of black geometry, text, and frames, and native canvas-format validity. Auto-Loose relaxes target thresholds and does not require native size. It omits preservation checks for spatial marking/filling tasks and retains weaker structure checks for symbol and suppression tasks. Filling uses red-mask overlap; point/ring tasks use components, centroids, or local-hit matching; letter/digit tasks use red content near the ground-truth glyph, red-mask overlap with the ground-truth answer, limits on extra red content, structure preservation, and format checks. Suppression uses red-pixel limits and structure-preservation checks. The automatic evaluator does not explicitly recognize glyph identity, so these are proxy pass rates, not human semantic accuracy. Red Dice~\cite{dice1945measures}, black F1, and structure F1 are task-dependent scoring components and diagnostics. Black F1 contributes to Auto-Strict for filling and point/ring tasks, while structure F1 contributes to the letter/digit and no-edit proxy rules. Format is checked before alignment or resizing. 

\noindent\textbf{Decomposing the frozen decisions.} We decompose all 13,200 shared outputs from the five editors and the separate structured-plan diagnostic into three nested criteria: the target criterion alone; the target criterion with preservation; and the target criterion with preservation and native format. The final stage reproduces every reported Auto-Strict decision with zero reconstruction mismatches. This changes how the recorded decisions are analyzed, not the outputs or their final pass/fail labels. We call the first stage a target criterion rather than reasoning accuracy because it can still depend on localization and rendering. Table~\ref{tab:target-final} reports all three stages.

\begin{table}[!t]
\caption{Cumulative proxy pass rates on frozen shared outputs (2,200 per system; 13,200 total). Preservation and native-format requirements are added to the target criterion in sequence; the final column is Auto-Strict. All rates are percentages. The InstructPix2Pix middle stage contains one pass. OCR+Gemini is a separate diagnostic.}
\label{tab:target-final}
\centering\small
\begin{tabular}{lrrr}
\toprule
System & Target & + Preserv. & + Format \\
\midrule
Nano Banana Pro & 53.0 & 48.7 & 48.7 \\
Qwen-Image-Edit & 19.0 & 13.4 & 13.4 \\
FLUX.2 Klein 4B API & 17.3 & 11.5 & 11.5 \\
FLUX.2 Klein 4B open-weight & 16.8 & 11.3 & 11.3 \\
InstructPix2Pix & 14.6 & $<0.1$ & 0.0 \\
\midrule
OCR+Gemini plan-render & 42.5 & 35.0 & 35.0 \\
\bottomrule
\end{tabular}
\end{table}

\noindent\textbf{First-shot policy and condition oracle.} For the end-to-end editors, we retry only after transport failure, an empty response, or file-decoding failure, never after an incorrect semantic answer. Size or aspect-ratio mismatch causes an Auto-Strict format failure. Auto-Loose may align such outputs for diagnostic scoring. The eight-condition oracle groups each base problem across V0--V7 and reports whether any condition passes the relevant metric. It is an upper envelope over tested conditions, not repeated sampling or a deployable selection rule. It is distinct from O1/O2, which supply additional task or target information as inputs.

\noindent\textbf{Human task-solvability check.} Four unpaid adult volunteers, one author and three non-authors, each answered 110 items using clicks or typed responses rather than pixel-level image editing. Two had a computer-science background and two did not. Before participating, volunteers were informed that their responses would be used for this research. No identifying information was collected. Across 440 responses, Solve-Strict and Solve-Loose were 96.1\% and 99.8\%. This is a solvability check under a different response interface, not a human image-editing baseline.

\noindent\textbf{Human validation and failure audit.} Three raters independently evaluated an identical balanced subset of 440 model outputs: 11 tasks $\times$ four no-reference conditions $\times$ two items $\times$ five models. All had a computer-science background; one was an author. None received compensation. The rating materials omitted model identities, condition labels, automatic scores, and other raters' judgments. Raters were instructed to work independently. A fixed codebook defined Strict/Loose judgments, failure categories, precedence for multiple errors, and handling of ambiguous cases. The protocol and codebook are available in the repository's human-evaluation materials. Inter-rater agreement was calculated before resolving disagreements, and automatic decisions were compared with the three-rater majority. These 440 model outputs are separate from the 440 human solve-only responses. The earlier single-rater audit of 2,200 FLUX API outputs characterizes failure forms and is not used to establish the main ranking.

\section{Results}
\label{sec:main-results}
\begin{figure}[!t]
\centering
\includegraphics[width=\linewidth]{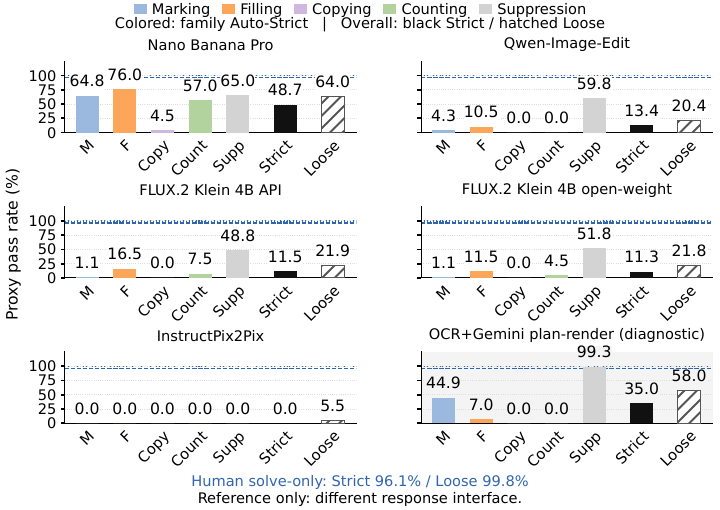}
\caption{Shared V0--V3 proxy pass rates (11 tasks; 2,200 outputs per system). Colored bars show family Auto-Strict; black/hatched bars show overall Auto-Strict/Loose. OCR+Gemini is diagnostic. Blue dashed/dotted lines show human Strict/Loose solve-only rates, using a different interface.}
\label{fig:model-comparison-alltask}
\end{figure}

\noindent\textbf{Shared model comparison.} Figure~\ref{fig:model-comparison-alltask} summarizes V0--V3. Nano Banana Pro leads the evaluated editors with 48.7\% Auto-Strict and 64.0\% Auto-Loose. Qwen-Image-Edit reaches 13.4/20.4\%; FLUX API and open-weight reach 11.5\% and 11.3\% Auto-Strict; InstructPix2Pix reaches 0.0/5.5\%. Overall rates include two no-edit suppression tasks. On the nine active tasks (1,800 outputs per editor), Auto-Strict is 811/1,800 (45.1\%) for Nano Banana Pro, 55/1,800 (3.1\%) for Qwen, 57/1,800 (3.2\%) for FLUX API, 41/1,800 (2.3\%) for FLUX open-weight, and 0/1,800 for InstructPix2Pix. Suppression therefore accounts for most of the Qwen and FLUX passes in the overall comparison.

\noindent\textbf{Target success and faithful execution.} Table~\ref{tab:target-final} separates preservation and format losses. Nano Banana Pro falls from 53.0\% at the target stage to 48.7\% after preservation, with no further loss at the format stage. InstructPix2Pix passes the target criterion on 322 outputs, but only one also passes preservation; that output fails native format. All 2,200 InstructPix2Pix outputs are $512\times512$, rather than the required $1024\times1024$. Its zero final rate thus reflects both preservation failures and a systematic format mismatch.

\noindent\textbf{Rendered instructions.} On the shared TXT subset (11 tasks, 50 problems per task; 550 outputs per system), Nano Banana Pro reaches 63.3/79.5\% Auto-Strict/Auto-Loose and Qwen 15.1/17.6\%. FLUX API reaches 17.8/18.2\%, entirely from the two no-edit suppression tasks. On the five spatial marking/filling tasks---Line-Intersection Marking, Line-Midpoint Marking, Dot-Circling Marking, Overlapping-Shape Center Marking, and Target-Shape Filling---Nano Banana Pro reaches 88.4/91.2\%, while both FLUX implementations and InstructPix2Pix reach 0.0/0.0\%. In the larger FLUX sweep, TXT passes likewise occur only on the two suppression tasks; the active editing tasks remain at zero.

\noindent\textbf{Reliability of the automatic evaluation.} Table~\ref{tab:human-validation} reports the independent three-rater validation. Pairwise Strict agreement is 96.1--97.5\%, with Cohen's $\kappa=0.863$--$0.912$. Against the three-rater majority, Strict precision and recall are 90.1\% and 83.1\%, respectively. Strict agreement is lowest for Counting (87.5\%, $n=40$); Loose agreement is lowest for Suppression (76.3\%, $n=80$). The supplementary material reports all 11 tasks separately and illustrates correct decisions, false positives, and false negatives using frozen outputs. 

\noindent\textbf{Copying and proxy sensitivity.} Nano Banana Pro reaches only 4.5\% Auto-Strict on Copying (27/600). In its 24 copying outputs included in human validation, the majority accepts nine while Auto-Strict accepts three. All six false negatives exceed the limit on red pixels outside the padded ground-truth glyph region. This demonstrates sensitivity to spatial tolerance, but the small validation subset cannot explain every failure across the 600 outputs.

\begin{table}[!t]
\caption{Independent three-rater validation on 440 balanced model outputs. Auto--human metrics use the three-rater majority. Values are percentages except Fleiss' $\kappa$.}
\label{tab:human-validation}
\centering
\begin{tabular}{lrr}
\toprule
Metric & Strict & Loose \\
\midrule
Human unanimous agreement & 95.5 & 88.9 \\
Fleiss' $\kappa$ & 0.894 & 0.783 \\
Auto--human agreement & 95.5 & 89.3 \\
Precision & 90.1 & 69.0 \\
Recall & 83.1 & 87.9 \\
F1 & 86.5 & 77.3 \\
Auto--human range (models) & 86.4--100 & 85.2--93.2 \\
Auto--human range (families) & 87.5--100 & 76.3--95.6 \\
\bottomrule
\end{tabular}
\end{table}

\noindent\textbf{Structured-plan diagnostic.} OCR+Gemini plan-render reaches 35.0\% Auto-Strict and 58.0\% Auto-Loose, with 100.0\% native-format validity. Its target-criterion rate is 42.5\% (Table~\ref{tab:target-final}). Controlled drawing therefore does not by itself solve task interpretation, candidate selection, or preservation. 

\begin{figure}[!t]
\centering
\includegraphics[width=\linewidth]{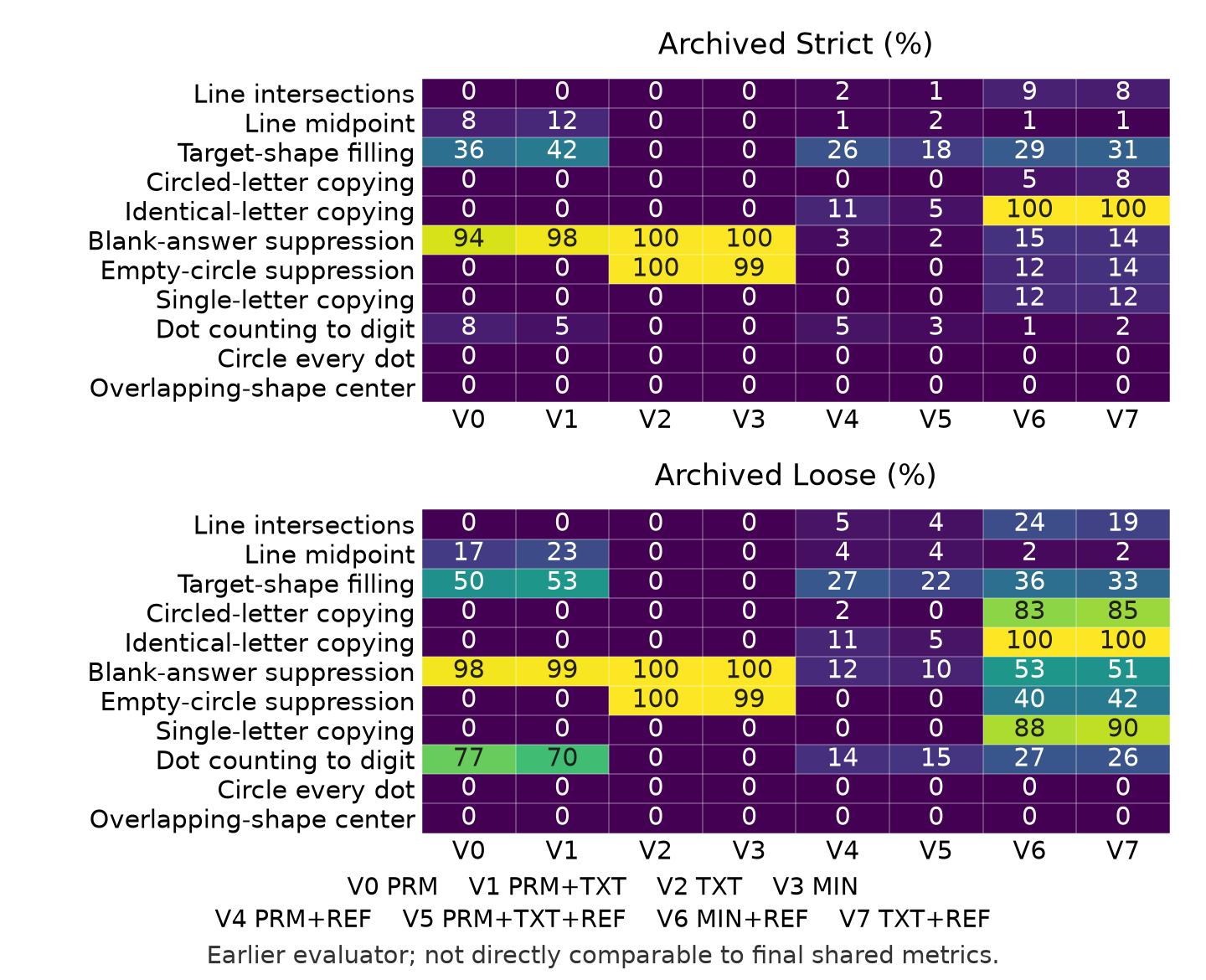}
\caption{Archived Strict/Loose pass rates for FLUX.2 Klein 4B API across V0--V7 (400 problems per task; 35,200 outputs). These scores use an earlier evaluator and are not directly comparable to the final shared Auto-Strict/Auto-Loose. High suppression scores can coexist with preservation failures. The shared evaluation uses the final scorer and replaces each task's index-zero problem to avoid overlap with reference examples. Values are rounded to integers: 0.25\% appears as 0 and 99.75\% as 100. The plot uses abbreviated task names: ``Circle every dot'' denotes Dot-Circling Marking, and ``Overlapping-shape center'' denotes Overlapping-Shape Center Marking.}
\label{fig:flux-heatmap}
\end{figure}

\noindent\textbf{Archived FLUX condition sweep.} Figure~\ref{fig:flux-heatmap} summarizes the full FLUX API sweep using an earlier evaluator. Its archived Strict/Loose scores are not directly comparable to the final shared Auto-Strict/Auto-Loose. Under the archived evaluator, Target-Shape Filling reaches 41.8\% Strict in its best individual condition, with an eight-condition oracle of 69.8\% Strict and 79.0\% Loose. Archived oracle Strict rates are 11.5\% for Line-Intersection Marking, 16.0\% for Line-Midpoint Marking, and 0.25\% for Dot-Circling Marking. Suppression scores require particular caution: in the single-rater audit, all 50 Blank-Answer outputs from V0/V1 pass archived Strict, but none passes Audit-Strict. Among unchanged shared items, the final scorer rejects 94 suppression outputs accepted by the archived rule because structure preservation is insufficient despite the absence of red marks. High archived suppression scores therefore do not establish faithful no-edit behavior.

\noindent\textbf{Failure forms in the single-rater audit.} The single-rater audit covers 2,200 FLUX API outputs (11 tasks $\times$ 25 problems $\times$ eight conditions), a different sample from the shared V0--V3 comparison. Human-coded Audit-Strict and Audit-Loose pass rates are 160/2,200 (7.3\%) and 683/2,200 (31.0\%), respectively; these criteria differ from the automatic proxies. Of the 2,040 Audit-Strict failures, the leading primary categories are omitted edits (567; 27.8\%), wrong targets (370; 18.1\%), structural damage or black-line redrawing (363; 17.8\%), no-edit violations (258; 12.6\%), wrong locations (153; 7.5\%), wrong glyphs (119; 5.8\%), and wrong counts (118; 5.8\%). The remaining 92 (4.5\%) comprise extra red marks (33), black erasure (18), wrong-region edits (15), black shifts (14), illegible glyphs (9), incomplete filling (2), and text changes (1). The audit records both answer-related errors and damage to the surrounding worksheet. This audit characterizes recurring failures; the separate three-rater study supports the reliability analysis.

\section{Diagnostic Analyses}
\label{sec:nb-analysis}
We next analyze instruction access and sensitivity, followed by reasoning-stress diagnostics.

\noindent\textbf{Instruction access and color conflict.} We use Nano Banana Pro for the D-, L-, and P-series. The D-series covers filling, line intersections, midpoint marking, circled-letter copying, and dot counting, with ten problems per task and condition. A separate red-mask overlap diagnostic classifies 41/50 image-only and 43/50 generic-prompt outputs as target-only edits. In the conflict condition, the image requests red ink and the external prompt requests blue; all 50 outputs are classified as blue-only in the worksheet body. With only a red-ink cue in the image, target-only classifications fall to 9/50. These color and overlap criteria differ from Auto-Strict.

\begin{figure}[!htbp]
\centering
\includegraphics[width=.40\linewidth]{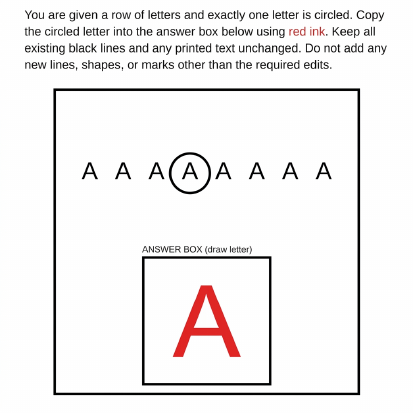}
\caption{Worksheet example for the P-series. The phrase ``red ink'' is red in the instruction region, and the answer is red in the box. The probes distinguish the words in the instruction from their displayed color; red already present in an input is not, by itself, evidence of a new edit.}
\label{fig:instruction-red-signal}
\end{figure}

The L-series tests 44 outputs per perturbation across the 11 tasks. Same-meaning paraphrases retain the intended task in 36/44 cases; keyword and role/order changes alter the task in 20/44 and 34/44 cases. Scrambled instructions still trigger edits in 35/44 cases. The supplementary material lists the instructions and outcome categories. In the P-series (Figure~\ref{fig:instruction-red-signal}), lexical color words and their displayed color are varied independently. Gray or blue phrases yield little instruction-region red signal, whereas red-rendered phrases trigger the detector. The measured flag therefore tracks displayed red content, including text already present in the input; it does not isolate whether the word ``red'' causes newly added marks.

\thisbottomragged

\noindent\textbf{Instruction-conditioned versus minimal-cue performance.} On the five spatial marking/filling tasks in the shared no-reference evaluation, Nano Banana Pro reaches 88.8\%, 88.4\%, and 88.4\% Auto-Strict with explicit instructions (V0--V2), but 2.4\% with the minimal cue (V3). Corresponding Auto-Loose rates are 91.6, 90.0, 91.2, and 6.4\%. These results show a substantial difference between explicit task instructions and the minimal cue under the shared evaluation protocol.

\noindent\textbf{Additional reference and oracle probes.} We collected additional outputs on five spatial marking/filling tasks for V4--V7, conflicting references (CREF), and FLUX oracle prompts (O1/O2). These probes are distinct from the archived full FLUX sweep. An audit revealed false rejections and target-matching inconsistencies in their archived evaluation. We therefore omit their pass rates pending re-evaluation with the final shared scorer and draw no quantitative conclusions from these additional probes.

\begin{figure}[!t]
\centering
\includegraphics[width=0.80\linewidth]{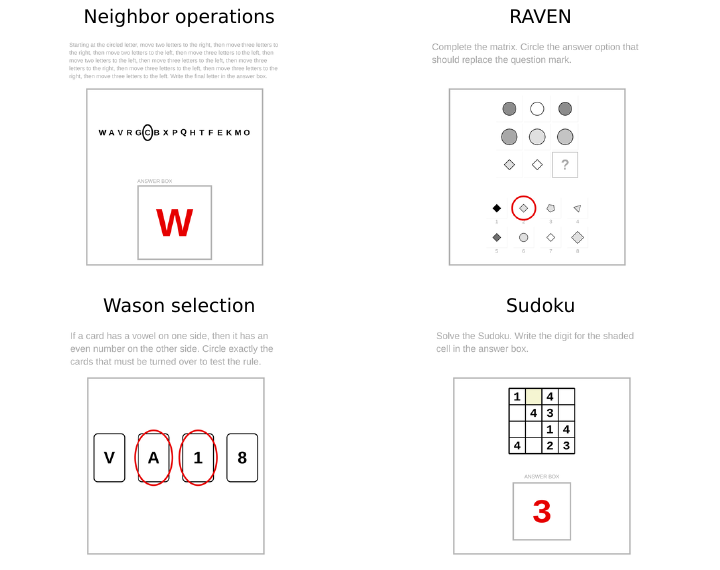}
\caption{Ground-truth examples for Table~\ref{tab:nb-reasoning}: follow left/right moves (Neighbor), complete a matrix (RAVEN), select cards to test a conditional (Wason), and fill a shaded cell (Sudoku). Red marks are prescribed answers, not model outputs.}
\label{fig:nb-reasoning}
\end{figure}

\noindent\textbf{Reasoning-stress tasks.} We evaluate 398 distinct additional Nano Banana Pro outputs: 378 in Table~\ref{tab:nb-reasoning} (examples in Figure~\ref{fig:nb-reasoning}) and 20 Sudoku contradiction variants. The probes target sequential state updates, abstract analogy, conditional selection, and constraint satisfaction. They are supplementary diagnostics, not core WISP templates or generalization tests.

\begin{table}[!t]
\caption{Nano Banana Pro reasoning-stress scores (378 items; percentages except $n$). Wason uses exact card-set selection; other rows use archived image-based proxies. Score+Fmt. additionally requires native format. The 20 Sudoku contradiction variants are separate.}
\label{tab:nb-reasoning}
\centering
\begin{tabular}{lrrrr}
\toprule
Probe & $n$ & Task score & Format & Score+Fmt. \\
\midrule
Neighbor, 1--10 ops & 100 & 94.0 & 100.0 & 94.0 \\
Neighbor, 10--300 ops & 98 & 91.8 & 99.0 & 91.8 \\
Public RAVEN & 70 & 22.9 & 100.0 & 22.9 \\
Wason selection & 100 & 49.0 & 97.0 & 49.0 \\
Sudoku $4\times4$ & 10 & 70.0 & 100.0 & 70.0 \\
\bottomrule
\end{tabular}
\end{table}

The neighbor-symbol task follows rendered left/right moves to a final symbol. The 1--10-operation schedule yields a 94.0\% proxy pass rate on 100 items, including 90.0\% at ten operations. A separate extension tests 10, 20, 30, 40, 50, 75, 100, 150, 200, and 300 operations. After removing two duplicate entries for a reused replacement output, it contains 98 distinct outputs: eight at 200 operations and ten at each other level. The proxy passes 90/98 (91.8\%). This coarser schedule probes longer sequences, with ten operations included in both schedules. The observed rates do not decrease monotonically across this sampled schedule: 100.0\% at 200 and 300 operations, but 70.0\% at 75 and 100. These small samples do not establish a maximum operation count or reliable performance on long sequences in general.

In public RAVEN~\cite{zhang2019raven}, the model chooses a missing matrix element from eight options. Across 70 items sampled evenly from seven configurations, the image-overlap proxy passes 16 outputs (22.9\%). It does not enforce selection of only one option, so this rate is not exact-choice accuracy. The Wason selection task~\cite{wason1968reasoning} asks which cards must be turned over to test a conditional rule. Across 100 items in five surface variants, red-region overlap passes 99.0\%, but selecting both and only $P$ and $\neg Q$ succeeds in 49/100; 97/100 outputs have valid native format. Exact success ranges from 19/28 for canonical vowel/even wording to 6/18--10/18 for the other surface variants. Overlap therefore overstates multi-card success.

The $4\times4$ Sudoku probe fills a shaded cell, yielding a 70.0\% proxy pass rate on ten items. All ten unsatisfiable variants explicitly requesting ``NONE'' return it, but two use black instead of red, so 80.0\% satisfy both the answer and color requirements. Without that instruction, none of ten unsatisfiable variants returns ``NONE'': nine return digits and one leaves the answer blank. Some outputs also alter puzzle cells or other worksheet content. Explicit no-solution compliance thus differs from spontaneous inconsistency detection. All reasoning instructions are rendered inside the image; the external prompt specifies only red-answer output and preservation.

\section{Discussion, Limitations, and Conclusion}
\noindent\textbf{What the benchmark measures.} Target success and final success differ because preservation and native format also matter. OCR+Gemini controls drawing, but still depends on OCR, candidate extraction, planning, and revision. Neither the target criterion nor this diagnostic isolates reasoning from localization and rendering.

\noindent\textbf{Instruction following and rule induction.} Nano Banana Pro's strong TXT performance supports rendered text as a task specification for this editor. The low shared MIN score shows difficulty on spatial tasks when task-specific instructions and visual references are absent. Familiar formats or learned templates may contribute to the reasoning-probe results; these data do not separate those effects from stepwise computation or establish transfer to unfamiliar rules, layouts, or symbols.

\noindent\textbf{Reasoning-stress comparisons.} Task and scoring differences limit comparisons between symbol tracking, RAVEN, Wason, and Sudoku. Wason outputs can overlap the correct cards without selecting the exact pair; Sudoku distinguishes explicit no-solution compliance from spontaneous inconsistency detection. These probes motivate reporting answer correctness separately from edit execution and format.

\noindent\textbf{Evaluation and scope.} Three-rater validation covers 440 outputs from five editors; future systems require task-wise analysis of false positives and negatives. Our limited model set and changing proprietary endpoints prevent controlled attribution to scale, architecture, or training. OCR+Gemini uses a different model and a revision procedure, so it is not a matched ablation of Nano Banana Pro.

\noindent\textbf{Future directions.} A concrete next experiment is to let an editor mark its intermediate position after each move in the neighbor task, then test whether these visible states improve final answers or merely propagate early mistakes. Held-out rules, unfamiliar symbols, counterfactual layouts, and permuted instructions could test transfer. Applying this interface to natural documents, user interfaces, or scientific diagrams would require separate measures of answer correctness and edit preservation.

\noindent\textbf{Conclusion.} Nano Banana Pro leads the evaluated editors on Auto-Strict, including when no-edit controls are excluded. For Nano Banana Pro, explicit instructions support spatial edits on the shared no-reference tasks. Decomposed decisions and human validation reveal distinct failures in target execution, preservation, suppression, and format. Producing correct answers without unrelated edits remains difficult.

\subsubsection*{Acknowledgments}
We thank the anonymous reviewers for their constructive comments and suggestions. This work was conducted as part of the Experts in Information Science Program at the National Institute of Informatics (NII), with financial support provided by NII.
\bibliographystyle{splncs04}
\bibliography{main}
\end{document}


\title{\texorpdfstring{Image-Space Rule Discovery:\\Supplementary Material}{Image-Space Rule Discovery: Supplementary Material}}
\titlerunning{Image-Space Rule Discovery: Supplementary Material}
\author{Misora Sugiyama\inst{1,2}\orcidlink{0009-0006-1245-8619} \and Toya Oyama\inst{3,2}\orcidlink{0009-0007-8108-6529} \and Hirokatsu Kataoka\inst{4,5,2}\orcidlink{0000-0001-8844-165X}}
\authorrunning{M.~Sugiyama et al.}
\institute{New York University \and LIMIT.Lab \and The University of Tokyo \and National Institute of Advanced Industrial Science and Technology (AIST) \and Visual Geometry Group, University of Oxford}
\maketitle
\flushbottom

\section{Introduction}
This supplementary material supports the main paper by providing additional details about benchmark generation, controlled information conditions, automatic scoring, diagnostic probes, and result tables. The main paper is self-contained; this document is intended to make the worksheet protocol and reported diagnostics easier to reproduce and audit. End-to-end editors use a first-shot policy, with retries only for transport or file failures. OCR+Gemini is a separate diagnostic. It permits at most one plan revision after validation against the input. Automatic pass rates are operational proxies rather than semantic accuracy.

\section{Benchmark Generator and Worksheet Format}

\subsection{Canvas and rendering constants}
All core worksheet instances are rendered at a fixed $1024\times 1024$ resolution. Rendering is performed at a higher internal scale and then downsampled to reduce aliasing. Table~\ref{tab:layout} summarizes the layout constants used by the generator.

\par\begingroup
\footnotesize
\captionsetup{hypcap=false,skip=4pt}
\setlength{\LTcapwidth}{\textwidth}
\begin{xltabular}{\textwidth}{lZ}
\caption{Core worksheet layout and rendering constants. Coordinates are in final 1024-pixel image space unless otherwise stated.}\label{tab:layout}\\
\toprule
Item & Value or rule \\

\midrule
\endfirsthead
\toprule
Item & Value or rule \\

\midrule
\endhead
\bottomrule
\endfoot
Canvas & $1024\times1024$ RGB worksheet on a white background. \\
Internal rendering scale & $4\times$ render-and-downsample pipeline. \\
Header region & Top 200 px. Rendered in-image instructions, when present, are placed in this region. \\
Instruction text box & Horizontal margin 60 px; approximately $x\in[60,964]$ and $y\in[20,190]$. \\
Problem frame & Light-gray square frame of side 760 px, with top-left coordinate $(132,220)$ and bottom-right coordinate $(892,980)$. \\
Problem content & Black visual elements are rendered inside the frame. Text-answer tasks include an answer box inside the frame. \\
Answer convention & Required answer edits are red. In the blue-conflict diagnostic only, the external prompt requests blue. \\
Colors & White $(255,255,255)$; black $(0,0,0)$; gray instruction text $(170,170,170)$; gray frame $(180,180,180)$; red $(230,0,0)$; blue $(0,90,255)$. \\
Line widths & Frame width $6s$; black input strokes $8s$; gray strokes $8s$; dot radius $10s$, where $s=4$ is the internal render scale. \\
Fonts & A standard sans-serif font is used when available; the generator searches system sans-serif fonts and falls back to the available default. \\
\end{xltabular}
\endgroup

\subsection{Dataset records and generated artifacts}
For each task instance and information condition, the generator creates the rendered input worksheet, the ground-truth output worksheet, and task metadata. The fixed public dataset stores inputs, ground-truth outputs, and metadata; its evaluator derives masks from the images during scoring. A dataset record contains at least the following fields:
\begin{itemize}[leftmargin=1.2em,itemsep=1pt]
    \item \texttt{id}: unique instance identifier containing task, base index, and condition ID;
    \item \texttt{task}: task template name;
    \item \texttt{idx}: base problem index;
    \item \texttt{var\_id}: information condition, V0--V7;
    \item \texttt{input\_path} and \texttt{gt\_path}: rendered input and ground-truth output images;
    \item \texttt{prompt}: external prompt sent to the editing model;
    \item \texttt{meta}: scoring metadata, such as answer-box coordinates, target points, dot coordinates, target letters, target digit, or shape centers.
\end{itemize}
The core generation setting uses 11 tasks, 400 base instances per task, and eight information conditions, yielding $11\times400\times8=35{,}200$ dataset rows. The shared cross-model subset uses 50 base instances per task and the four no-reference conditions V0--V3, yielding $11\times50\times4=2{,}200$ outputs per model. Before the shared freeze, the index-zero evaluation problem in each of the 11 tasks was replaced by the index-50 problem to avoid overlap with reference examples, affecting 44 shared records. Reference images were kept fixed. The archived full-sweep scores retain the earlier evaluation set.

\subsection{Core task templates}
Table~\ref{tab:tasks} lists the 11 core worksheet tasks. Red is reserved for the answer edit, and every task requires all non-target content to remain unchanged.

\par\begingroup
\fontsize{8}{10}\selectfont
\captionsetup{hypcap=false,skip=4pt}
\setlength{\LTcapwidth}{\textwidth}
\begin{xltabular}{\textwidth}{L{.13\textwidth}L{.26\textwidth}L{.17\textwidth}Z}
\caption{Core worksheet task templates. ``Output'' describes the required red edit.}\label{tab:tasks}\\
\toprule
Family & Task ID & Short name & Output requirement \\

\midrule
\endfirsthead
\toprule
Family & Task ID & Short name & Output requirement \\

\midrule
\endhead
\bottomrule
\endfoot
Marking & \texttt{line\_\allowbreak intersections} & Line intersections & Place a small red dot at every intersection of the black line segments. \\
Marking & \texttt{line\_\allowbreak midpoint\_\allowbreak mark} & Line midpoint & Place a small red dot at the midpoint of the black line segment. \\
Marking & \texttt{count\_\allowbreak dots\_\allowbreak visual} & Circle every dot & Draw a thin red circle around each black dot, with no extra circles. \\
Marking & \texttt{overlapping\_\allowbreak shapes} & Shape centers & Mark the center of every outlined shape with a small red dot. \\
Filling & \texttt{fill\_\allowbreak target\_\allowbreak shape} & Target-shape filling & Fill the outlined target shape solid red. \\
Copying & \texttt{circled\_\allowbreak letter\_\allowbreak base} & Circled-letter copying & Copy the circled letter into the answer box using red ink. \\
Copying & \texttt{circled\_\allowbreak letter\_\allowbreak allA} & All-A control & Copy the circled letter in a row of identical letters. \\
Copying & \texttt{circled\_\allowbreak letter\_\allowbreak copy\_\allowbreak single} & Single-letter copying & Copy the single circled letter into the answer box using red ink. \\
Counting & \texttt{count\_\allowbreak dots\_\allowbreak digit} & Dot counting to digit & Count black dots and write the correct digit in red ink. \\
Suppression & \texttt{circled\_\allowbreak letter\_\allowbreak blank} & Blank response & Leave the answer box blank and do not draw anything. \\
Suppression & \texttt{circled\_\allowbreak letter\_\allowbreak blankcircle} & Empty-circle response & If the circled position is blank, leave the answer box blank. \\
\end{xltabular}
\endgroup

\section{Controlled Information Conditions}

\subsection{V0--V7 condition definitions}
The eight controlled information conditions vary three channels: external prompt specificity, rendered in-image instruction, and visual references. Table~\ref{tab:conditions} gives the condition definitions used across tasks.

\par\begingroup
\fontsize{8}{10}\selectfont
\captionsetup{hypcap=false,skip=4pt}
\setlength{\LTcapwidth}{\textwidth}
\begin{xltabular}{\textwidth}{lp{.20\textwidth}L{.20\textwidth}L{.10\textwidth}Z}
\caption{Information conditions. ``Strong'' means a task-specific external prompt containing the operation and preservation constraints. ``Generic'' means the external prompt only asks the model to follow the written instruction in the image. ``Minimal'' is the prompt \emph{Edit the image.}}\label{tab:conditions}\\
\toprule
ID & Label & External prompt & Image text & Visual refs / purpose \\

\midrule
\endfirsthead
\toprule
ID & Label & External prompt & Image text & Visual refs / purpose \\

\midrule
\endhead
\bottomrule
\endfoot
V0 & PRM & Strong & No & None. Tests ordinary prompt-driven editing. \\
V1 & PRM+\allowbreak TXT & Strong & Yes & None. Tests prompt plus redundant rendered instruction. \\
V2 & TXT & Generic & Yes & None. Tests whether rendered image text can specify the task. \\
V3 & MIN & Minimal & No & None. Tests weak visual-structure-only behavior. \\
V4 & PRM+\allowbreak REF & Strong & No & Style and in-context references. \\
V5 & PRM+\allowbreak TXT+\allowbreak REF & Strong & Yes & Style and in-context references. \\
V6 & MIN+\allowbreak REF & Minimal & No & References without task-specific prompt or image text. \\
V7 & TXT+\allowbreak REF & Generic & Yes & Rendered image text plus references. \\
\end{xltabular}
\endgroup

\paragraph{Visual reference construction.}
For V4--V7, the reference images are auxiliary images supplied with the editing request; they are not the target worksheet itself. We use two lightweight reference types. A \emph{style reference} shows the output appearance of the required edit, such as the intended red fill, red dot, red ring, or red answer stroke, while omitting the specific target geometry of the current worksheet. An \emph{in-context reference} shows a separate miniature worksheet-like example with the same operation applied to a different object or layout. Thus, references can demonstrate the visual form of the edit and, in some conditions, the input-output relationship, but they do not provide the exact answer for the current item. In the conflicting-reference condition, the reference cue is intentionally made incompatible with the worksheet instruction; those outputs are scored only against the canonical worksheet target.

\subsection{Prompt templates}
A strong prompt combines the task context and instruction, any extra constraints, and a common preservation constraint. The common preservation constraint is:
\begin{quote}\small
Keep all existing black lines and any printed text unchanged. Do not add any new lines, shapes, or marks other than the required edits.
\end{quote}
The generic weak prompt is:
\begin{quote}\small
Edit the image by following the written instructions that appear inside the image.
\end{quote}
The minimal prompt is:
\begin{quote}\small
Edit the image.
\end{quote}
For rendered image text, the header contains the task context and the task instruction, without task IDs or hidden labels. Examples of task-specific instructions are shown in Table~\ref{tab:instruction-examples}.

\par\begingroup
\fontsize{8}{10}\selectfont
\captionsetup{hypcap=false,skip=4pt}
\setlength{\LTcapwidth}{\textwidth}
\begin{xltabular}{\textwidth}{lZ}
\caption{Representative rendered instruction templates. The full generator uses the same style for all tasks.}\label{tab:instruction-examples}\\
\toprule
Task & Rendered task instruction \\

\midrule
\endfirsthead
\toprule
Task & Rendered task instruction \\

\midrule
\endhead
\bottomrule
\endfoot
Target-shape filling & You are given an outlined target shape inside a light gray square frame. Fill the outlined target shape solid red. \\
Line midpoint & You are given a black line segment inside a light gray square frame. Place a small red dot at the midpoint of the line segment. \\
Line intersections & You are given black line segments inside a light gray square frame. Place a small red dot at every intersection point of the black lines. \\
Circled-letter copying & You are given a row of letters and exactly one letter is circled. Copy the circled letter into the answer box below using red ink. \\
Dot counting to digit & You are given black dots inside a light gray square frame. Count the dots and write the correct digit from 1 to 9 in the answer box below using red ink. \\
Circle every dot & You are given black dots inside a light gray square frame. Draw a thin red circle around each black dot. \\
Blank response & You are given a row of letters and exactly one position is circled. Leave the answer box below completely blank and do not draw anything. \\
Empty-circle response & You are given a row of letters and exactly one position is circled. Copy the circled letter into the answer box below using red ink. If the circled position is blank, leave the answer box below completely blank. \\
Overlapping shapes & You are given multiple overlapping outlined shapes inside a light gray square frame. Mark the center of every shape with a small red dot. \\
\end{xltabular}
\endgroup

\section{Automatic Scoring Protocol}

\subsection{Masks and diagnostic metrics}
The shared V0--V3 comparison uses the final public evaluator. Native output dimensions are checked before bilinear resizing to the input dimensions for mask computation; a size mismatch fails Auto-Strict. Auto-Loose does not require native size. Red answer pixels satisfy
\[
R>150,\quad G<130,\quad B<130,\quad R-G>35,\quad R-B>35.
\]
Black and structure masks use mean RGB intensity below 90 and 225, respectively, excluding red and blue pixels. The blue mask uses $B>140$, $R<130$, $G<170$, and $B-R>35$. For masks $A$ and $B$, Dice is $2|A\cap B|/(|A|+|B|)$. Black-F1 and structure-F1 compare the corresponding input and output masks using $2TP/(2TP+FP+FN)$. Empty-mask comparisons return 1. These metrics enter task-dependent proxy rules; they do not independently establish semantic correctness.

\subsection{Task-specific proxy rules}
Table~\ref{tab:scoring} specifies the final shared scorer. For component-based tasks, eight-connected components smaller than eight pixels are discarded. Ground-truth and candidate component centroids are greedily matched one-to-one within the stated distance. Strict point scoring also caps each candidate component's area at $\max(1200,\lfloor4\,\mathrm{median}(\text{GT component areas})\rfloor)$ pixels. Symbol rules use the ground-truth red bounding box padded by 12 pixels, not the worksheet answer-box boundary. No OCR or glyph-identity recognition is performed by these rules.

\par\begingroup
\small
\captionsetup{hypcap=false,skip=4pt}
\setlength{\LTcapwidth}{\textwidth}
\begin{xltabular}{\textwidth}{L{.20\textwidth}ZZ}
\caption{Final shared-evaluation proxy rules. RD is red Dice, BF1 is black-F1, and SF1 is structure-F1. Auto-Strict additionally requires native canvas dimensions in every row.}\label{tab:scoring}\\
\toprule
Task group & Auto-Strict & Auto-Loose \\

\midrule
\endfirsthead
\toprule
Task group & Auto-Strict & Auto-Loose \\

\midrule
\endhead
\bottomrule
\endfoot
Target-shape filling & RD $\ge .70$ and BF1 $\ge .85$. & RD $\ge .45$. \\
Point marking & All GT components matched within 18 px, no extra components, area cap above, BF1 $\ge .85$. & All GT components matched within 30 px and at most one extra component. \\
Circle every dot & All GT components matched within 22 px, no extra components, BF1 $\ge .85$. & All GT components matched within 36 px and at most one extra component. \\
Letter/digit answers & RD $\ge .35$, at least 20 red pixels inside padded GT box, at most 250 outside, SF1 $\ge .75$. & At least 20 red pixels inside padded GT box, at most 1,000 outside, SF1 $\ge .60$. \\
Suppression & At most 25 red pixels in the whole canvas, SF1 $\ge .90$. & At most 150 red pixels in the whole canvas, SF1 $\ge .75$. \\
\end{xltabular}
\endgroup
If an answer-task ground truth has no red pixels, the scorer uses the suppression thresholds. The decomposition into target success, preservation, and native format reproduces all 13,200 frozen shared Auto-Strict decisions. In the main paper's Nano Banana Pro copying validation subset, 9/24 outputs pass the human majority and 3/24 pass Auto-Strict. All six false negatives exceed the limit on red outside the padded ground-truth glyph region; these sample-level errors do not determine the semantic success rate across all 600 copying outputs. The large FLUX V0--V7 sweep is reported separately using archived Strict/Loose scores, which are not directly comparable to the final shared metrics. Archived pass rates for the additional reference, conflicting-reference, and oracle-prompt probes are excluded because of target-matching inconsistencies.

\subsection{Oracle condition}
The condition oracle is not repeated sampling under a single condition. It reports whether a base problem passes the relevant evaluator under any tested information condition. For V0--V7, the same base problem is grouped across the eight conditions. The large-scale FLUX oracle uses the archived evaluator and is distinct from the O1/O2 prompts, which supply additional information to the model.

\section{Additional Quantitative Results}

\subsection{Shared cross-model summary}
Table~\ref{tab:crossmodel} gives the shared all-task proxy strict summary reported for the V0--V3 subset. The subset contains 11 tasks, 50 base problems per task, and four information conditions, or 2,200 outputs per model.

\par\begingroup
\small
\captionsetup{hypcap=false,skip=4pt}
\setlength{\LTcapwidth}{\textwidth}
\begin{longtable}{lrr}
\caption{Shared V0--V3 proxy pass rates over 2,200 outputs per system. OCR+Gemini is a separate structured-plan diagnostic.}\label{tab:crossmodel}\\
\toprule
System & Auto-Strict (\%) & Auto-Loose (\%) \\

\midrule
\endfirsthead
\toprule
System & Auto-Strict (\%) & Auto-Loose (\%) \\

\midrule
\endhead
\bottomrule
\endfoot
Nano Banana Pro & 48.7 & 64.0 \\
Qwen-Image-Edit & 13.4 & 20.4 \\
FLUX.2 Klein 4B API & 11.5 & 21.9 \\
FLUX.2 Klein 4B open-weight & 11.3 & 21.8 \\
InstructPix2Pix & 0.0 & 5.5 \\
\midrule
OCR+Gemini plan-render & 35.0 & 58.0 \\
\end{longtable}
\endgroup

\subsection{\nbp{} V0--V3 condition summary}
Table~\ref{tab:nb-all} summarizes the Nano Banana Pro condition summary over all 11 tasks for V0--V3.
\par\begingroup
\fontsize{8}{10}\selectfont
\captionsetup{hypcap=false,skip=4pt}
\setlength{\LTcapwidth}{\textwidth}
\begin{longtable}{lrrrrrr}
\caption{\nbp{} condition summary over all 11 tasks, V0--V3. Strict/Loose are Auto-Strict/Auto-Loose; pass and format rates are percentages.}\label{tab:nb-all}\\
\toprule
Condition & $n$ & Strict & Loose & Format & Red Dice & Black-F1 \\

\midrule
\endfirsthead
\toprule
Condition & $n$ & Strict & Loose & Format & Red Dice & Black-F1 \\

\midrule
\endhead
\bottomrule
\endfoot
V0 PRM & 550 & 55.8 & 79.1 & 100.0 & 0.542 & 0.761 \\
V1 PRM+TXT & 550 & 64.4 & 77.1 & 100.0 & 0.560 & 0.823 \\
V2 TXT & 550 & 63.3 & 79.5 & 100.0 & 0.569 & 0.851 \\
V3 MIN & 550 & 11.3 & 20.2 & 99.8 & 0.195 & 0.510 \\
\end{longtable}
\endgroup

\par\begingroup
\fontsize{8}{10}\selectfont
\captionsetup{hypcap=false,skip=4pt}
\setlength{\LTcapwidth}{\textwidth}
\begin{longtable}{lrrrrrr}
\caption{\nbp{} condition summary over the five active marking/filling tasks. Strict/Loose are Auto-Strict/Auto-Loose; pass and format rates are percentages.}\label{tab:nb-active}\\
\toprule
Condition & $n$ & Strict & Loose & Format & Red Dice & Black-F1 \\

\midrule
\endfirsthead
\toprule
Condition & $n$ & Strict & Loose & Format & Red Dice & Black-F1 \\

\midrule
\endhead
\bottomrule
\endfoot
V0 PRM & 250 & 88.8 & 91.6 & 100.0 & 0.628 & 0.969 \\
V1 PRM+TXT & 250 & 88.4 & 90.0 & 100.0 & 0.637 & 0.952 \\
V2 TXT & 250 & 88.4 & 91.2 & 100.0 & 0.638 & 0.944 \\
V3 MIN & 250 & 2.4 & 6.4 & 99.6 & 0.037 & 0.479 \\
\end{longtable}
\endgroup

\paragraph{TXT-slice clarification.}
The all-task V2/TXT slice for Nano Banana Pro is shown in Table~\ref{tab:nb-all}: 63.3\% Strict and 79.5\% Loose over all 11 tasks. The 88.4\% Strict and 91.2\% Loose TXT values correspond to the five active marking/filling task subset in Table~\ref{tab:nb-active}, which is the subset used for the reference, conflicting-reference, and oracle diagnostic ablations.

\subsection{\nbp{} task-by-condition Auto-Strict pass rate}
Table~\ref{tab:nb-task} expands the \nbp{} Auto-Strict proxy pass rate by task and V0--V3 condition. Each cell uses 50 outputs.

\par\begingroup
\renewcommand{\arraystretch}{1.16}
\fontsize{8}{10}\selectfont
\captionsetup{hypcap=false,skip=4pt}
\setlength{\LTcapwidth}{\textwidth}
\begin{longtable}{lrrrr}
\caption{\nbp{} Auto-Strict pass rate by task and V0--V3 condition. Values are percentages.}\label{tab:nb-task}\\
\toprule
Task & V0 & V1 & V2 & V3 \\

\midrule
\endfirsthead
\toprule
Task & V0 & V1 & V2 & V3 \\

\midrule
\endhead
\bottomrule
\endfoot
Circled-letter all-A control & 6.0 & 20.0 & 6.0 & 0.0 \\
Circled-letter copying & 6.0 & 6.0 & 0.0 & 0.0 \\
Blank response & 66.0 & 100.0 & 100.0 & 70.0 \\
Empty-circle response & 20.0 & 56.0 & 66.0 & 42.0 \\
Single-letter copying & 2.0 & 6.0 & 2.0 & 0.0 \\
Dot counting to digit & 70.0 & 78.0 & 80.0 & 0.0 \\
Circle every dot & 78.0 & 86.0 & 84.0 & 0.0 \\
Target-shape filling & 100.0 & 96.0 & 98.0 & 10.0 \\
Line intersections & 92.0 & 100.0 & 100.0 & 2.0 \\
Line midpoint & 84.0 & 70.0 & 68.0 & 0.0 \\
Overlapping shapes & 90.0 & 90.0 & 92.0 & 0.0 \\
\end{longtable}
\endgroup

\subsection{Large-scale \flux{} API archived Strict table}
Table~\ref{tab:flux-strict} gives the archived Strict pass rates underlying the \flux{} API heatmap for the full V0--V7 condition sweep. Each cell uses 400 outputs. Values are rounded to the nearest integer, so 0.25\% can appear as 0 and 99.75\% as 100. These scores use an earlier evaluator and are not directly comparable to the final shared Auto-Strict/Auto-Loose metrics. In particular, high archived suppression scores can coexist with structure-preservation failures and do not establish faithful no-edit behavior.

\par\begingroup
\renewcommand{\arraystretch}{1.16}
\fontsize{8}{10}\selectfont
\captionsetup{hypcap=false,skip=4pt}
\setlength{\LTcapwidth}{\textwidth}
\begin{xltabular}{\textwidth}{Zrrrrrrrr@{}}
\caption{\flux{} API archived Strict pass rate by task and condition. Values are percentages from an earlier evaluator, not final shared Auto-Strict scores.}\label{tab:flux-strict}\\
\toprule
Task & V0 & V1 & V2 & V3 & V4 & V5 & V6 & V7 \\

\midrule
\endfirsthead
\toprule
Task & V0 & V1 & V2 & V3 & V4 & V5 & V6 & V7 \\

\midrule
\endhead
\bottomrule
\endfoot
Line intersections & 0 & 0 & 0 & 0 & 2 & 1 & 9 & 8 \\
Line midpoint & 8 & 12 & 0 & 0 & 1 & 2 & 1 & 1 \\
Fill target shape & 36 & 42 & 0 & 0 & 26 & 18 & 29 & 31 \\
Copy circled letter & 0 & 0 & 0 & 0 & 0 & 0 & 5 & 8 \\
All-A control & 0 & 0 & 0 & 0 & 11 & 5 & 100 & 100 \\
Blank response & 94 & 98 & 100 & 100 & 3 & 2 & 15 & 14 \\
Blank-circle response & 0 & 0 & 100 & 99 & 0 & 0 & 12 & 14 \\
Copy single letter & 0 & 0 & 0 & 0 & 0 & 0 & 12 & 12 \\
Count dots, write digit & 8 & 5 & 0 & 0 & 5 & 3 & 1 & 2 \\
Circle every dot & 0 & 0 & 0 & 0 & 0 & 0 & 0 & 0 \\
Overlapping shapes & 0 & 0 & 0 & 0 & 0 & 0 & 0 & 0 \\
\end{xltabular}
\endgroup

\section{Human Task-Solvability Sanity Check}
The main paper reports a solve-only human sanity check to verify that the worksheet questions themselves are generally solvable by people. The scored summary uses four anonymized participants and 110 sampled worksheet items, for 440 rateable responses in total. Participants answered by clicking or typing rather than by performing pixel-level image editing, so this is not a human image-editing baseline. Four unpaid adult volunteers, one author and three non-authors, participated. Two had a computer-science background and two did not. Before participating, volunteers were informed that their responses would be used for this research. No identifying information was collected. Only pooled summaries are released; participant-level responses and summaries remain private.

\par\begingroup

\captionsetup{hypcap=false,skip=4pt}
\setlength{\LTcapwidth}{\textwidth}
\begin{longtable}{lrrrrr}
\caption{Human solve-only sanity check. Strict and loose accuracy are percentages.}\label{tab:human-solve-overall}\\
\toprule
Participants used & Responses & Rateable & Items & Strict (\%) & Loose (\%) \\

\midrule
\endfirsthead
\toprule
Participants used & Responses & Rateable & Items & Strict (\%) & Loose (\%) \\

\midrule
\endhead
\bottomrule
\endfoot
4 & 440 & 440 & 110 & 96.1 & 99.8 \\
\end{longtable}
\endgroup

\par\begingroup
\fontsize{8}{10}\selectfont
\captionsetup{hypcap=false,skip=4pt}
\setlength{\LTcapwidth}{\textwidth}
\begin{longtable}{lrrrrr}
\caption{Human solve-only accuracy by task. Values pool four participants; individual responses remain private.}\label{tab:human-solve-task}\\
\toprule
Task & Responses & Participants & Items & Strict (\%) & Loose (\%) \\

\midrule
\endfirsthead
\toprule
Task & Responses & Participants & Items & Strict (\%) & Loose (\%) \\

\midrule
\endhead
\bottomrule
\endfoot
All-A control & 40 & 4 & 10 & 100.0 & 100.0 \\
Blank response & 40 & 4 & 10 & 100.0 & 100.0 \\
Circle every dot & 40 & 4 & 10 & 60.0 & 100.0 \\
Circled-letter copying & 40 & 4 & 10 & 100.0 & 100.0 \\
Dot counting to digit & 40 & 4 & 10 & 100.0 & 100.0 \\
Empty-circle response & 40 & 4 & 10 & 100.0 & 100.0 \\
Line intersections & 40 & 4 & 10 & 97.5 & 97.5 \\
Line midpoint & 40 & 4 & 10 & 100.0 & 100.0 \\
Overlapping shapes & 40 & 4 & 10 & 100.0 & 100.0 \\
Single-letter copying & 40 & 4 & 10 & 100.0 & 100.0 \\
Target-shape filling & 40 & 4 & 10 & 100.0 & 100.0 \\
\end{longtable}
\endgroup

\section{Reference, Conflicting-Reference, and Oracle Ablations}
\par\begingroup
\fontsize{8}{10}\selectfont
\captionsetup{hypcap=false,skip=4pt}
\setlength{\LTcapwidth}{\textwidth}
\begin{xltabular}{\textwidth}{ZrrZ}
\caption{Large-scale FLUX.2 Klein 4B API archived condition-oracle scores. Using the earlier evaluator, the oracle counts a base problem as passing if any V0--V7 condition passes. Values are percentages and are not directly comparable to the final shared Auto-Strict/Auto-Loose metrics.}\label{tab:flux-large-oracle-summary}\\
\toprule
Task & Oracle Strict & Oracle Loose & Notes \\

\midrule
\endfirsthead
\toprule
Task & Oracle Strict & Oracle Loose & Notes \\

\midrule
\endhead
\bottomrule
\endfoot
Target-shape filling & 69.8 & 79.0 & Full V0--V7 condition oracle. \\
Line intersections & 11.5 & -- & Strict value quoted in the main paper. \\
Line midpoint & 16.0 & -- & Strict value quoted in the main paper. \\
Circle every dot & 0.25 & -- & Strict value quoted in the main paper. \\
\end{xltabular}
\endgroup

We collected additional V4--V7, conflicting-reference (CREF), and FLUX oracle-prompt (O1/O2) outputs on five spatial marking/filling tasks. An audit found target-matching inconsistencies in their archived evaluation. We omit those pass rates, tables, and summary plot pending evaluation with the final shared scorer. These additional records are distinct from the shared V0--V3 comparison and the full FLUX sweep.

\paragraph{Condition definitions.}
V4--V7 follow Table~\ref{tab:conditions}. CREF supplies visual reference cues that conflict with the worksheet instruction; its intended scoring target is the canonical worksheet answer. O1 provides an explicit task-specific edit instruction, and O2 additionally identifies the target or answer region.

\begin{figure}[!htbp]
  \centering
  \setlength{\tabcolsep}{4pt}
  \renewcommand{\arraystretch}{0.92}
  \fontsize{8}{10}\selectfont

  \resizebox{0.98\textwidth}{!}{%
  \begin{tabular}{@{}ccc@{}}
    \textbf{Input} & \textbf{Nano Banana Pro output} & \textbf{Ground truth} \\
    \includegraphics[width=0.31\textwidth,height=0.173\textheight,keepaspectratio]{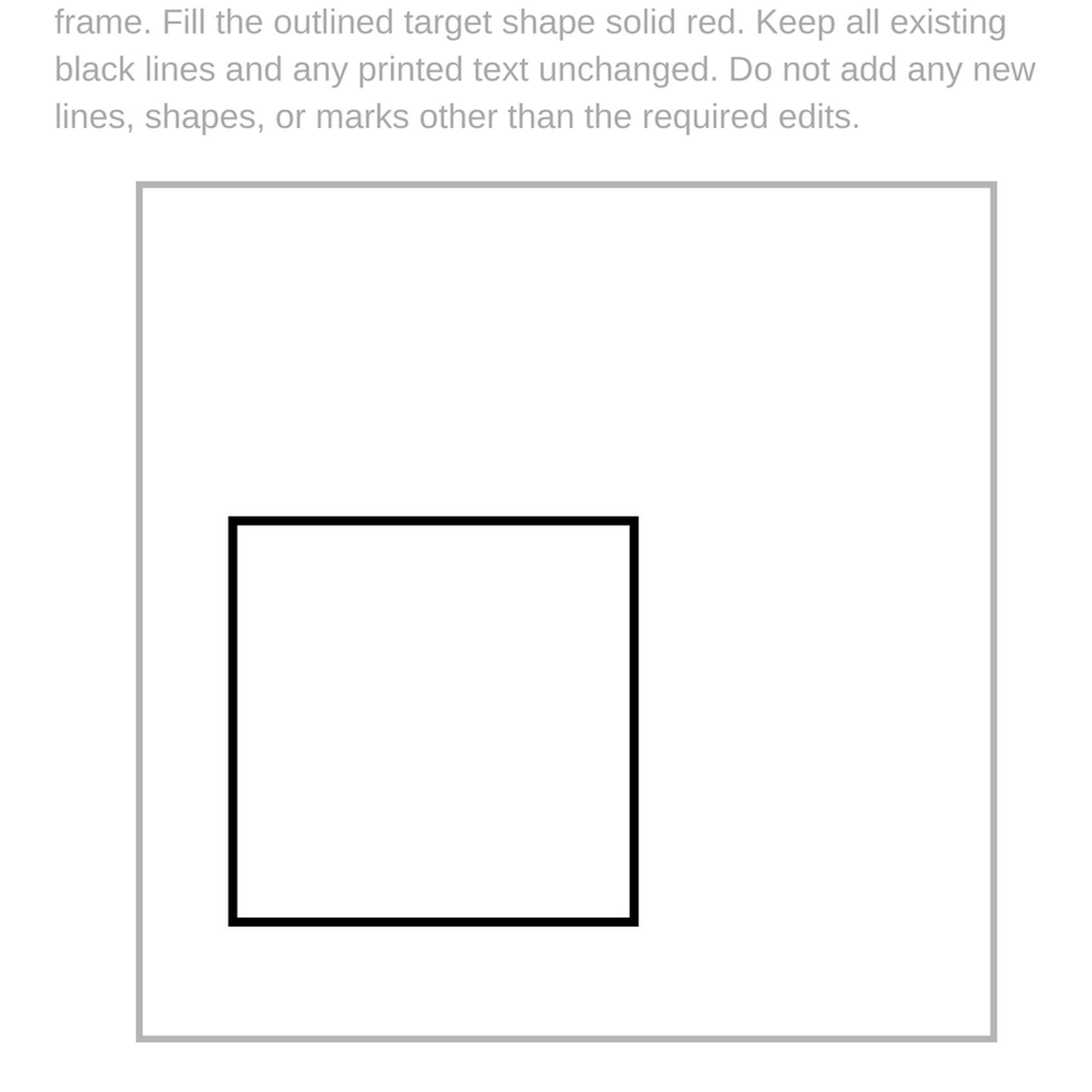} &
    \includegraphics[width=0.31\textwidth,height=0.173\textheight,keepaspectratio]{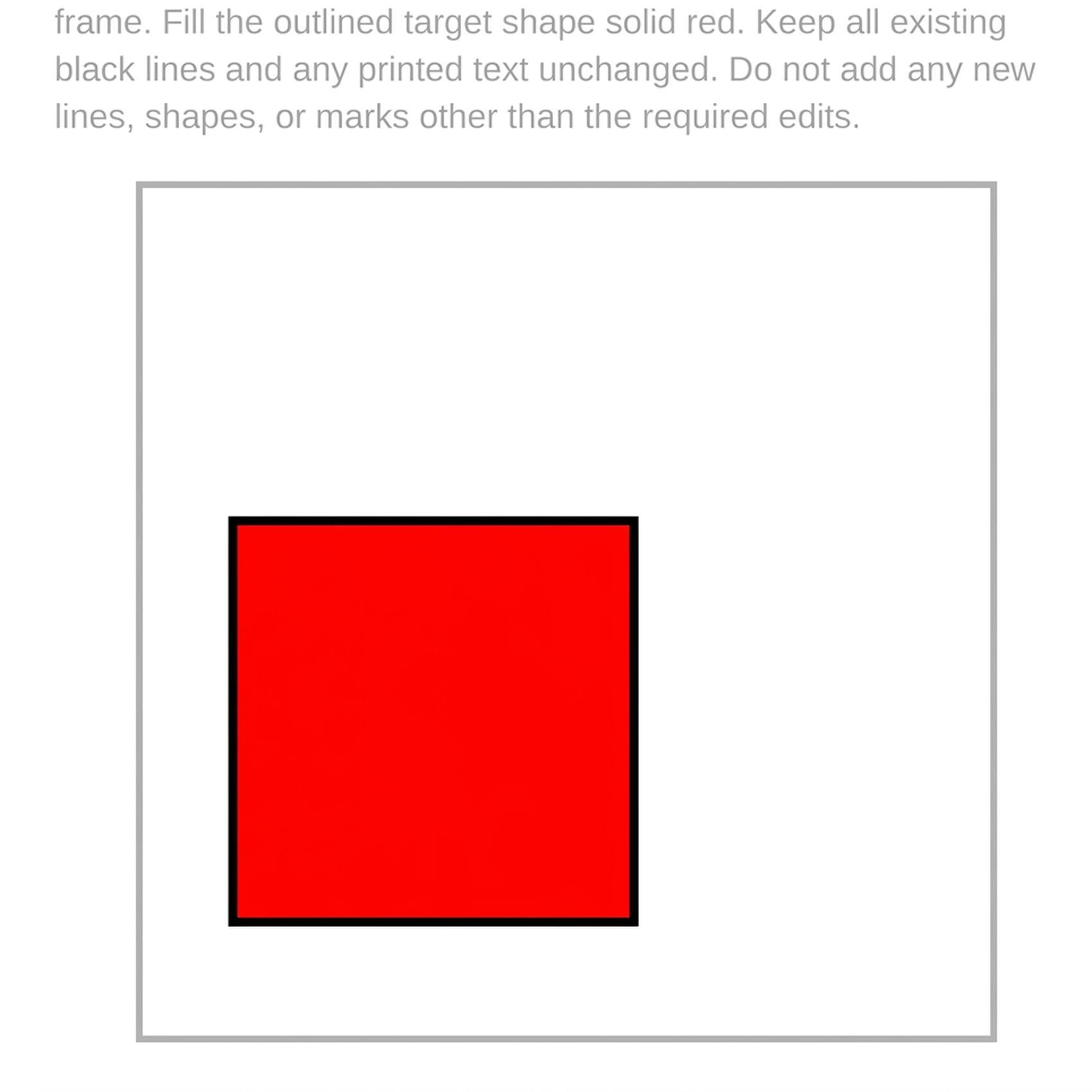} &
    \includegraphics[width=0.31\textwidth,height=0.173\textheight,keepaspectratio]{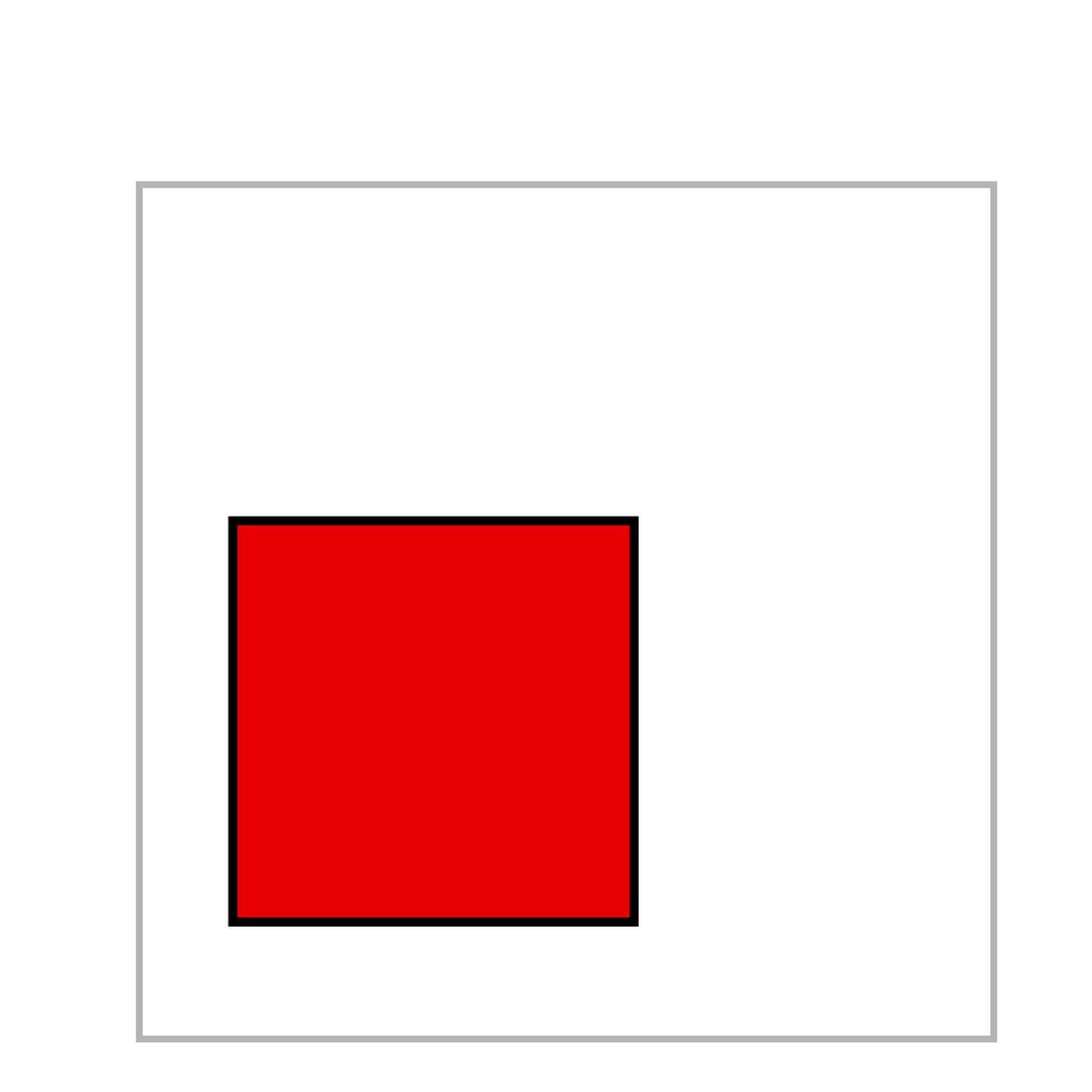} \\
    \multicolumn{3}{@{}L{0.96\textwidth}@{}}{\footnotesize\textbf{(a) Target-shape filling, V5.} The displayed output visually matches the target shape.}\\[0.90em]

    \textbf{Input} & \textbf{Nano Banana Pro output} & \textbf{Ground truth} \\
    \includegraphics[width=0.31\textwidth,height=0.173\textheight,keepaspectratio]{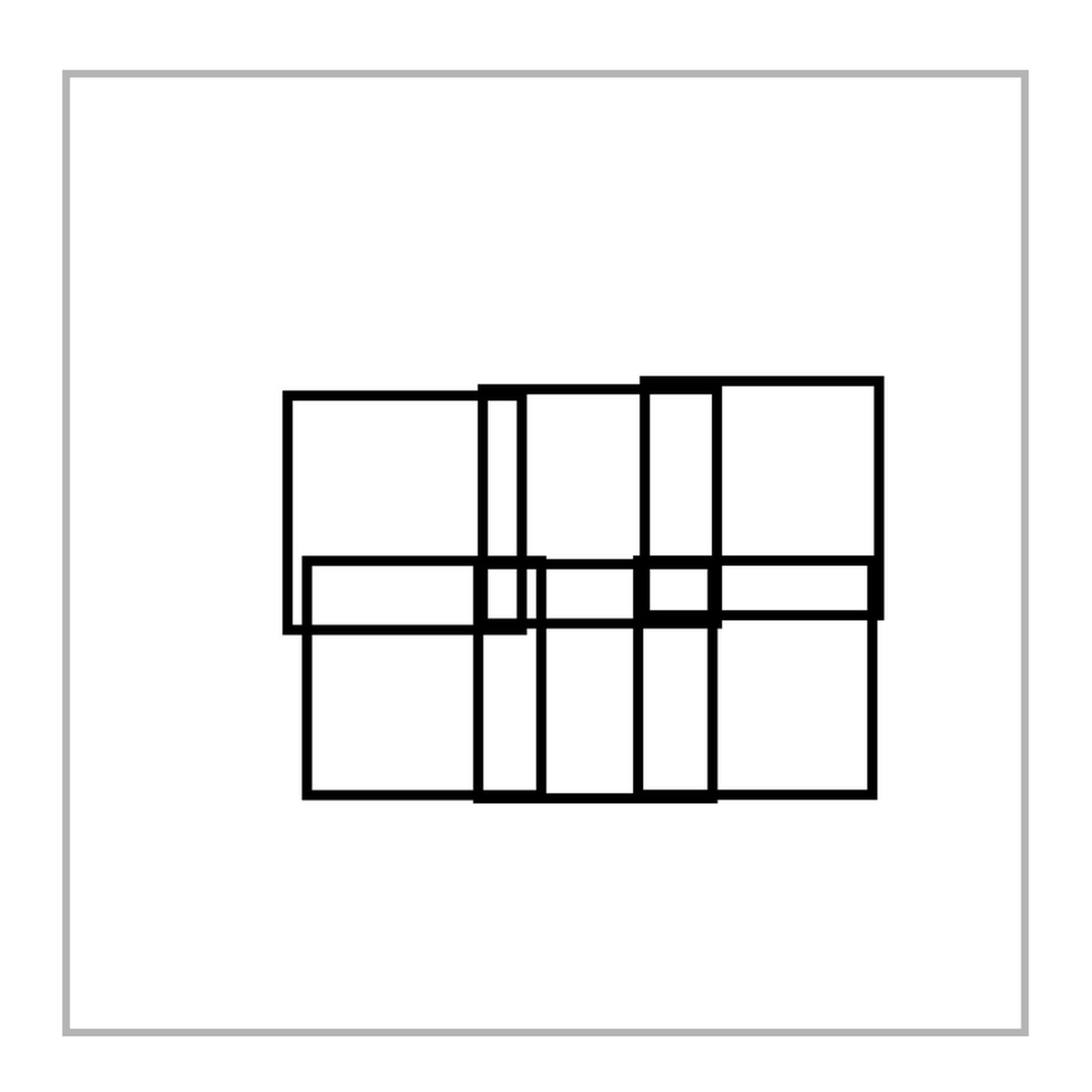} &
    \includegraphics[width=0.31\textwidth,height=0.173\textheight,keepaspectratio]{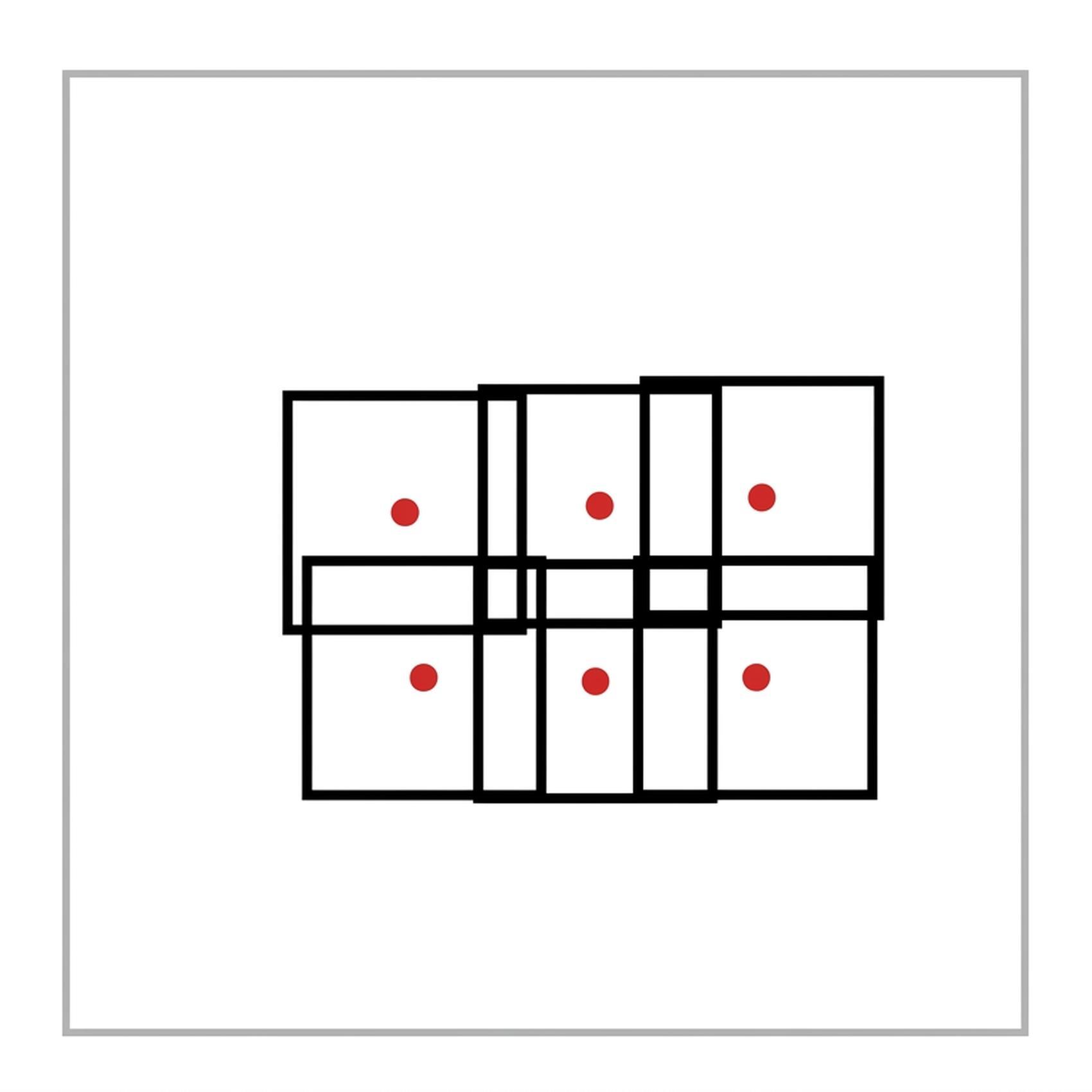} &
    \includegraphics[width=0.31\textwidth,height=0.173\textheight,keepaspectratio]{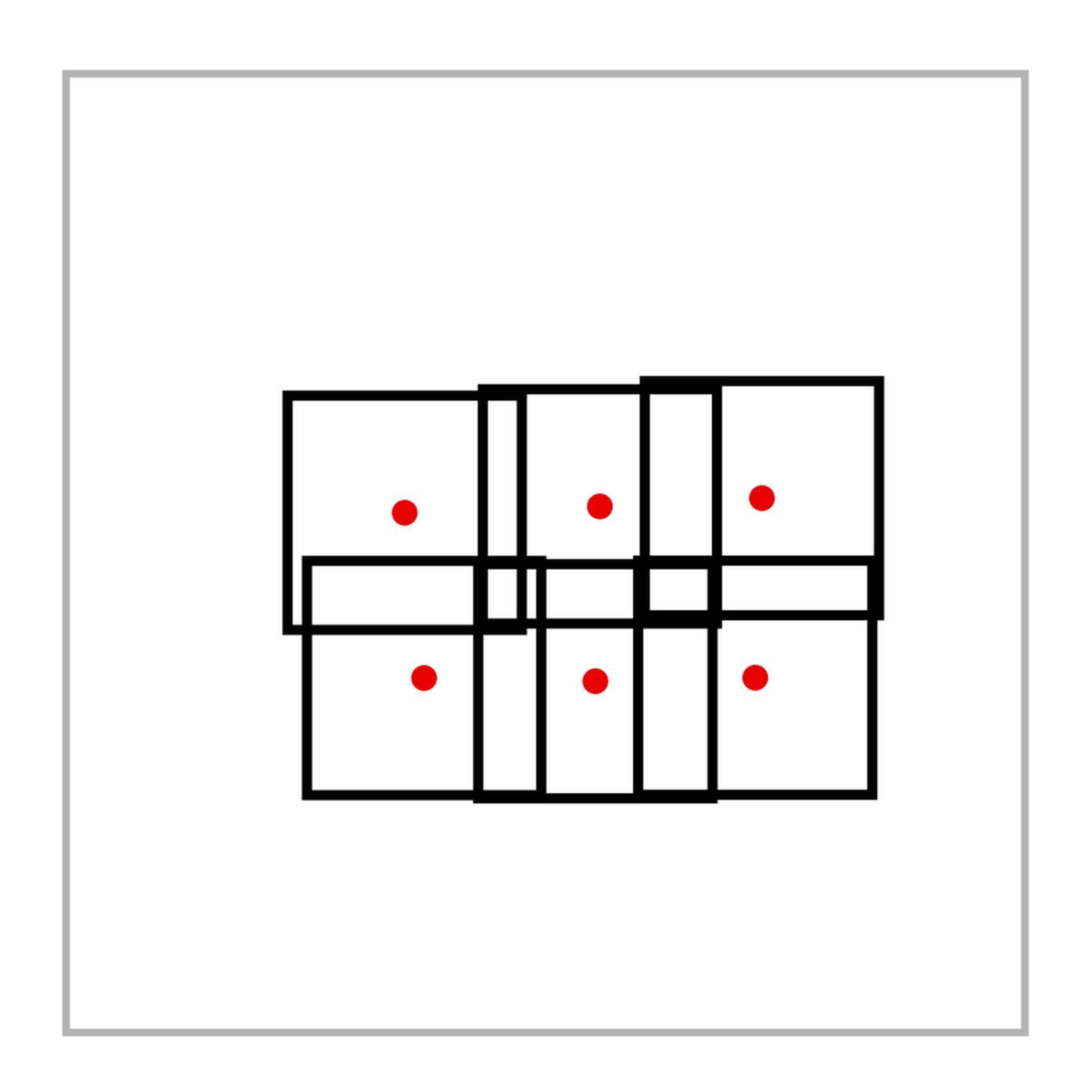} \\
    \multicolumn{3}{@{}L{0.96\textwidth}@{}}{\footnotesize\textbf{(b) Overlapping shapes, V4.} The displayed red marks align closely with the canonical target. The archived failure label is not used as evidence of a model error because target-matching inconsistencies were found in this evaluation.}\\[0.90em]

    \textbf{Input} & \textbf{Nano Banana Pro output} & \textbf{Ground truth} \\
    \includegraphics[width=0.31\textwidth,height=0.173\textheight,keepaspectratio]{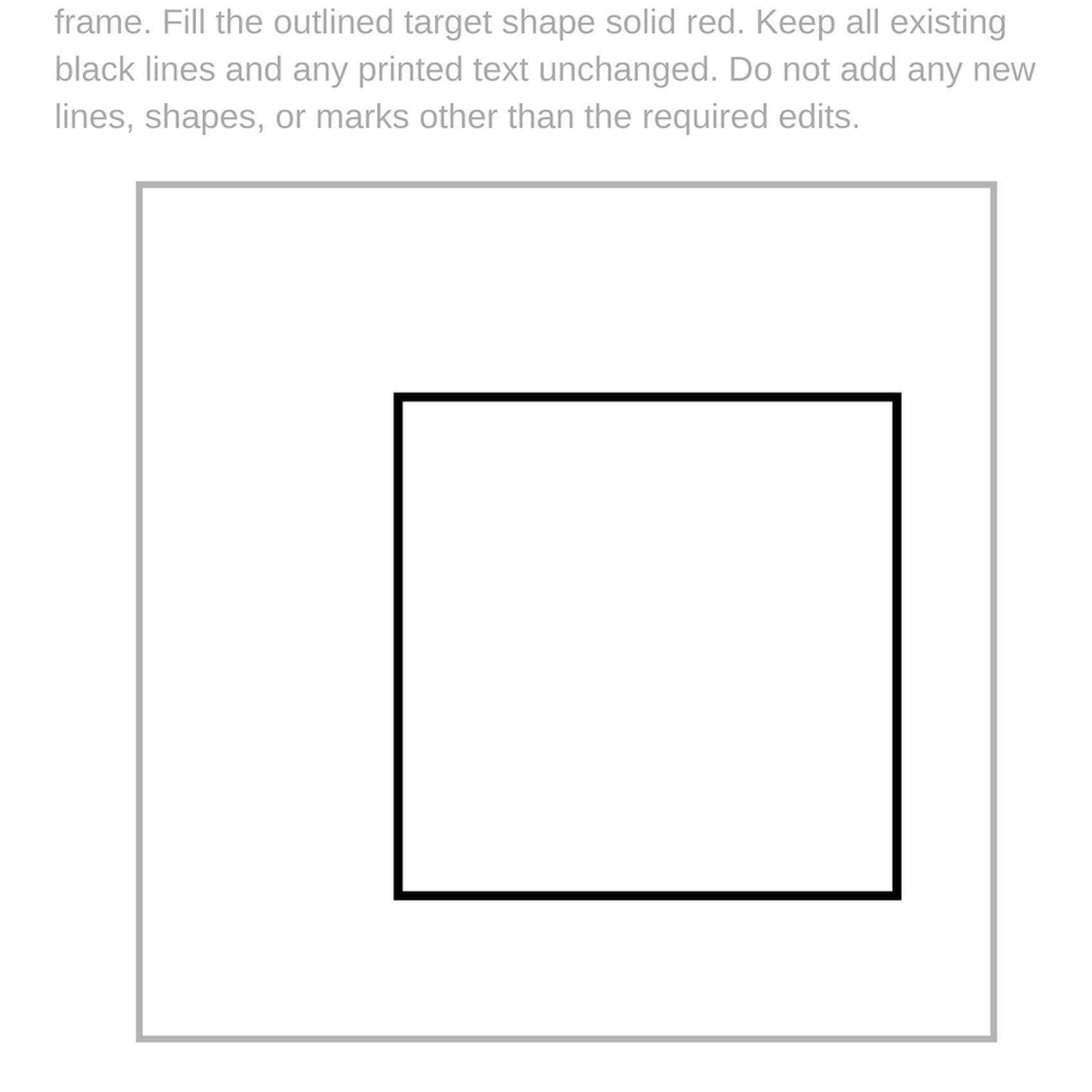} &
    \includegraphics[width=0.31\textwidth,height=0.173\textheight,keepaspectratio]{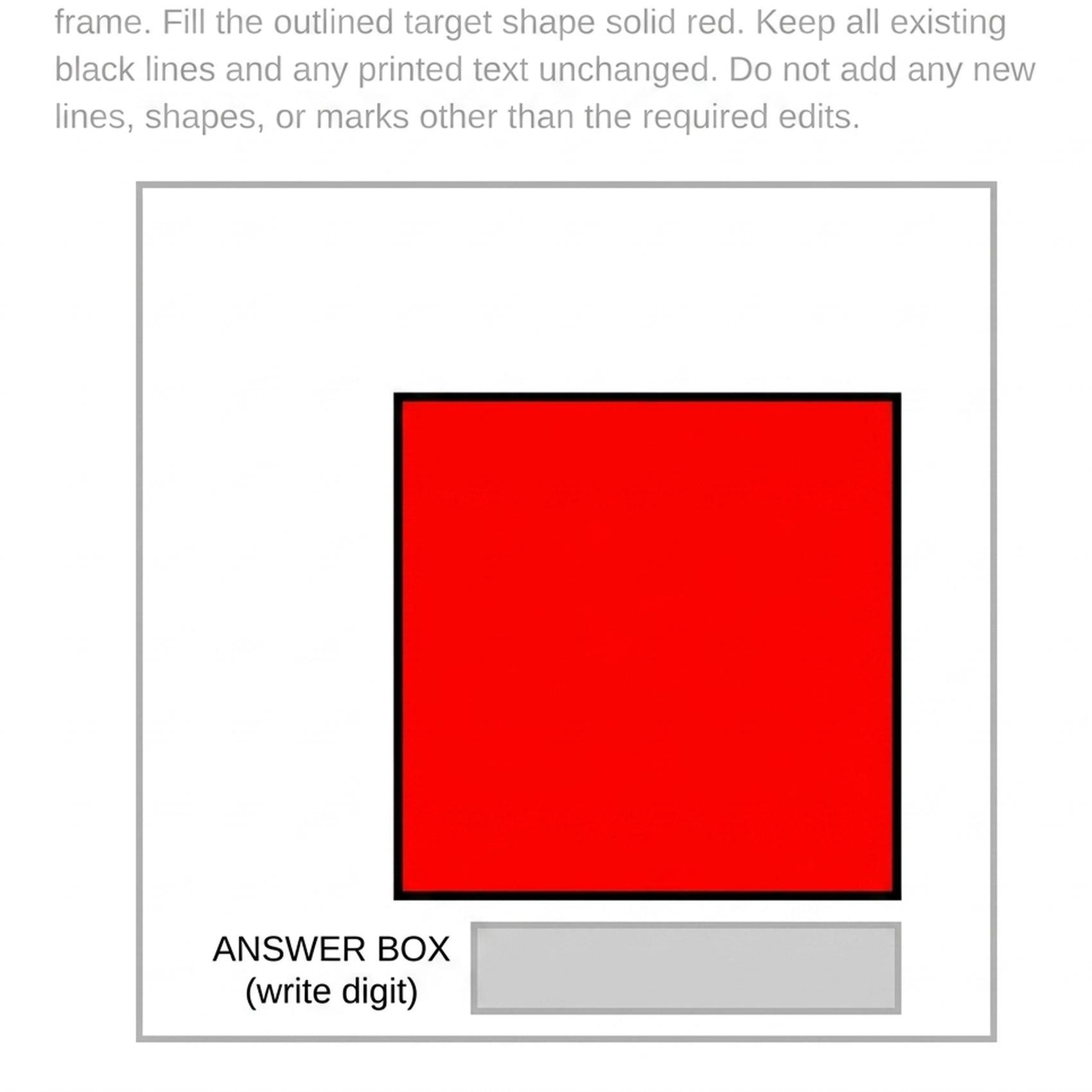} &
    \includegraphics[width=0.31\textwidth,height=0.173\textheight,keepaspectratio]{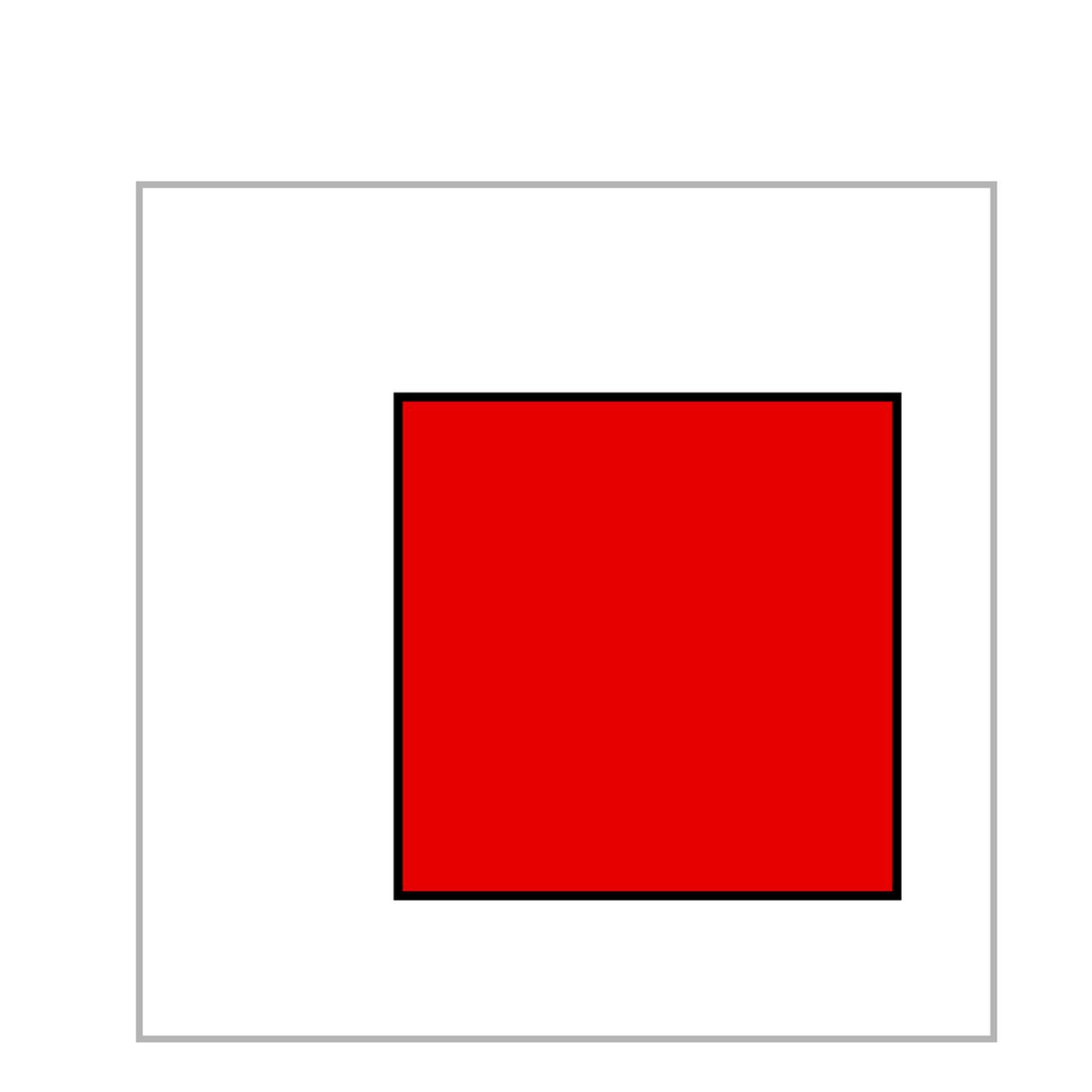} \\
    \multicolumn{3}{@{}L{0.96\textwidth}@{}}{\footnotesize\textbf{(c) Target-shape filling, CREF.} Nano Banana Pro is scored against the canonical worksheet target, not against the misleading reference cue.}\\[0.90em]

    \textbf{Input} & \textbf{FLUX.2 Klein 4B API output} & \textbf{Ground truth} \\
    \includegraphics[width=0.31\textwidth,height=0.173\textheight,keepaspectratio]{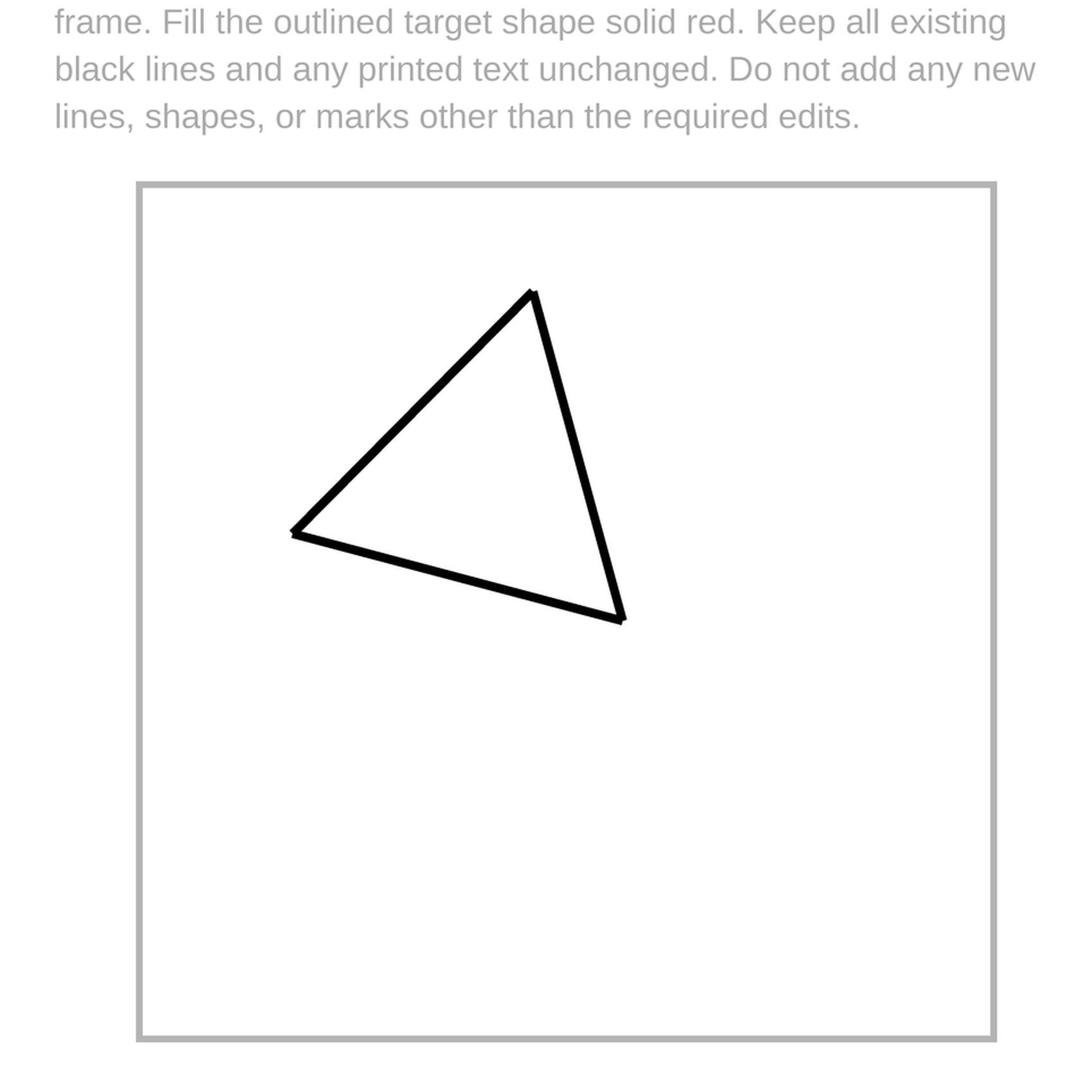} &
    \includegraphics[width=0.31\textwidth,height=0.173\textheight,keepaspectratio]{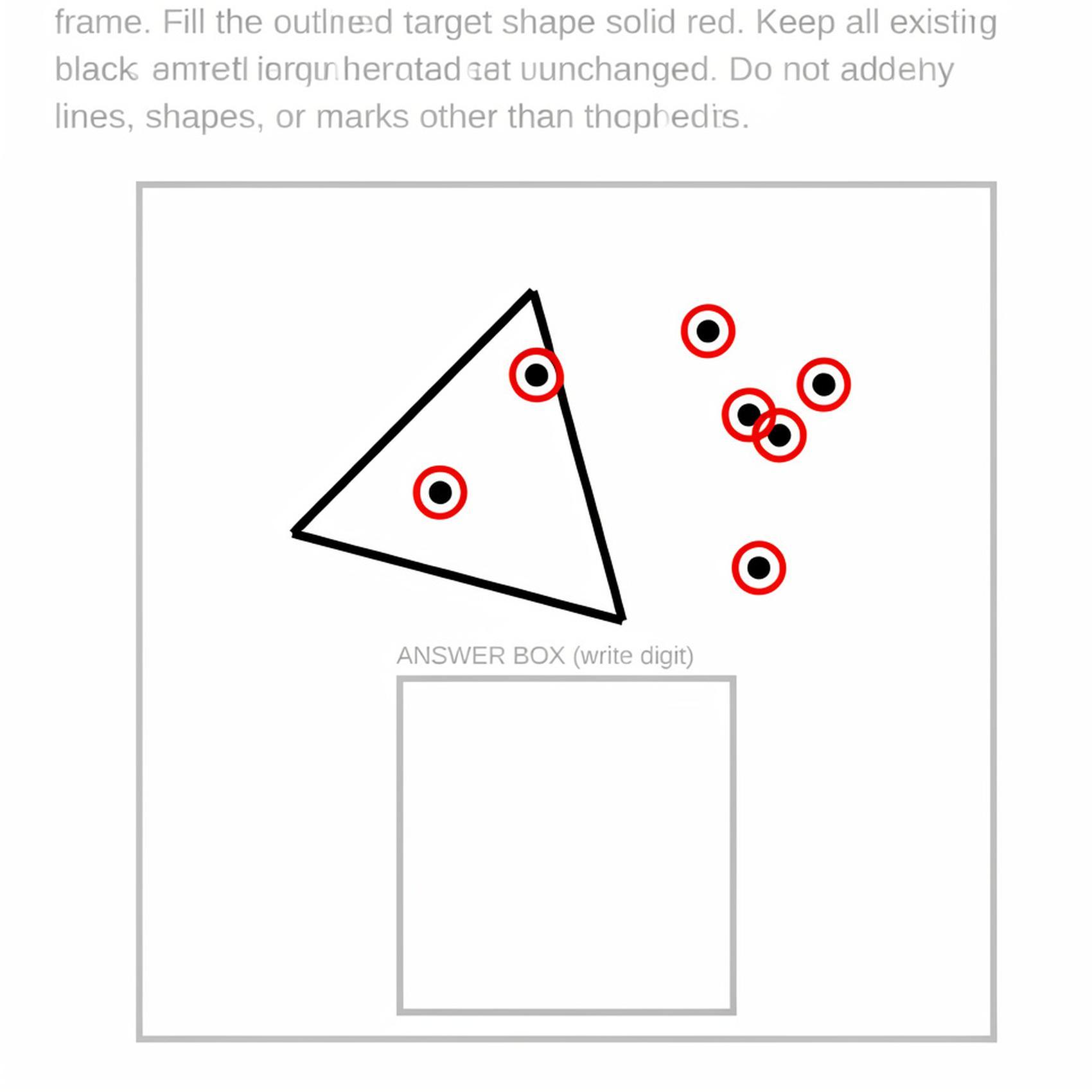} &
    \includegraphics[width=0.31\textwidth,height=0.173\textheight,keepaspectratio]{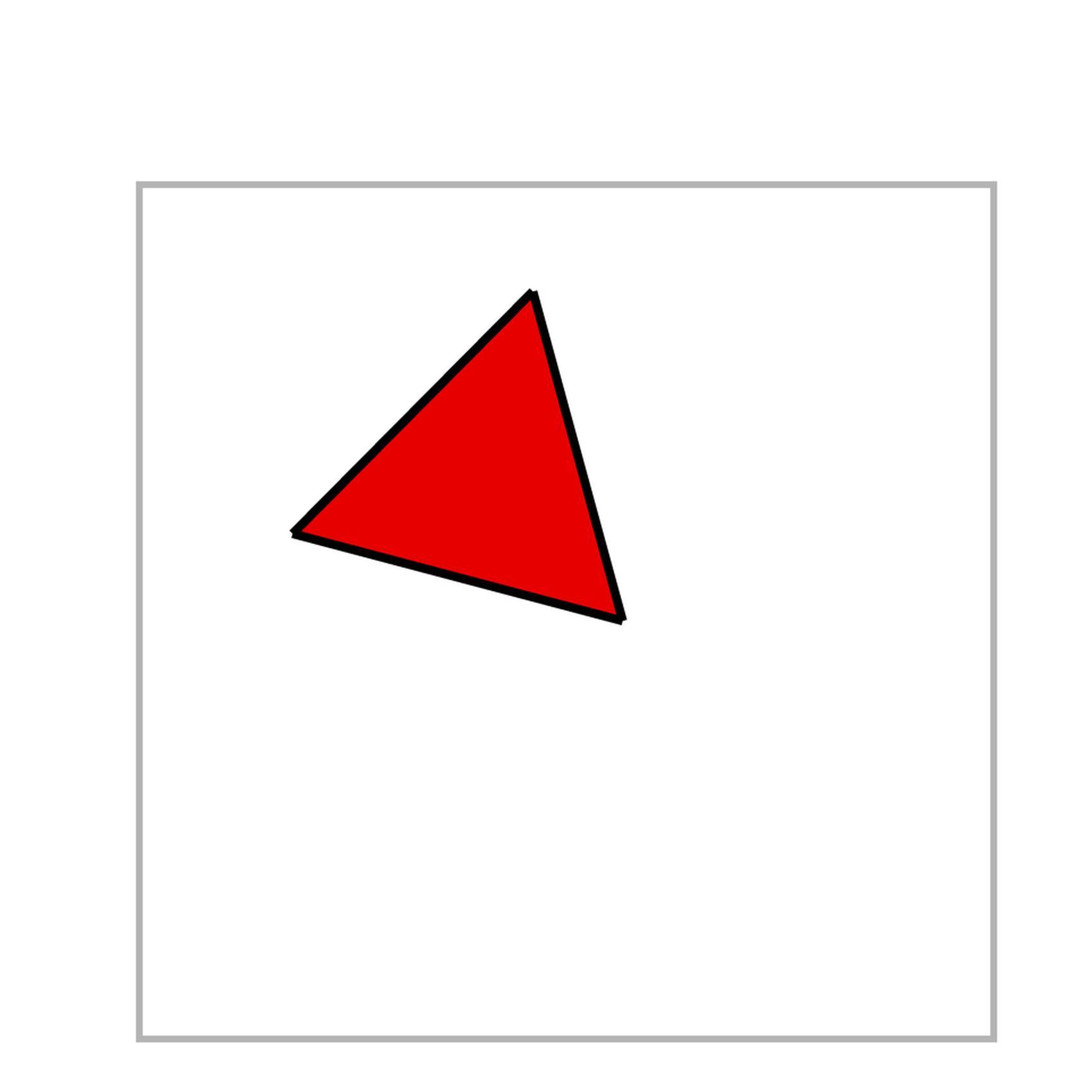} \\
    \multicolumn{3}{@{}L{0.96\textwidth}@{}}{\footnotesize\textbf{(d) Target-shape filling, CREF.} FLUX.2 Klein 4B API does not match the canonical worksheet target. This example alone does not establish which cue the model followed.}
  \end{tabular}%
  }

  \caption{Representative reference-condition and conflicting-reference diagnostic examples.
  The Input column shows the image sent to the model. The Ground truth column shows the
  canonical scoring target and intentionally omits the instruction header and reference context. Preservation is assessed against the input. These examples illustrate the collected outputs; they do not validate the archived additional-condition pass rates.}
  \label{fig:extra_examples}
\end{figure}

\section{Human Audit Protocol}
A single-rater audit was used to characterize recurring failure forms in a balanced \flux{} API subset. The completed audit contains 2,200 rateable outputs, covering 11 tasks, eight information conditions, and 25 paired base problems per task. The audit is used for failure analysis and proxy sanity checking, not for the main benchmark ranking. The recorded failures are distributed across multiple stages of the solve-and-edit pipeline rather than dominated by a single error type.

\par\begingroup

\captionsetup{hypcap=false,skip=4pt}
\setlength{\LTcapwidth}{\textwidth}
\begin{longtable}{lrrrrr}
\caption{Human audit aggregate over 2,200 sampled \flux{} API outputs.}\label{tab:audit-aggregate}\\
\toprule
Outputs & Strict correct & Strict (\%) & Loose correct & Loose (\%) & Strict failures \\

\midrule
\endfirsthead
\toprule
Outputs & Strict correct & Strict (\%) & Loose correct & Loose (\%) & Strict failures \\

\midrule
\endhead
\bottomrule
\endfoot
2,200 & 160 & 7.3 & 683 & 31.0 & 2,040 \\
\end{longtable}
\endgroup

The primary failure-code definitions are:
\begin{itemize}[leftmargin=1.2em,itemsep=1pt]
    \item \textbf{Missing action}: the required red edit is absent or nearly absent.
    \item \textbf{Black-line redrawing}: input geometry, text, frame, or answer boxes are redrawn, thickened, moved, deleted, shifted, or otherwise altered.
    \item \textbf{Wrong target}: the model edits the wrong object or wrong answer region.
    \item \textbf{No-edit violation}: a suppression/control item receives a red mark or other edit.
    \item \textbf{Wrong location}: the right kind of mark is placed, but not at the required point or region.
    \item \textbf{Wrong count}: in dot-counting-to-digit items, the model writes an incorrect numeral/count in the answer box.
    \item \textbf{Wrong glyph}: in letter-copying items, the model writes an incorrect letter or letter-like glyph in the answer box.
\end{itemize}

\par\begingroup
\fontsize{8}{10}\selectfont
\captionsetup{hypcap=false,skip=4pt}
\setlength{\LTcapwidth}{\textwidth}
\begin{longtable}{lrr}
\caption{Complete primary-failure counts for the 2,040 Audit-Strict failures. Percentages are rounded to one decimal; the single text-change case rounds to 0.0\%.}\label{tab:audit-failure-counts}\\
\toprule
Failure code & Count & Percent of strict failures \\

\midrule
\endfirsthead
\toprule
Failure code & Count & Percent of strict failures \\

\midrule
\endhead
\bottomrule
\endfoot
Missing action & 567 & 27.8 \\
Wrong target & 370 & 18.1 \\
Black-line redrawing & 363 & 17.8 \\
No-edit violation & 258 & 12.6 \\
Wrong location & 153 & 7.5 \\
Wrong glyph & 119 & 5.8 \\
Wrong count & 118 & 5.8 \\
Extra red marks & 33 & 1.6 \\
Black erased & 18 & 0.9 \\
Wrong region & 15 & 0.7 \\
Black shifted & 14 & 0.7 \\
Illegible glyph & 9 & 0.4 \\
Incomplete filling & 2 & 0.1 \\
Text changed & 1 & 0.0 \\
\end{longtable}
\endgroup

\par\begingroup
\fontsize{8}{10}\selectfont
\captionsetup{hypcap=false,skip=4pt}
\setlength{\LTcapwidth}{\textwidth}
\begin{longtable}{lrrrrr}
\caption{Single-rater Audit-Strict/Audit-Loose pass rates by task. Each task has 200 rateable outputs.}\label{tab:audit-by-task}\\
\toprule
Task & Outputs & Strict correct & Strict (\%) & Loose correct & Loose (\%) \\

\midrule
\endfirsthead
\toprule
Task & Outputs & Strict correct & Strict (\%) & Loose correct & Loose (\%) \\

\midrule
\endhead
\bottomrule
\endfoot
All-A control & 200 & 8 & 4.0 & 99 & 49.5 \\
Blank response & 200 & 51 & 25.5 & 195 & 97.5 \\
Circle every dot & 200 & 0 & 0.0 & 21 & 10.5 \\
Circled-letter copying & 200 & 0 & 0.0 & 7 & 3.5 \\
Dot counting to digit & 200 & 1 & 0.5 & 7 & 3.5 \\
Empty-circle response & 200 & 56 & 28.0 & 193 & 96.5 \\
Line intersections & 200 & 0 & 0.0 & 57 & 28.5 \\
Line midpoint & 200 & 9 & 4.5 & 19 & 9.5 \\
Overlapping shapes & 200 & 0 & 0.0 & 0 & 0.0 \\
Single-letter copying & 200 & 0 & 0.0 & 0 & 0.0 \\
Target-shape filling & 200 & 35 & 17.5 & 85 & 42.5 \\
\end{longtable}
\endgroup

\par\begingroup
\fontsize{8}{10}\selectfont
\captionsetup{hypcap=false,skip=4pt}
\setlength{\LTcapwidth}{\textwidth}
\begin{longtable}{lrrrrr}
\caption{Single-rater Audit-Strict/Audit-Loose pass rates by information condition. Each condition has 275 rateable outputs.}\label{tab:audit-by-condition}\\
\toprule
Condition & Outputs & Strict correct & Strict (\%) & Loose correct & Loose (\%) \\

\midrule
\endfirsthead
\toprule
Condition & Outputs & Strict correct & Strict (\%) & Loose correct & Loose (\%) \\

\midrule
\endhead
\bottomrule
\endfoot
V0 & 275 & 9 & 3.3 & 88 & 32.0 \\
V1 & 275 & 12 & 4.4 & 87 & 31.6 \\
V2 & 275 & 48 & 17.5 & 49 & 17.8 \\
V3 & 275 & 50 & 18.2 & 50 & 18.2 \\
V4 & 275 & 7 & 2.5 & 113 & 41.1 \\
V5 & 275 & 6 & 2.2 & 104 & 37.8 \\
V6 & 275 & 14 & 5.1 & 98 & 35.6 \\
V7 & 275 & 14 & 5.1 & 94 & 34.2 \\
\end{longtable}
\endgroup

\par\begingroup
\fontsize{8}{10}\selectfont
\captionsetup{hypcap=false,skip=4pt}
\setlength{\LTcapwidth}{\textwidth}
\begin{longtable}{lrrrrr}
\caption{Compact primary-failure breakdown by task, part 1. Columns group related displayed labels to keep the table readable. }\label{tab:audit-failure-by-task-a}\\
\toprule
Task & Failures & Missing & Wrong target & Black damage & No-edit viol. \\

\midrule
\endfirsthead
\toprule
Task & Failures & Missing & Wrong target & Black damage & No-edit viol. \\

\midrule
\endhead
\bottomrule
\endfoot
All-A control & 192 & 55 & 90 & 42 & 0 \\
Blank response & 149 & 0 & 0 & 14 & 135 \\
Circle every dot & 200 & 86 & 18 & 39 & 0 \\
Circled-letter copying & 200 & 56 & 59 & 14 & 0 \\
Dot counting to digit & 199 & 50 & 3 & 24 & 0 \\
Empty-circle response & 144 & 1 & 10 & 9 & 123 \\
Line intersections & 200 & 55 & 16 & 96 & 0 \\
Line midpoint & 191 & 51 & 22 & 86 & 0 \\
Overlapping shapes & 200 & 108 & 5 & 31 & 0 \\
Single-letter copying & 200 & 56 & 81 & 9 & 0 \\
Target-shape filling & 165 & 49 & 66 & 31 & 0 \\
\end{longtable}
\endgroup

\par\begingroup
\fontsize{8}{10}\selectfont
\captionsetup{hypcap=false,skip=4pt}
\setlength{\LTcapwidth}{\textwidth}
\begin{longtable}{lrrrrr}
\caption{Compact primary-failure breakdown by task, part 2. ``Other'' groups extra red marks, wrong-region labels, illegible glyphs, incomplete filling, and text changes; black erasures and black shifts are included in black damage in part 1.}\label{tab:audit-failure-by-task-b}\\
\toprule
Task & Failures & Wrong loc. & Wrong glyph & Wrong count & Other \\

\midrule
\endfirsthead
\toprule
Task & Failures & Wrong loc. & Wrong glyph & Wrong count & Other \\

\midrule
\endhead
\bottomrule
\endfoot
All-A control & 192 & 0 & 0 & 0 & 5 \\
Blank response & 149 & 0 & 0 & 0 & 0 \\
Circle every dot & 200 & 56 & 0 & 0 & 1 \\
Circled-letter copying & 200 & 0 & 68 & 0 & 3 \\
Dot counting to digit & 199 & 0 & 1 & 118 & 3 \\
Empty-circle response & 144 & 0 & 0 & 0 & 1 \\
Line intersections & 200 & 10 & 0 & 0 & 23 \\
Line midpoint & 191 & 30 & 0 & 0 & 2 \\
Overlapping shapes & 200 & 56 & 0 & 0 & 0 \\
Single-letter copying & 200 & 0 & 50 & 0 & 4 \\
Target-shape filling & 165 & 1 & 0 & 0 & 18 \\
\end{longtable}
\endgroup

\begin{figure}[H]
\centering
\includegraphics[width=0.9\linewidth]{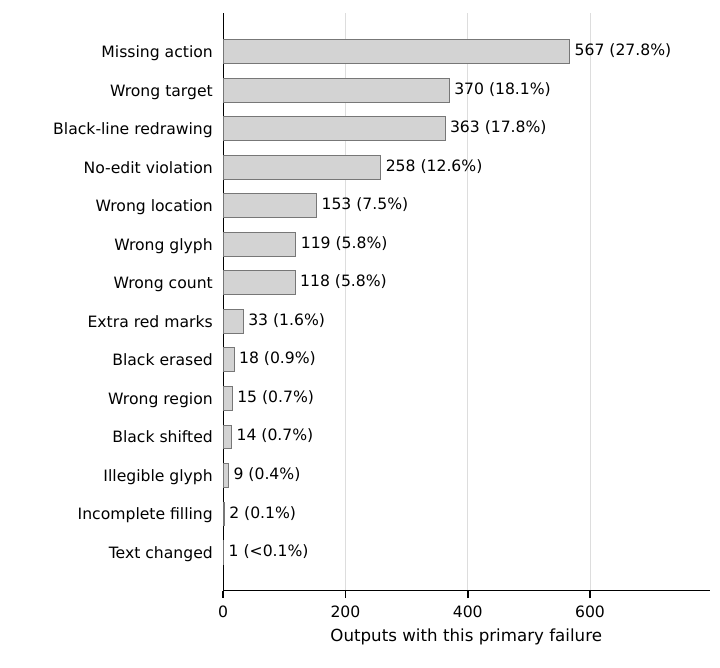}
\caption{Primary failure-code distribution in the single-rater audit of 2,200 FLUX.2 Klein 4B API outputs. Counts and percentages cover all 2,040 Audit-Strict failures; each output has one primary code.}
\label{fig:supp-human-audit-primary-failures}
\end{figure}

\subsection{Human-audit qualitative examples}

The following examples illustrate recurring audit labels; their frequencies are reported in the complete failure-count table above. The canonical ground-truth image may omit the input's instruction header.

\par\begingroup
\renewcommand{\LTcaptype}{figure}
\captionsetup{hypcap=false,skip=4pt}
\setlength{\tabcolsep}{3pt}
\setlength{\LTcapwidth}{\textwidth}
\begin{longtable}{@{}p{\textwidth}@{}}
\caption{Representative FLUX.2 Klein 4B API human-audit failure examples, part 1.}\label{fig:supp-flux-audit-gallery-part1}\\
\endfirsthead
\multicolumn{1}{@{}l@{}}{\small\bfseries Fig.~\thefigure\ (continued)}\\[6pt]
\endhead
\endfoot
\begin{minipage}{\linewidth}
\centering\fontsize{8}{10}\selectfont
\begin{tabular}{@{}ccc@{}}
\textbf{Input} & \textbf{FLUX.2 Klein 4B API output} & \textbf{Ground truth}\\
\includegraphics[width=0.31\textwidth,height=0.183\textheight,keepaspectratio]{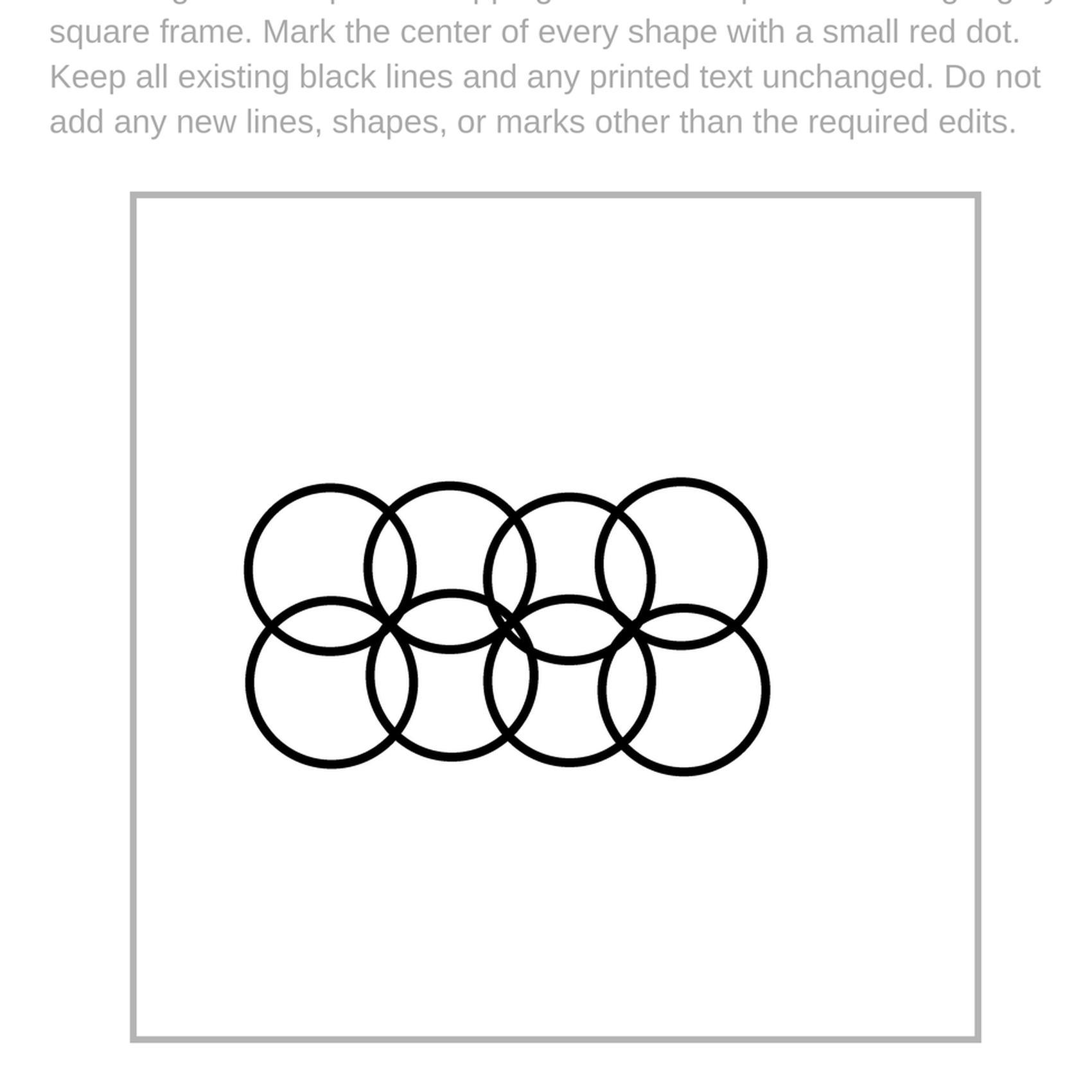} &
    \includegraphics[width=0.31\textwidth,height=0.183\textheight,keepaspectratio]{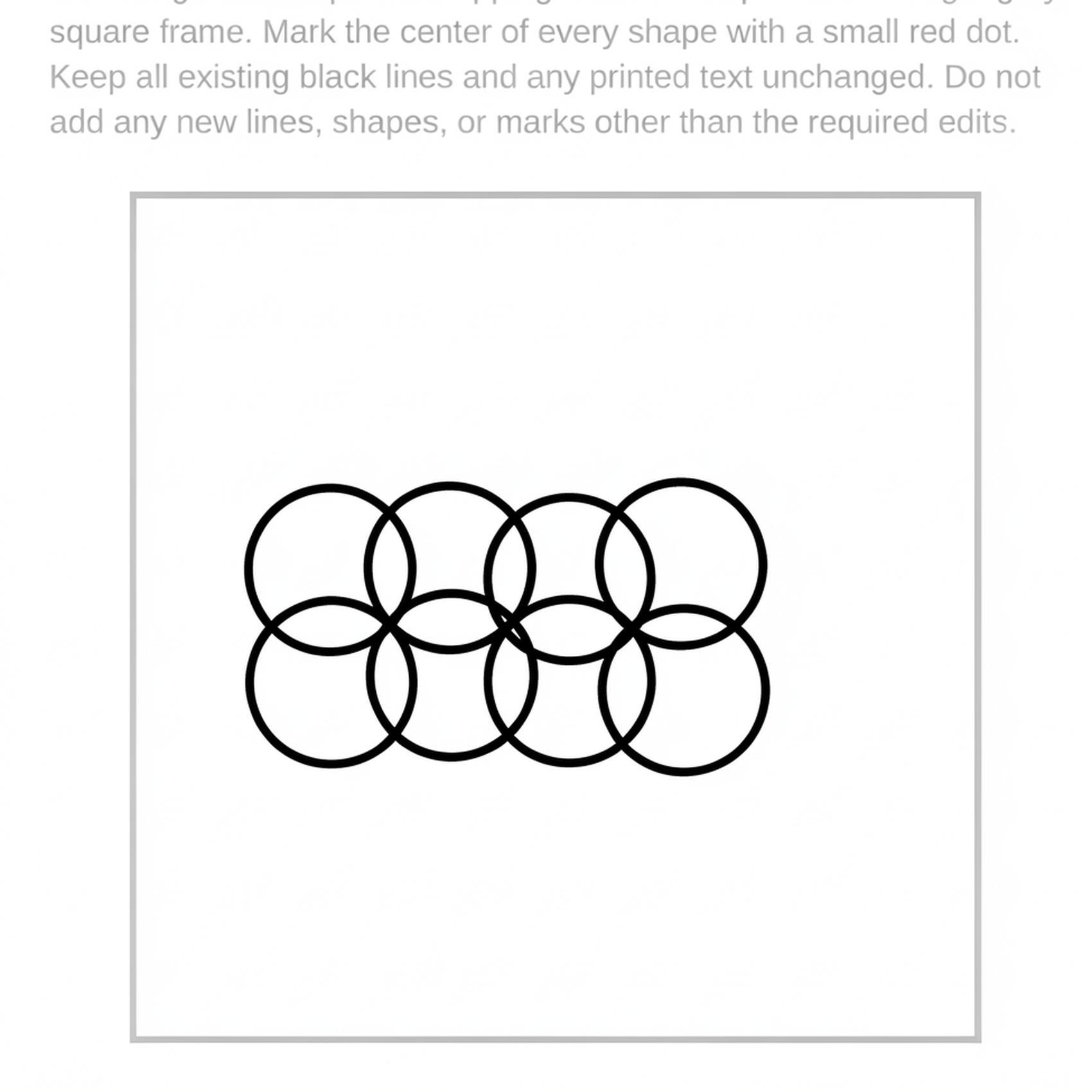} &
    \includegraphics[width=0.31\textwidth,height=0.183\textheight,keepaspectratio]{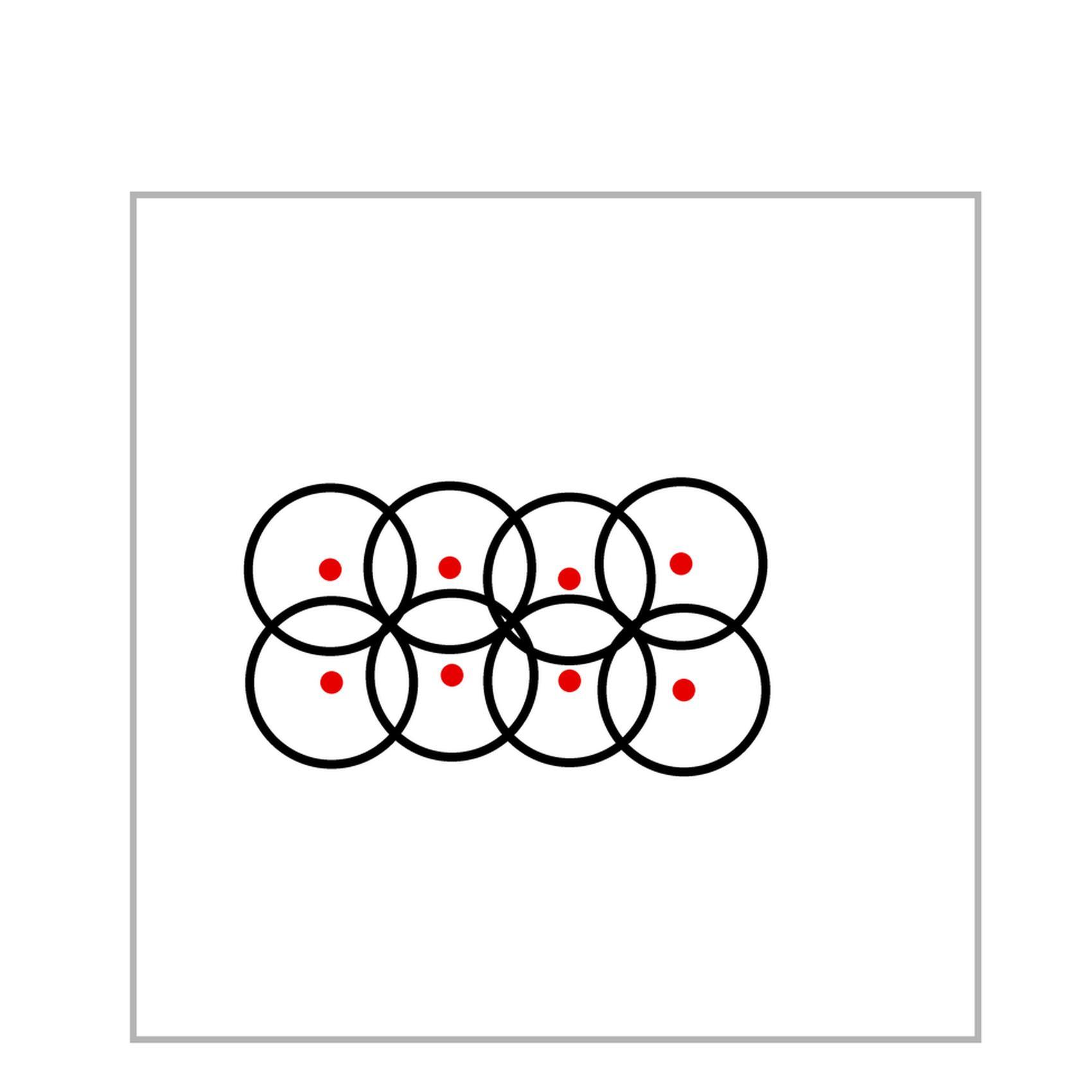}
\end{tabular}
\par\smallskip\raggedright \footnotesize\textbf{(a) Missing action.} The required red edit is absent or nearly absent.
\end{minipage}\\[7pt]
\begin{minipage}{\linewidth}
\centering\fontsize{8}{10}\selectfont
\begin{tabular}{@{}ccc@{}}
\textbf{Input} & \textbf{FLUX.2 Klein 4B API output} & \textbf{Ground truth}\\
\includegraphics[width=0.31\textwidth,height=0.183\textheight,keepaspectratio]{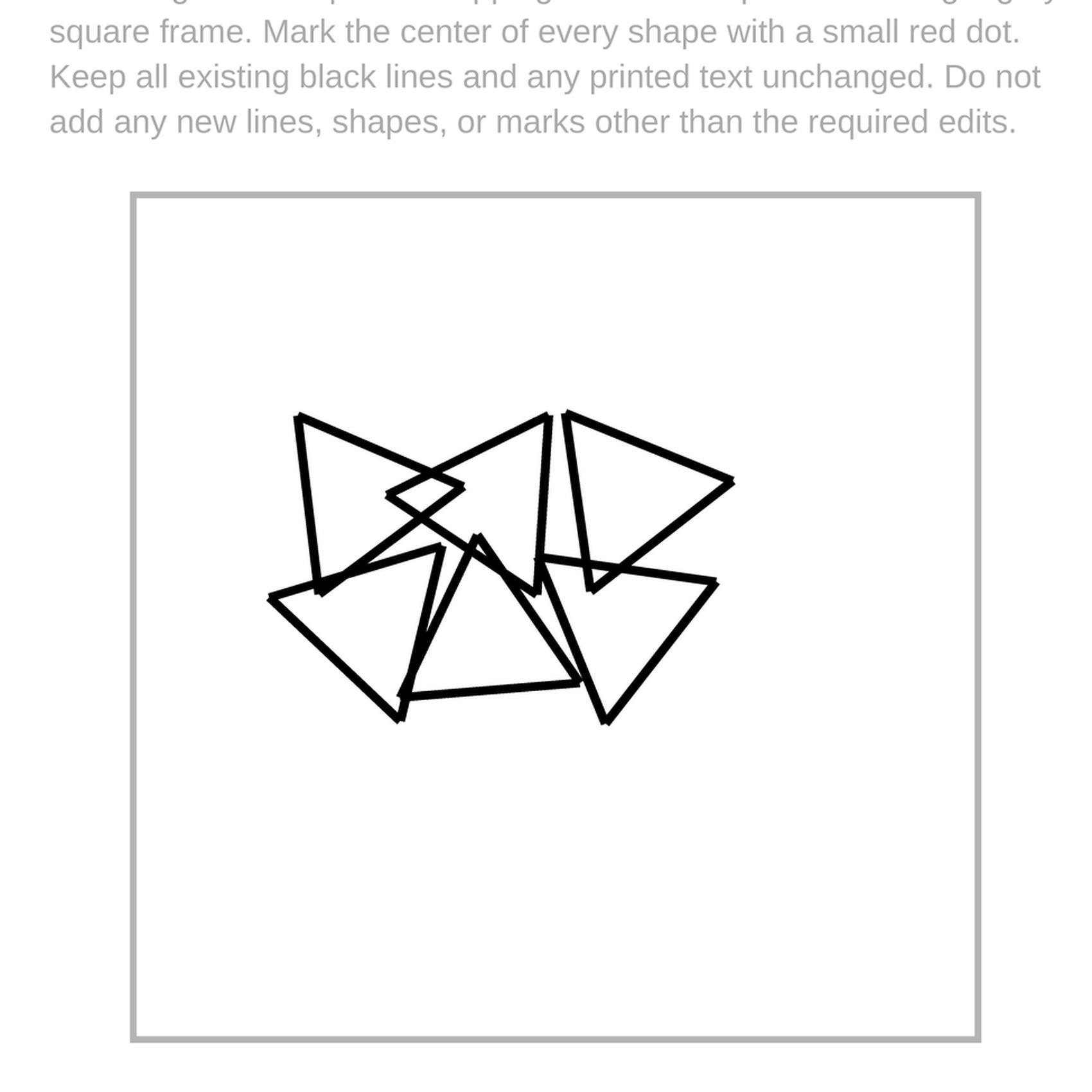} &
    \includegraphics[width=0.31\textwidth,height=0.183\textheight,keepaspectratio]{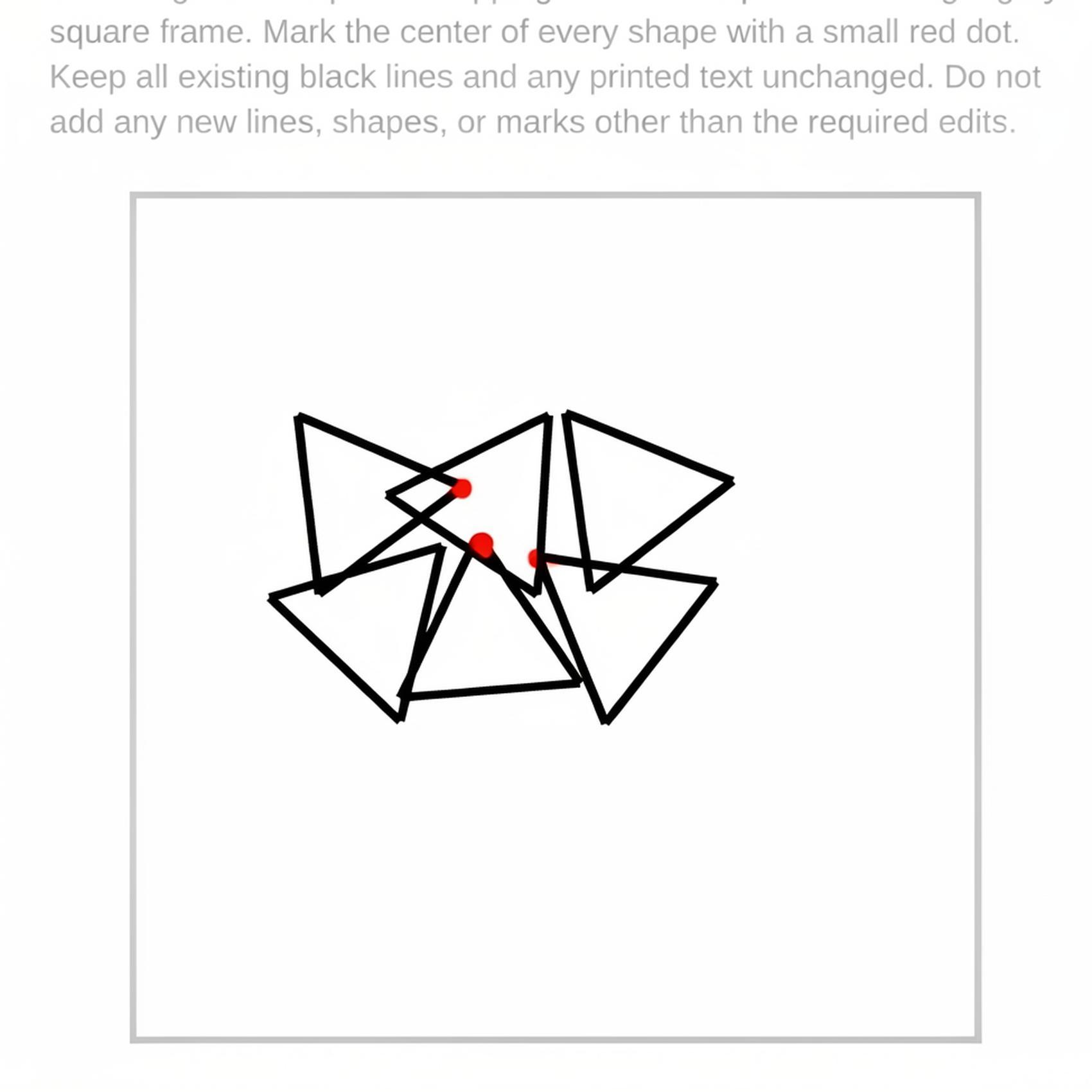} &
    \includegraphics[width=0.31\textwidth,height=0.183\textheight,keepaspectratio]{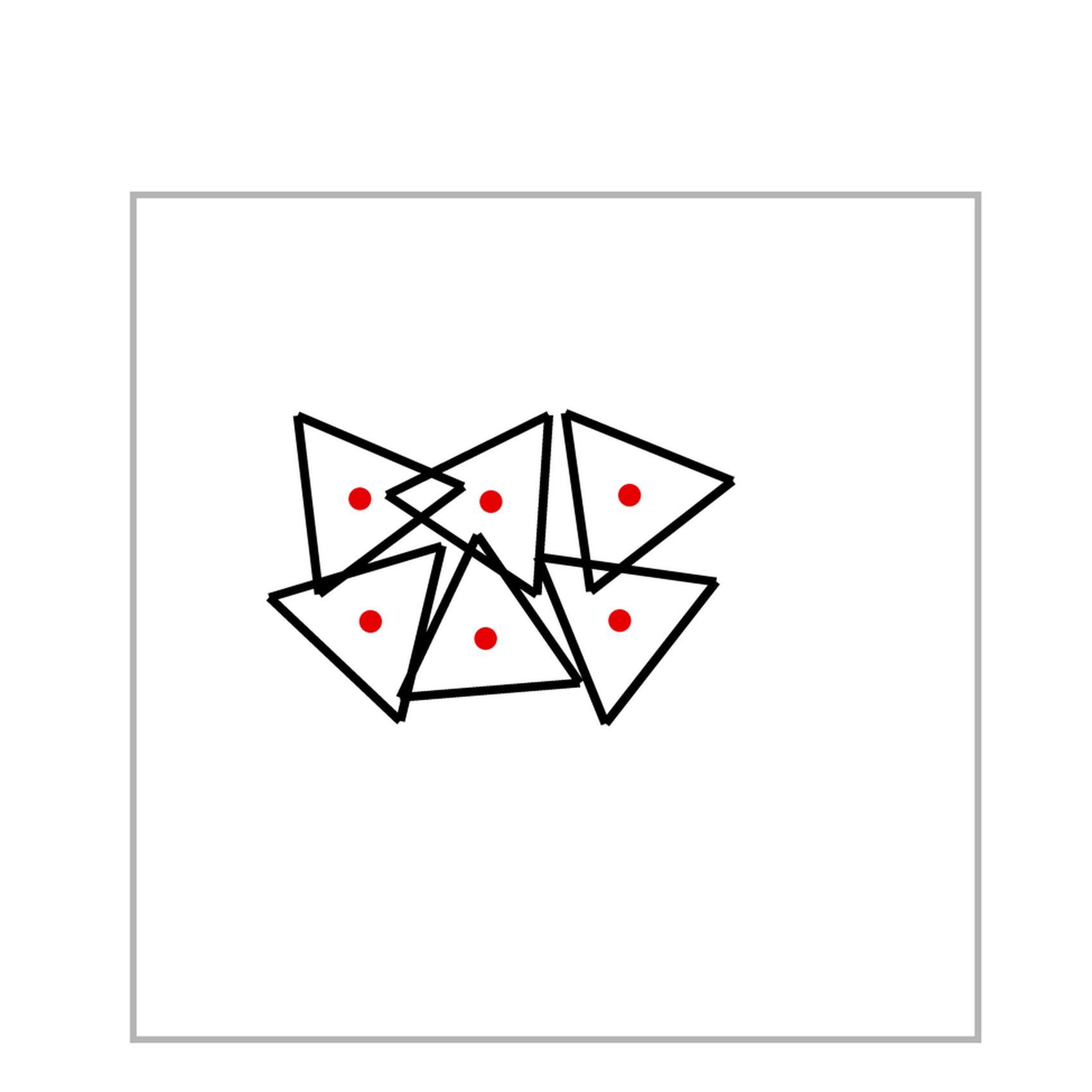}
\end{tabular}
\par\smallskip\raggedright \footnotesize\textbf{(b) Wrong target.} The model edits the wrong object or answer region.
\end{minipage}\\[7pt]
\begin{minipage}{\linewidth}
\centering\fontsize{8}{10}\selectfont
\begin{tabular}{@{}ccc@{}}
\textbf{Input} & \textbf{FLUX.2 Klein 4B API output} & \textbf{Ground truth}\\
\includegraphics[width=0.31\textwidth,height=0.183\textheight,keepaspectratio]{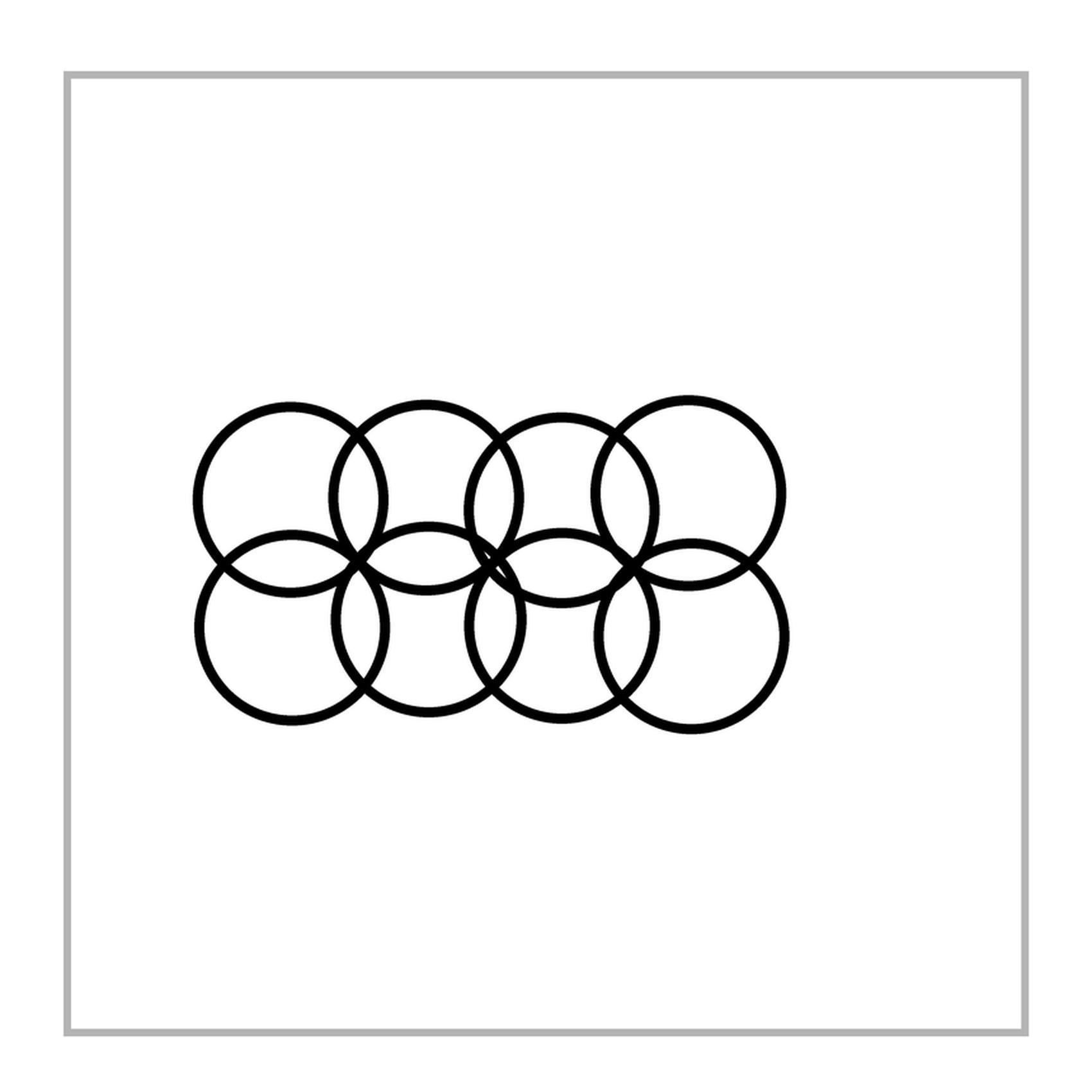} &
    \includegraphics[width=0.31\textwidth,height=0.183\textheight,keepaspectratio]{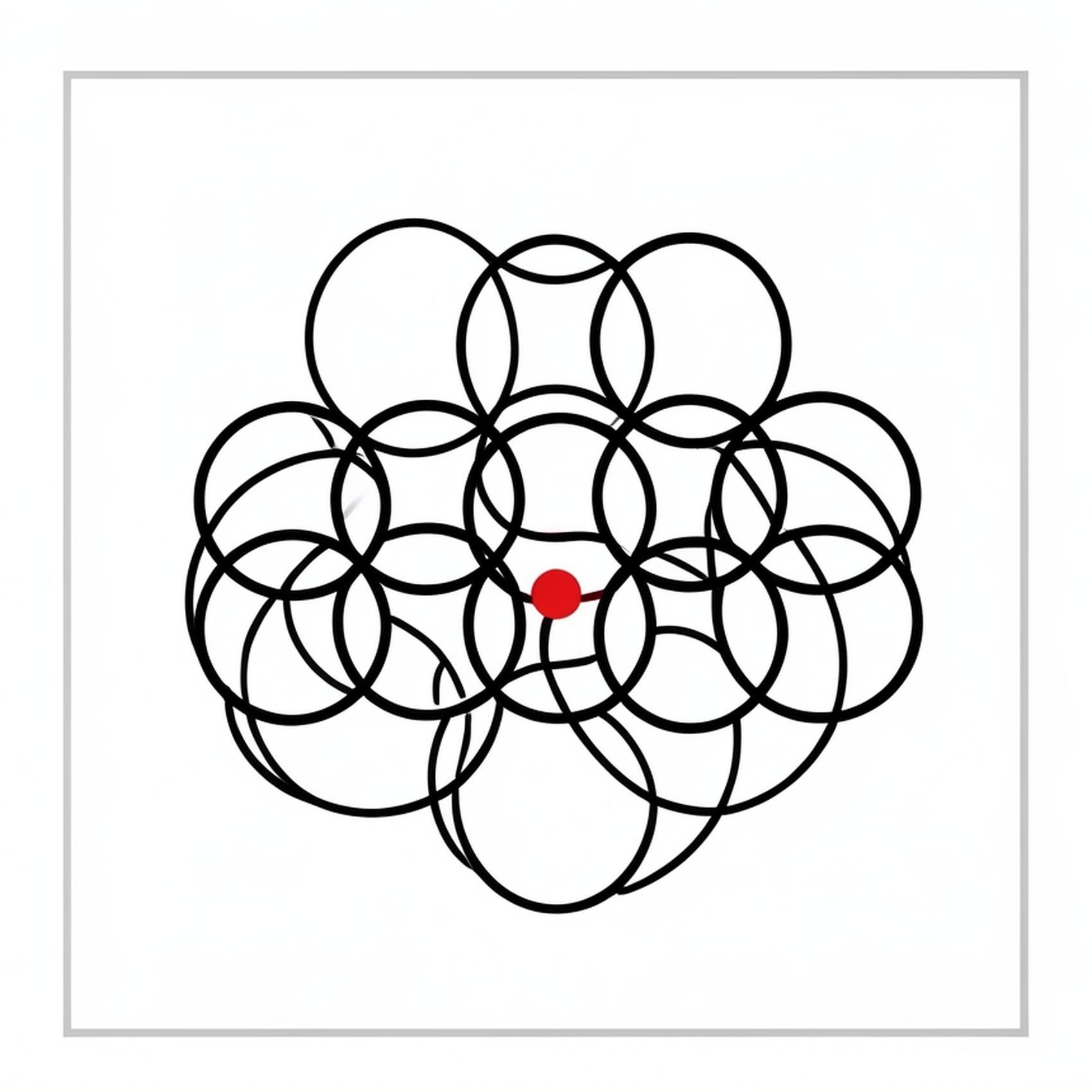} &
    \includegraphics[width=0.31\textwidth,height=0.183\textheight,keepaspectratio]{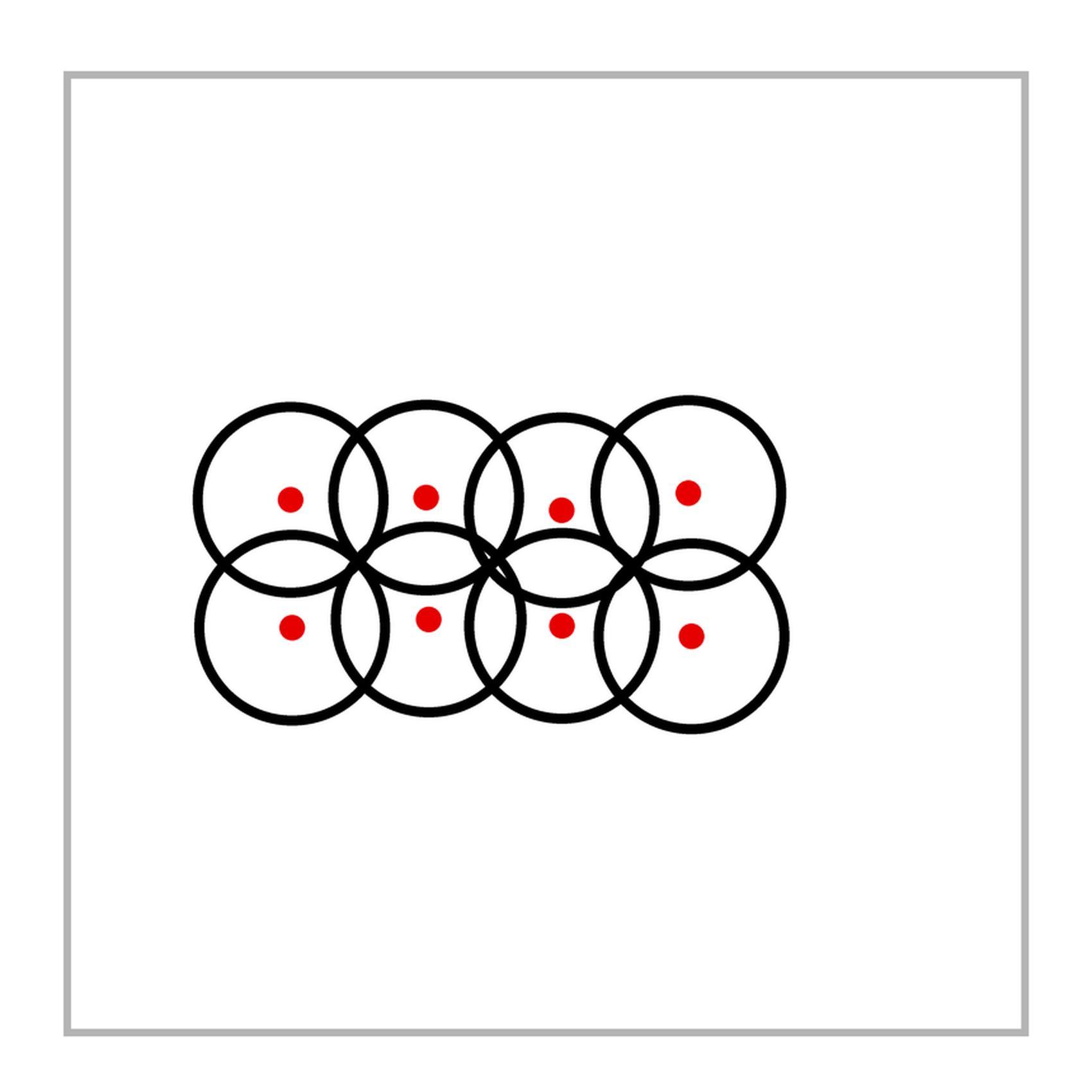}
\end{tabular}
\par\smallskip\raggedright \footnotesize\textbf{(c) Black-line redrawing.} Input geometry or worksheet structure is redrawn, moved, or damaged.
\end{minipage}\\[7pt]
\begin{minipage}{\linewidth}
\centering\fontsize{8}{10}\selectfont
\begin{tabular}{@{}ccc@{}}
\textbf{Input} & \textbf{FLUX.2 Klein 4B API output} & \textbf{Ground truth}\\
\includegraphics[width=0.31\textwidth,height=0.183\textheight,keepaspectratio]{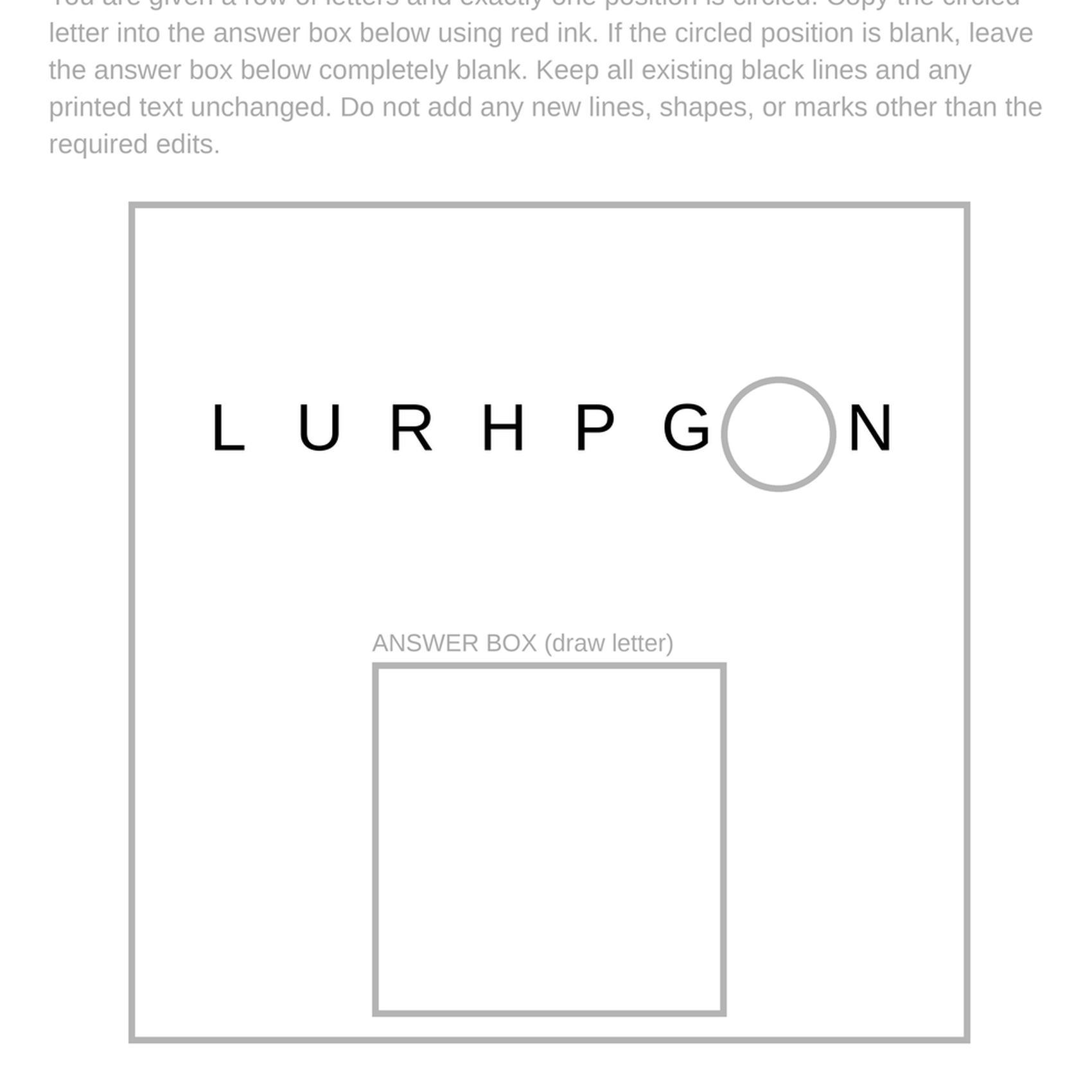} &
    \includegraphics[width=0.31\textwidth,height=0.183\textheight,keepaspectratio]{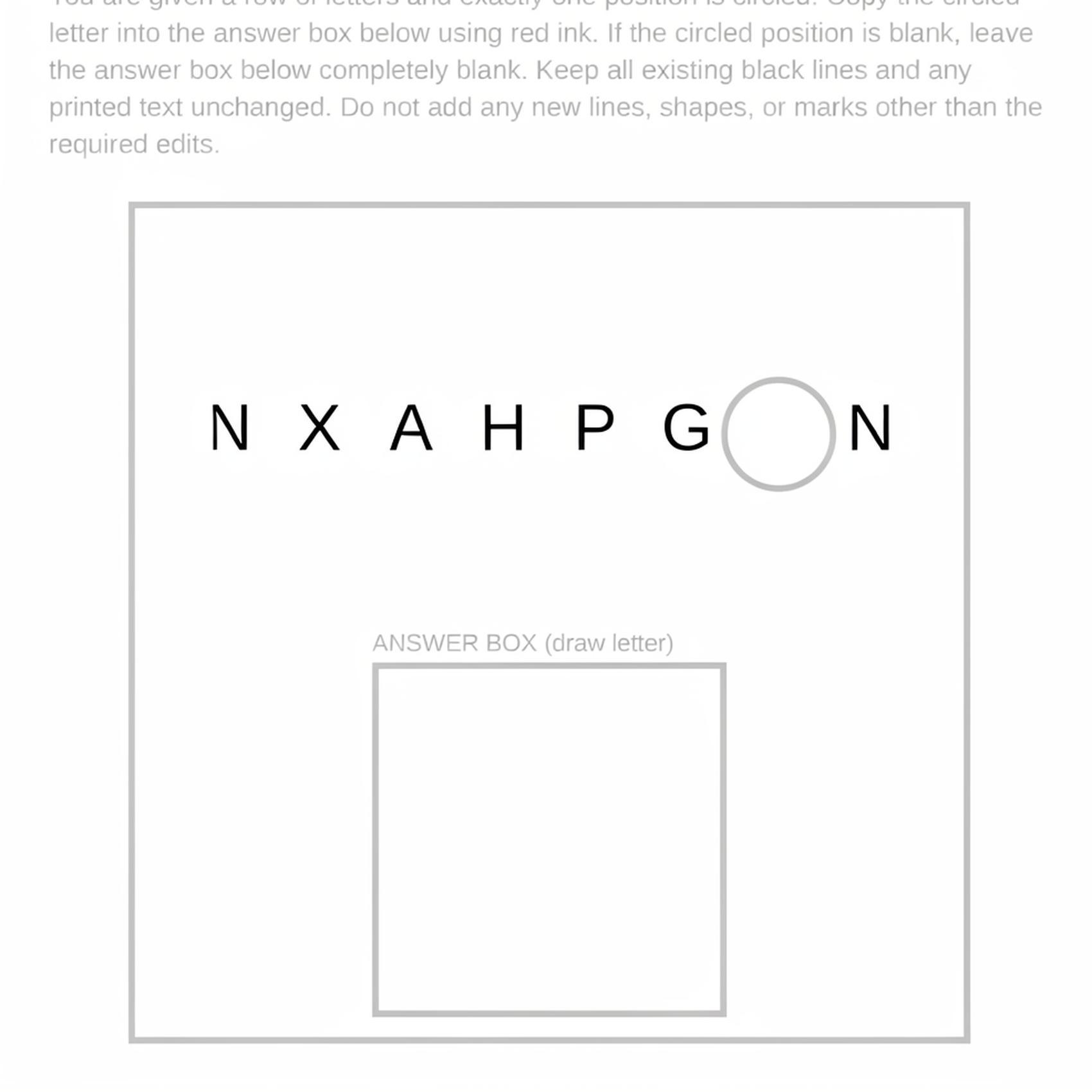} &
    \includegraphics[width=0.31\textwidth,height=0.183\textheight,keepaspectratio]{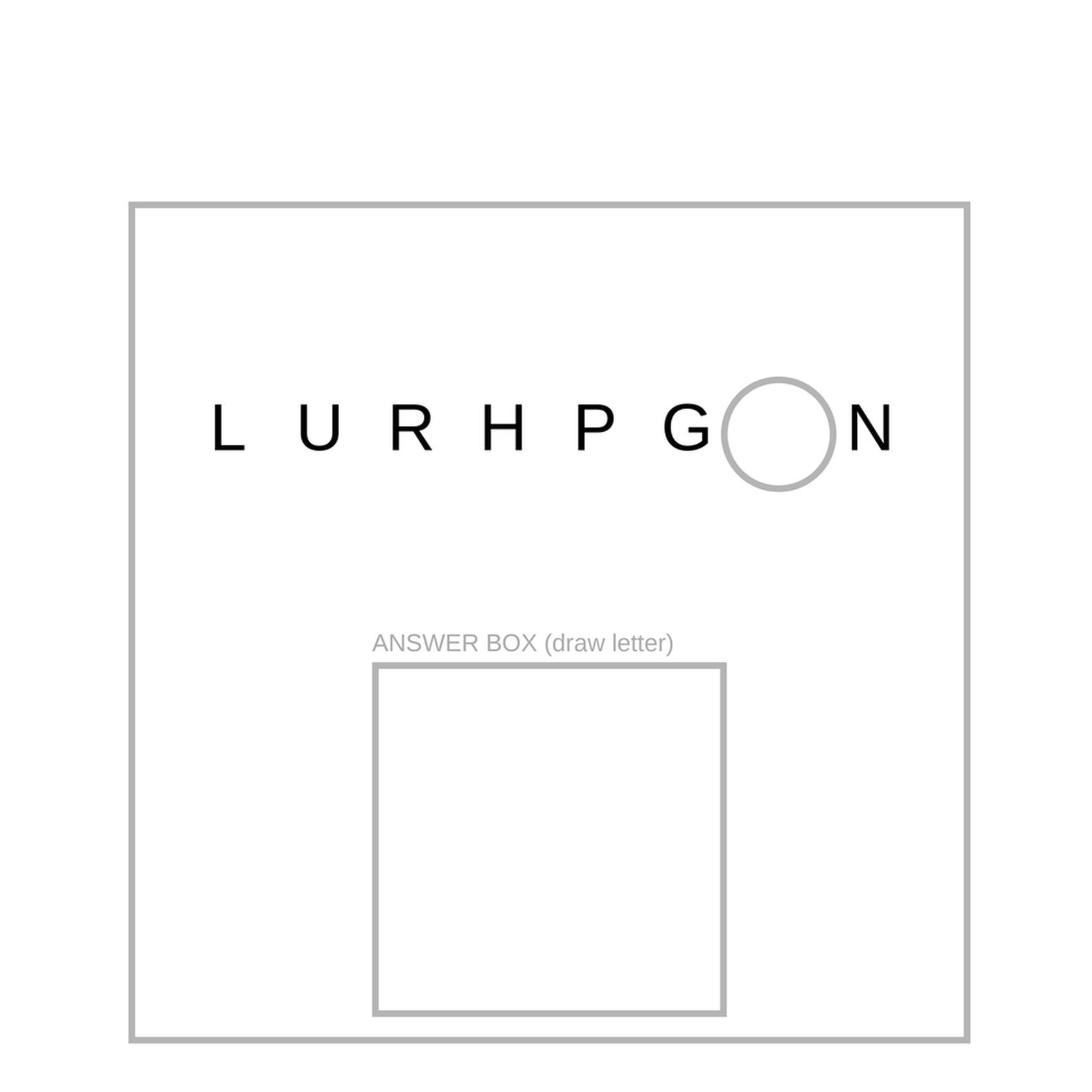}
\end{tabular}
\par\smallskip\raggedright \footnotesize\textbf{(d) No-edit violation.} A suppression/control item receives a red mark or other edit.
\end{minipage}\\[7pt]
\end{longtable}
\endgroup

\par\begingroup
\renewcommand{\LTcaptype}{figure}
\captionsetup{hypcap=false,skip=4pt}
\setlength{\tabcolsep}{3pt}
\setlength{\LTcapwidth}{\textwidth}
\begin{longtable}{@{}p{\textwidth}@{}}
\caption{Representative FLUX.2 Klein 4B API human-audit failure examples, part 2.}\label{fig:supp-flux-audit-gallery-part2}\\
\endfirsthead
\multicolumn{1}{@{}l@{}}{\small\bfseries Fig.~\thefigure\ (continued)}\\[6pt]
\endhead
\endfoot
\begin{minipage}{\linewidth}
\centering\fontsize{8}{10}\selectfont
\begin{tabular}{@{}ccc@{}}
\textbf{Input} & \textbf{FLUX.2 Klein 4B API output} & \textbf{Ground truth}\\
\includegraphics[width=0.32\textwidth,height=0.205\textheight,keepaspectratio]{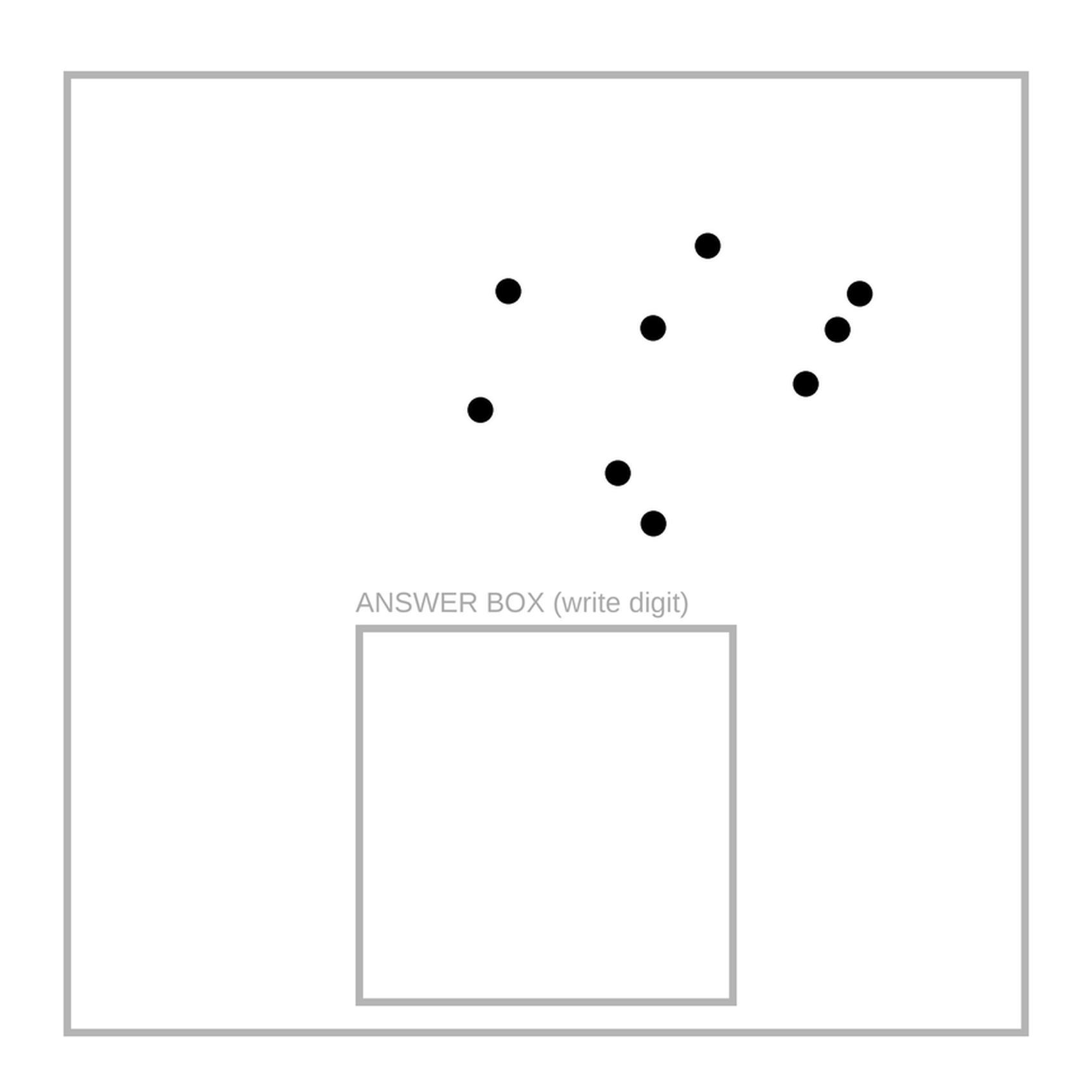} &
    \includegraphics[width=0.32\textwidth,height=0.205\textheight,keepaspectratio]{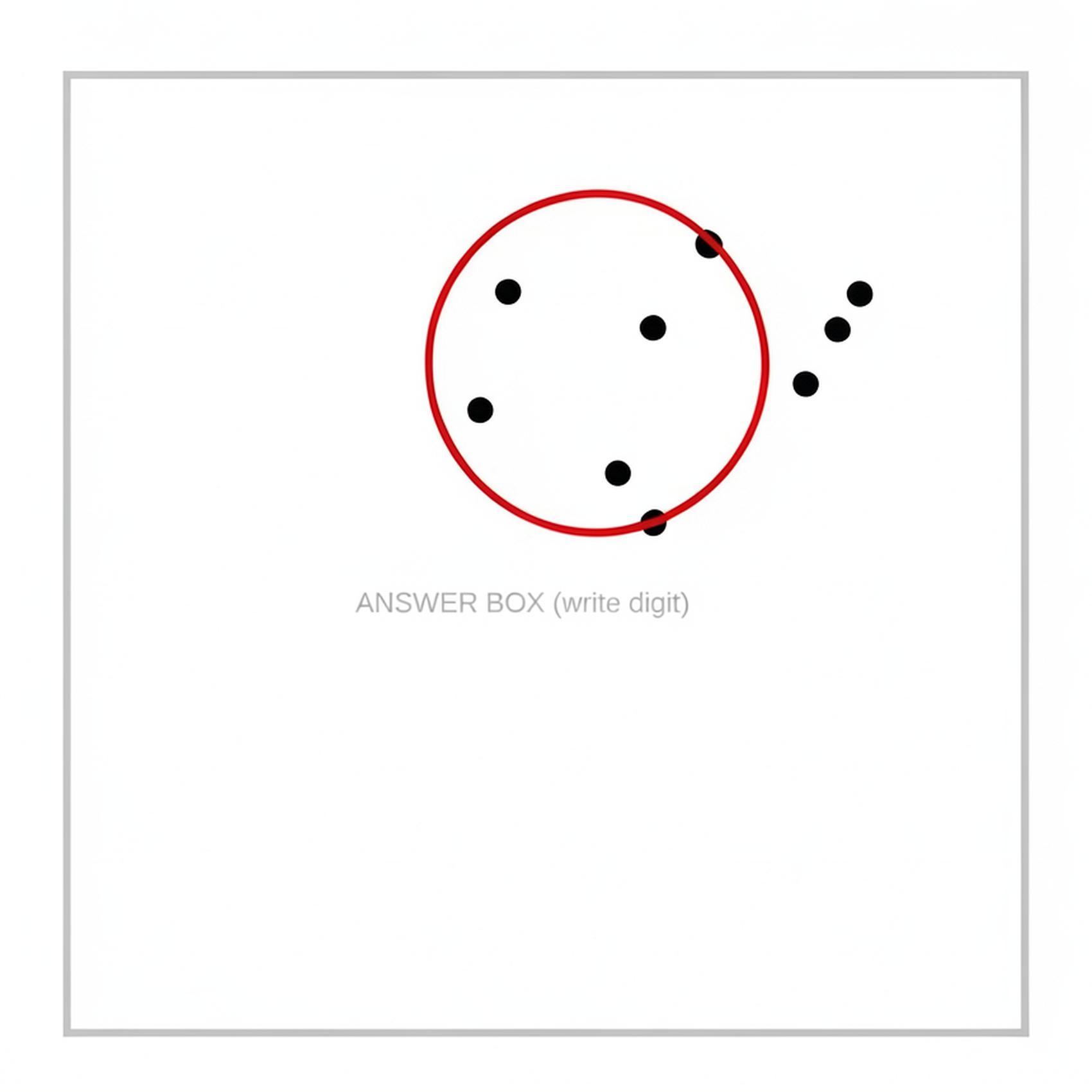} &
    \includegraphics[width=0.32\textwidth,height=0.205\textheight,keepaspectratio]{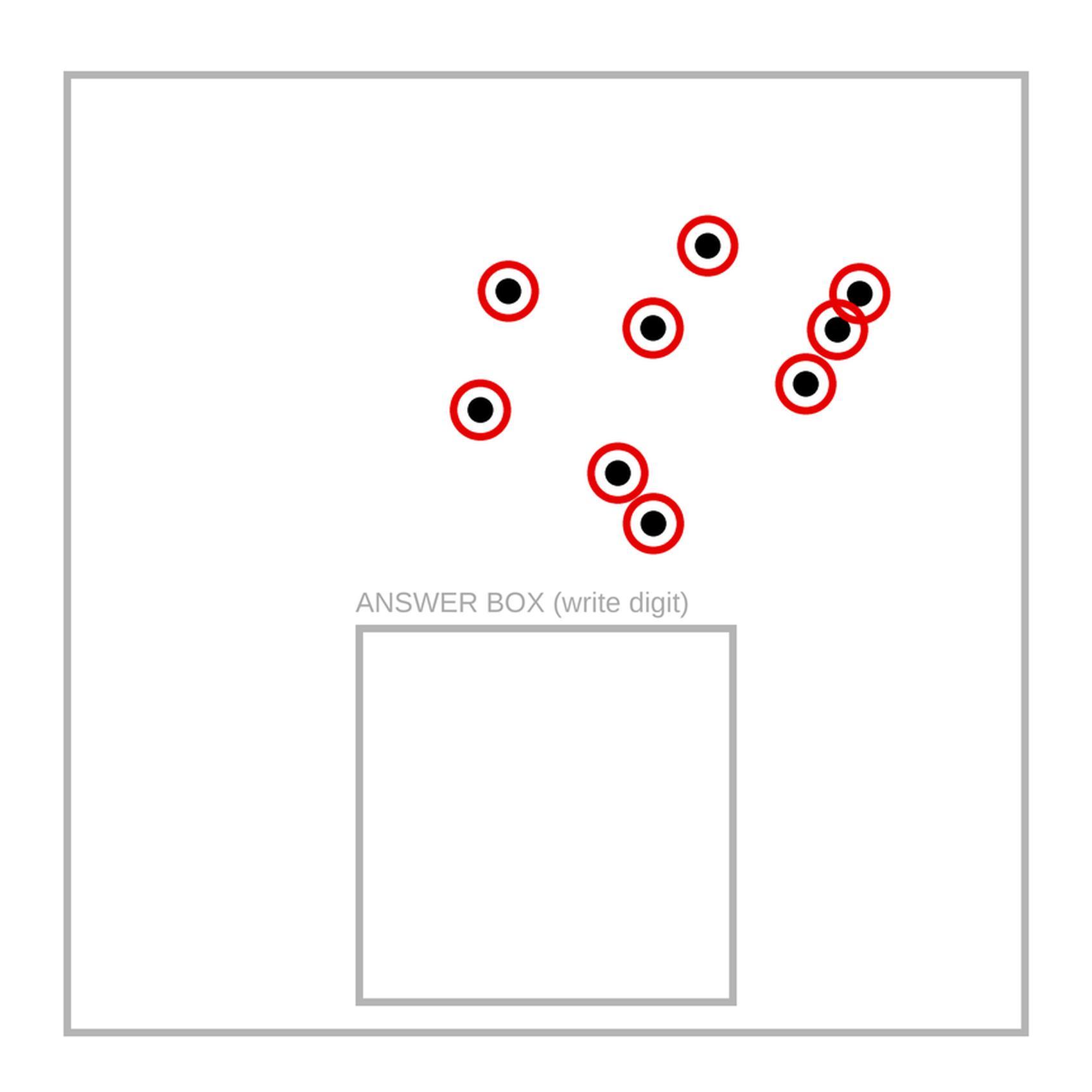}
\end{tabular}
\par\smallskip\raggedright \footnotesize\textbf{(a) Wrong location.} The right kind of mark is placed, but not at the required point or region.
\end{minipage}\\[7pt]
\begin{minipage}{\linewidth}
\centering\fontsize{8}{10}\selectfont
\begin{tabular}{@{}ccc@{}}
\textbf{Input} & \textbf{FLUX.2 Klein 4B API output} & \textbf{Ground truth}\\
\includegraphics[width=0.32\textwidth,height=0.205\textheight,keepaspectratio]{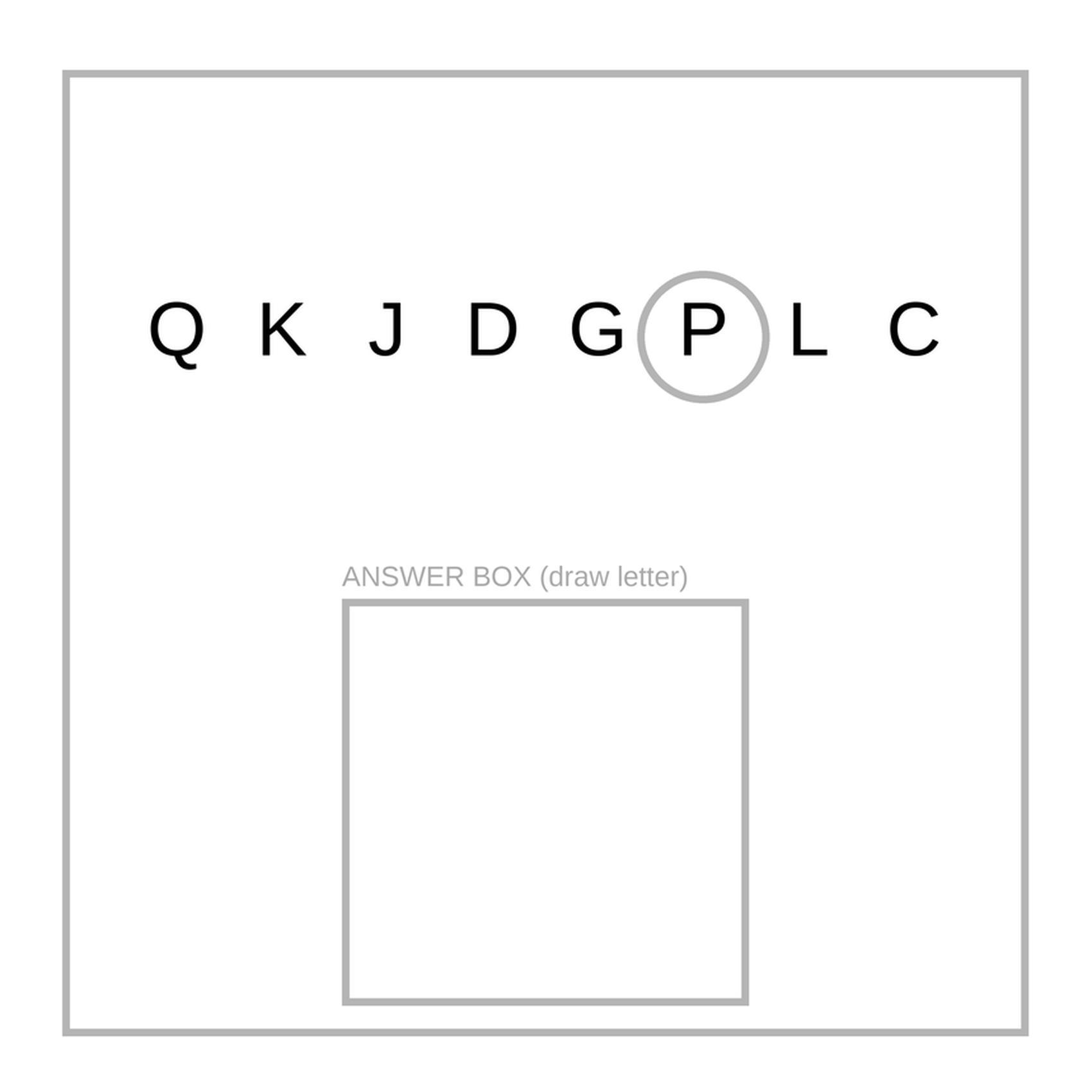} &
    \includegraphics[width=0.32\textwidth,height=0.205\textheight,keepaspectratio]{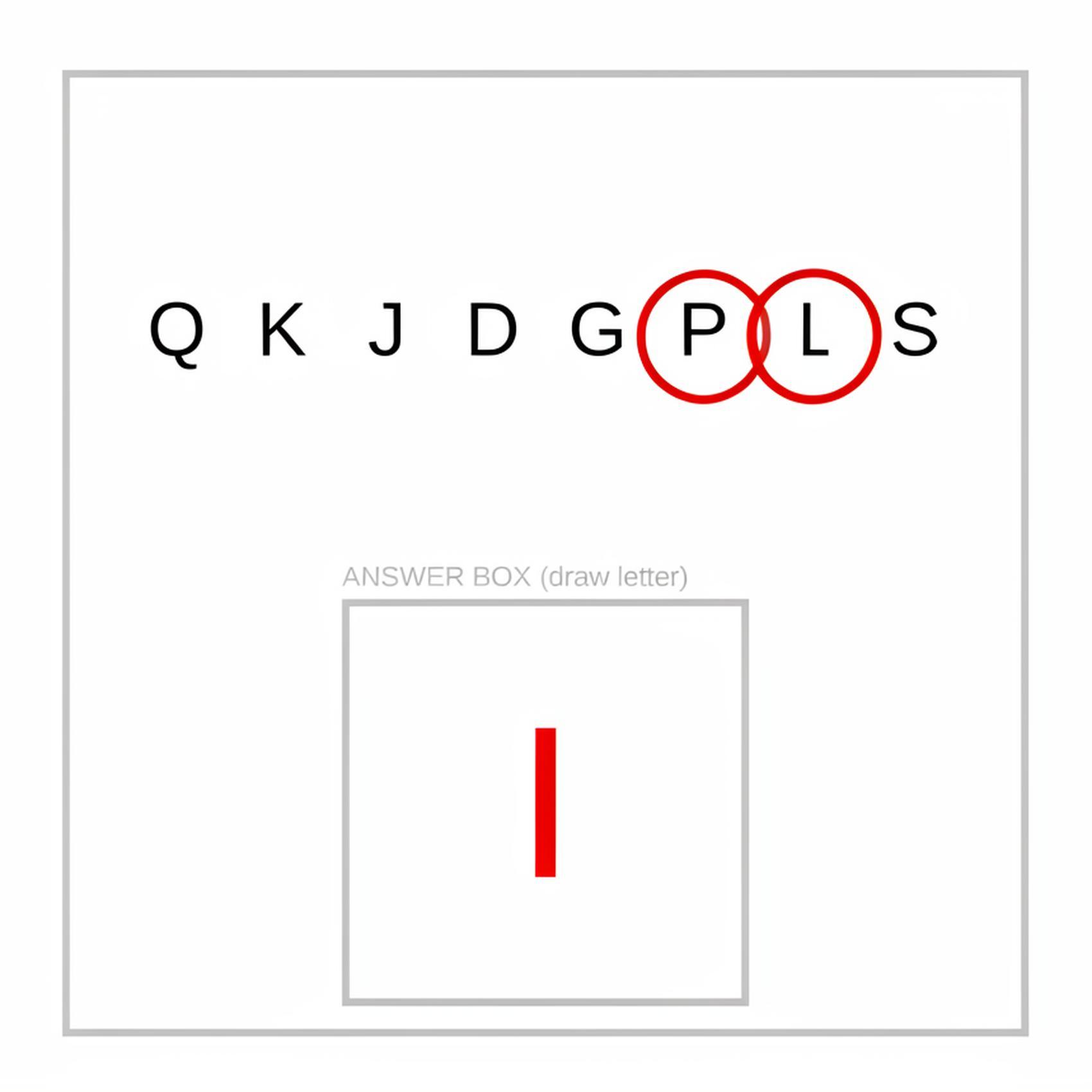} &
    \includegraphics[width=0.32\textwidth,height=0.205\textheight,keepaspectratio]{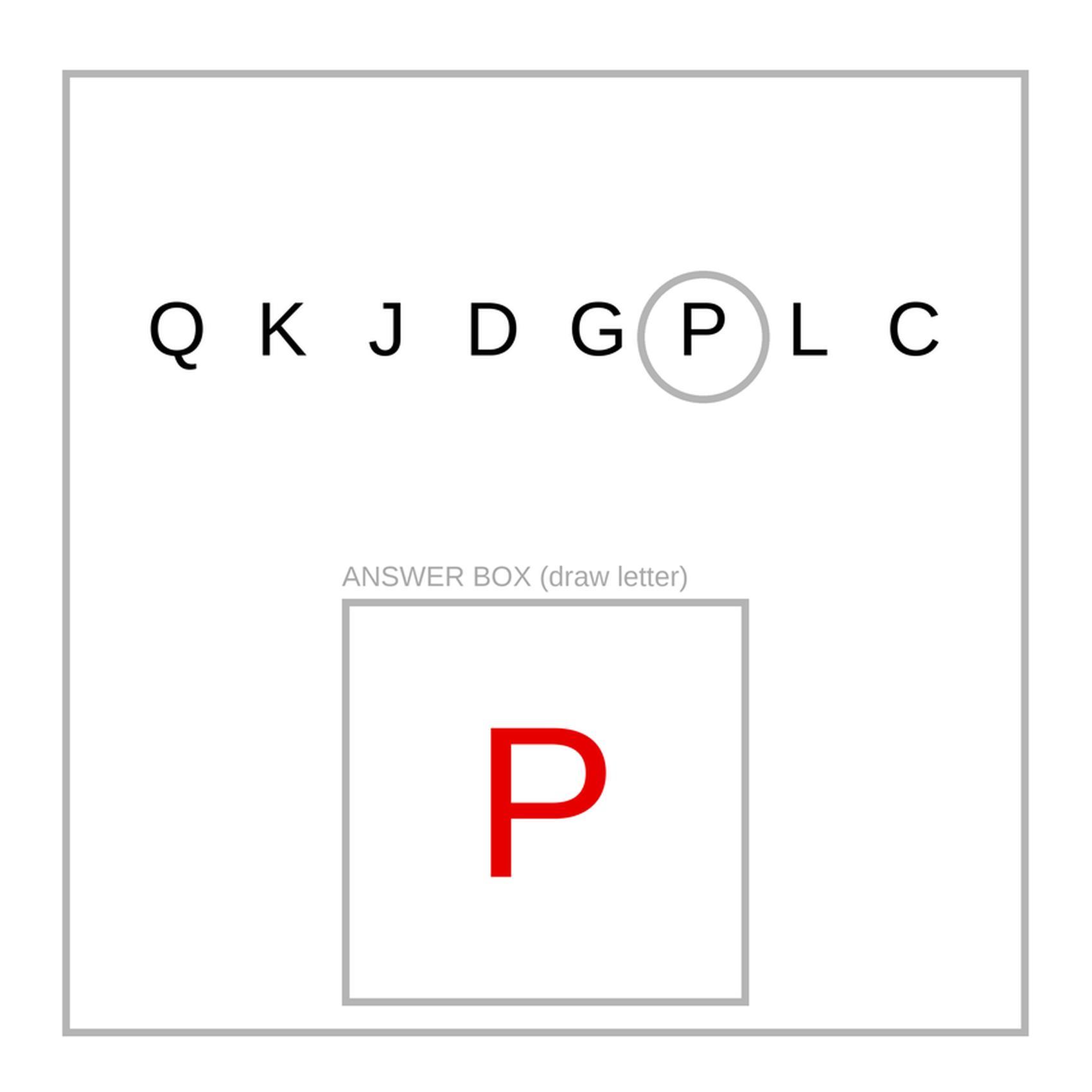}
\end{tabular}
\par\smallskip\raggedright \footnotesize\textbf{(b) Wrong glyph.} The model writes an incorrect alphabetic letter in a letter-copying item.
\end{minipage}\\[7pt]
\begin{minipage}{\linewidth}
\centering\fontsize{8}{10}\selectfont
\begin{tabular}{@{}ccc@{}}
\textbf{Input} & \textbf{FLUX.2 Klein 4B API output} & \textbf{Ground truth}\\
\includegraphics[width=0.32\textwidth,height=0.205\textheight,keepaspectratio]{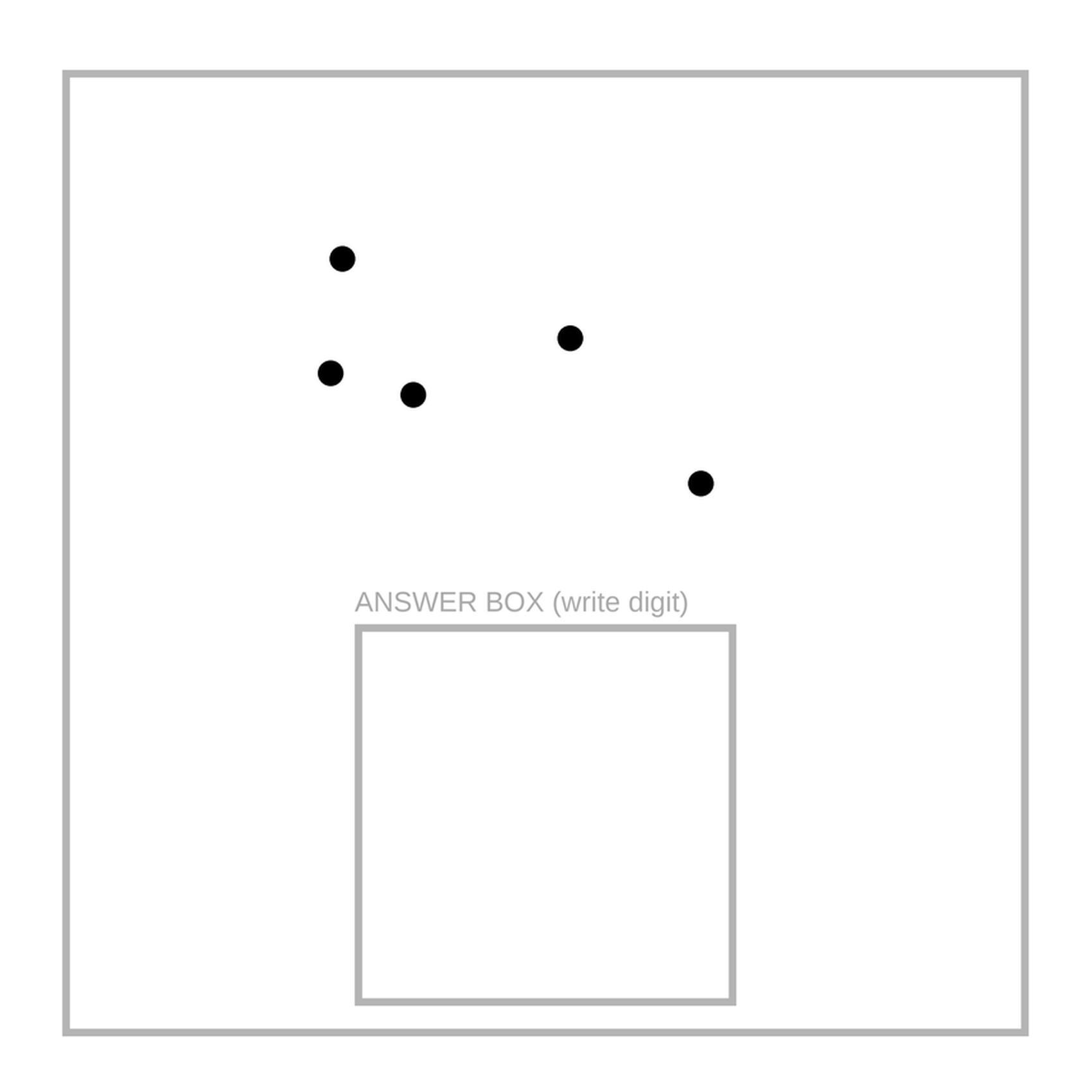} &
    \includegraphics[width=0.32\textwidth,height=0.205\textheight,keepaspectratio]{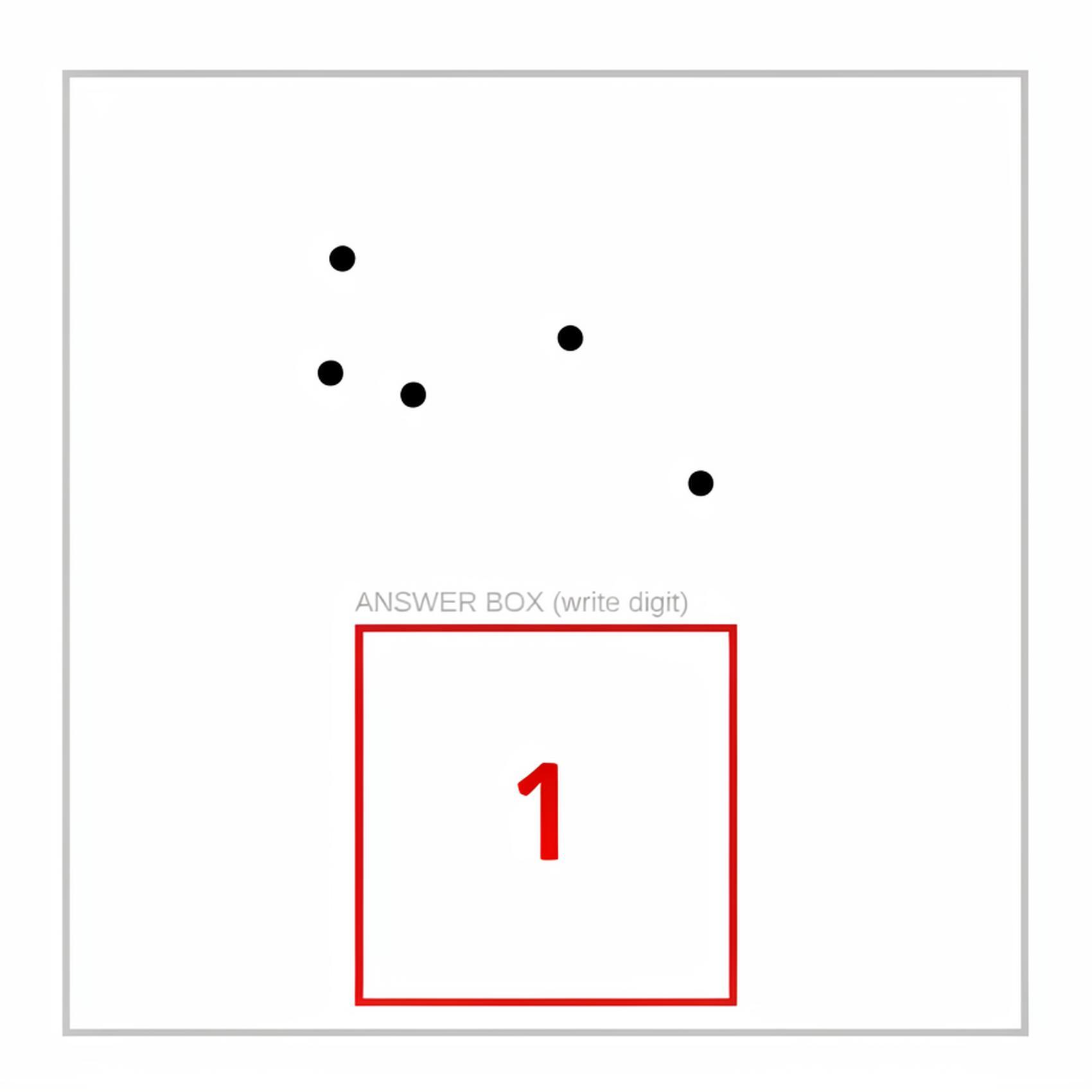} &
    \includegraphics[width=0.32\textwidth,height=0.205\textheight,keepaspectratio]{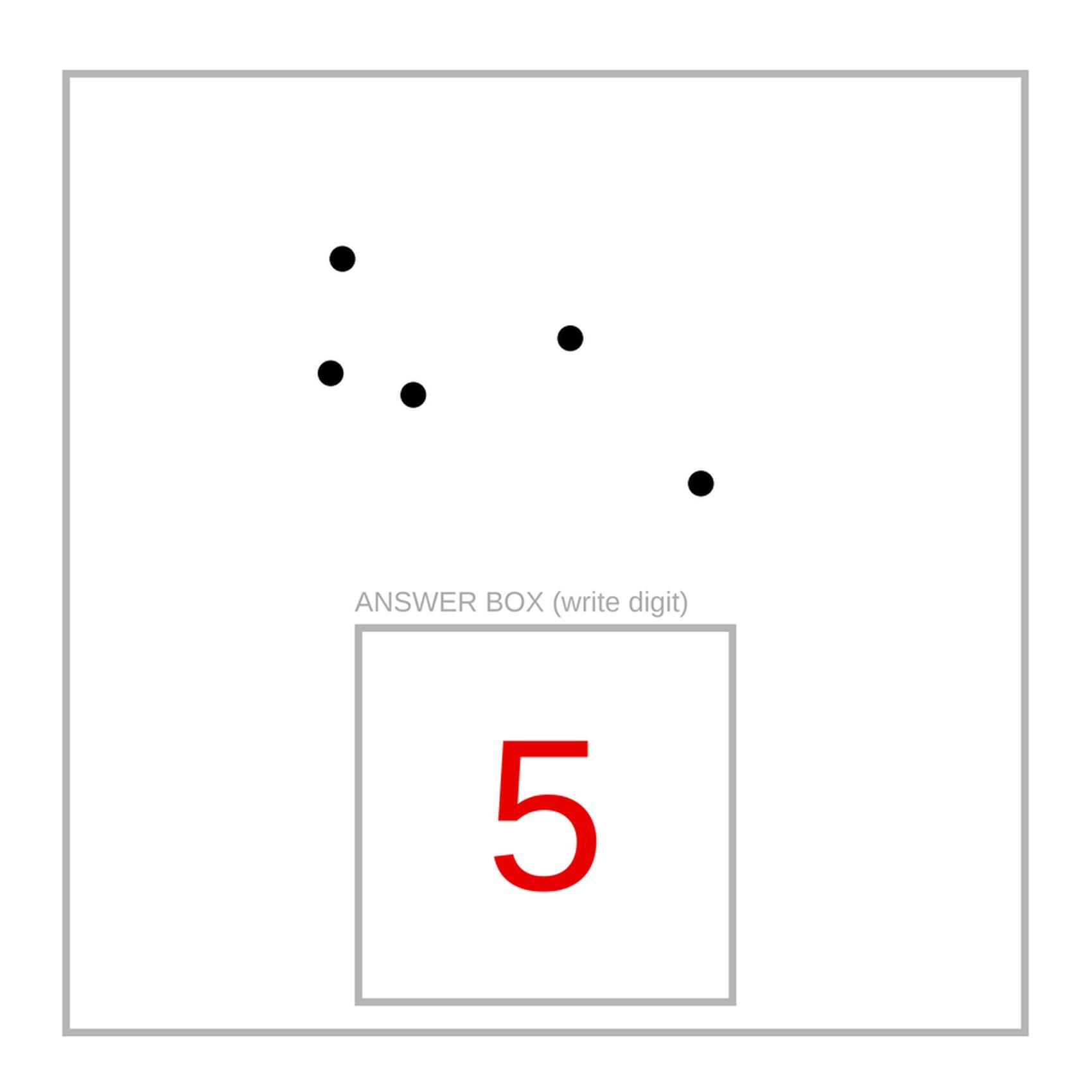}
\end{tabular}
\par\smallskip\raggedright \footnotesize\textbf{(c) Wrong count.} The model writes an incorrect numeric answer in a dot-counting-to-digit item.
\end{minipage}\\[7pt]
\end{longtable}
\endgroup

\section{Three-rater validation}
\label{sec:supp_human_validation}
Three unpaid raters with computer-science backgrounds, including one author, independently judged 440 outputs: 11 tasks $\times$ four conditions (V0--V3) $\times$ two items $\times$ five editors. The editors were Nano Banana Pro, Qwen-Image-Edit, FLUX.2 Klein 4B API, FLUX.2 Klein 4B open-weight, and InstructPix2Pix. The rating materials omitted model identities, condition labels, automatic scores, and other raters' judgments. Raters were instructed to work independently.

Raters used a fixed codebook specifying Strict/Loose decisions, failure categories, precedence for multiple errors, and ambiguous cases. Statistics use the final explicit Human-Strict and Human-Loose labels, rather than a logical AND of component checkboxes. The human reference is a majority of three decisions. Table~\ref{tab:supp_human_task_agreement} compares automatic decisions with this reference. Individual annotations and correspondence remain private; only pooled summaries and four selected examples' majority labels are released.

\par\begingroup
\small
\setlength{\tabcolsep}{4pt}
\captionsetup{hypcap=false,skip=4pt}
\setlength{\LTcapwidth}{\textwidth}
\begin{longtable}{lrrrrrr}
\caption{Automatic--human majority agreement (\%) on 40 outputs per task. Positive means a pass; FP and FN denote automatic false positives and false negatives relative to the human majority. Agreement is not task success.}\label{tab:supp_human_task_agreement}\\
\toprule
& \multicolumn{3}{c}{Strict} & \multicolumn{3}{c}{Loose} \\
\cmidrule(lr){2-4}\cmidrule(lr){5-7}
Task & Agree. & FP & FN & Agree. & FP & FN \\

\midrule
\endfirsthead
\toprule
& \multicolumn{3}{c}{Strict} & \multicolumn{3}{c}{Loose} \\
\cmidrule(lr){2-4}\cmidrule(lr){5-7}
Task & Agree. & FP & FN & Agree. & FP & FN \\

\midrule
\endhead
\bottomrule
\endfoot
Line intersections & 100.0 & 0 & 0 & 95.0 & 2 & 0 \\
Line midpoint & 97.5 & 0 & 1 & 92.5 & 1 & 2 \\
Circle every dot & 97.5 & 0 & 1 & 95.0 & 1 & 1 \\
Overlapping-shape center & 100.0 & 0 & 0 & 100.0 & 0 & 0 \\
Target-shape filling & 100.0 & 0 & 0 & 82.5 & 7 & 0 \\
Circled-letter copying & 92.5 & 0 & 3 & 95.0 & 0 & 2 \\
Identical-letter copying & 95.0 & 0 & 2 & 95.0 & 2 & 0 \\
Single-letter copying & 97.5 & 0 & 1 & 92.5 & 1 & 2 \\
Dot counting to digit & 87.5 & 1 & 4 & 82.5 & 7 & 0 \\
Blank-answer suppression & 87.5 & 4 & 1 & 62.5 & 13 & 2 \\
Empty-circle suppression & 95.0 & 2 & 0 & 90.0 & 2 & 2 \\
\end{longtable}
\endgroup

\par\begingroup
\small
\captionsetup{hypcap=false,skip=4pt}
\setlength{\LTcapwidth}{\textwidth}
\begin{longtable}{lrr}
\caption{Pooled three-rater validation on 440 outputs. Automatic metrics use the human majority; agreement is not task success. Percentages except Fleiss' $\kappa$.}\label{tab:supp_human_overall}\\
\toprule
Metric & Strict & Loose \\

\midrule
\endfirsthead
\toprule
Metric & Strict & Loose \\

\midrule
\endhead
\bottomrule
\endfoot
Unanimous human agreement & 95.5 & 88.9 \\
Fleiss' $\kappa$ & 0.894 & 0.783 \\
Automatic--human agreement & 95.5 & 89.3 \\
Precision & 90.1 & 69.0 \\
Recall & 83.1 & 87.9 \\
F1 & 86.5 & 77.3 \\
\end{longtable}
\endgroup

\section{Rendered-Instruction Diagnostic Probes}
The rendered-instruction diagnostics test whether \nbp{} uses task text embedded in the image, how it resolves conflicting prompt channels, how semantic changes to the rendered instruction affect behavior, and whether red text inside the instruction header can confound red-mask diagnostics. The probes are diagnostic only and are not used for the main model ranking.

\par\begingroup
\fontsize{8}{10}\selectfont
\captionsetup{hypcap=false,skip=4pt}
\setlength{\LTcapwidth}{\textwidth}
\begin{xltabular}{\textwidth}{lrZZ}
\caption{Rendered-instruction diagnostic sizes. Evaluated rows equal tasks multiplied by base items and conditions.}\label{tab:dlp-eval-sizes}\\
\toprule
Probe & Rows & Design calculation & Diagnostic purpose \\

\midrule
\endfirsthead
\toprule
Probe & Rows & Design calculation & Diagnostic purpose \\

\midrule
\endhead
\bottomrule
\endfoot
D-series & 250 & 5 tasks $\times$ 10 base items $\times$ 5 conditions & Instruction access and prompt-priority conflicts. \\
L-series & 176 & 11 tasks $\times$ 4 base items $\times$ 4 perturbation conditions & Semantic sensitivity to paraphrases, keyword swaps, role/order changes, and scrambled word order. \\
P-series & 240 & 4 tasks $\times$ 10 base items $\times$ 6 color conditions & Separation of lexical color phrases from visually red instruction text. \\
\end{xltabular}
\endgroup

\subsection{D-series: instruction access and prompt priority}

The five tasks are filling, line intersections, midpoint marking, circled-letter copying, and dot counting, with ten problems per task and condition. ``Target'' requires red Dice of at least .35 on the task or answer mask and no more than 50 added red pixels in the instruction region. D2 uses a separate color rule: more than 50 blue pixels and no more than 50 red pixels in the worksheet body. Neither rule includes all Auto-Strict requirements.

\par\begingroup
\fontsize{8}{10}\selectfont
\captionsetup{hypcap=false,skip=4pt}
\setlength{\LTcapwidth}{\textwidth}
\begin{xltabular}{\textwidth}{lZrrrr}
\caption{D-series diagnostic counts. ``Target'' is a red-mask overlap classification; ``Text'' flags added red in the instruction region. D2 requests red in the image and blue externally; all 50 outputs are classified as blue-only in the worksheet body. These categories do not establish full edit correctness.}\label{tab:dseries}\\
\toprule
Condition & Definition & Target & Text & External & Other \\

\midrule
\endfirsthead
\toprule
Condition & Definition & Target & Text & External & Other \\

\midrule
\endhead
\bottomrule
\endfoot
D0 & Full in-image instruction; no external text prompt. & 41 & 0 & -- & 9 \\
D1 & Full in-image instruction; external prompt only says to follow the image instruction. & 43 & 2 & -- & 5 \\
D2 & In-image red instruction conflicts with external blue instruction. & 0 & 0 & 50 & 0 \\
D3 & Image header contains irrelevant leading text followed by the real task; external prompt only says to follow image instruction. & 41 & 0 & -- & 9 \\
D4 & Image header only specifies a red-ink cue; external prompt only says to follow image instruction. & 9 & 1 & -- & 40 \\
\end{xltabular}
\endgroup

\par\begingroup
\fontsize{8}{10}\selectfont
\captionsetup{hypcap=false,skip=4pt}
\setlength{\LTcapwidth}{\textwidth}
\begin{xltabular}{\textwidth}{lZZ}
\caption{D-series task-specific rendered header line for D3 and external blue-conflict suffix for D2. The task order is shared across the D-series tables.}\label{tab:dseries-task-texts}\\
\toprule
Task & D3 real instruction line & D2 blue-conflict prompt suffix \\

\midrule
\endfirsthead
\toprule
Task & D3 real instruction line & D2 blue-conflict prompt suffix \\

\midrule
\endhead
\bottomrule
\endfoot
Fill shape & INSTEAD, fill the target outlined shape solid red. & You are given a target outlined shape inside a light gray square frame. Fill the target outlined shape solid blue. Do not change any black lines. Do not add new lines or extra marks. \\
Intersections & INSTEAD, mark all intersections with small red dots. & You are given black line segments inside a light gray square frame. Mark all intersections with small blue dots. Do not change any black lines. Do not add new lines or extra marks. \\
Midpoint & INSTEAD, mark the midpoint of the line with a small red dot. & You are given a black line segment inside a light gray square frame. Mark the midpoint of the line with a small blue dot. Do not change any black lines. Do not add new lines or extra marks. \\
Circled-letter & INSTEAD, draw the circled letter in the answer box below in red. & You are given letters and one of them is circled. Draw the circled letter in the answer box below in blue. Do not change any existing letters or lines. Do not add extra marks. \\
Count-to-digit & INSTEAD, count the dots and write the correct digit in red. & You are given black dots inside a light gray square frame. Count the dots and write the correct digit in the answer box below in blue. Do not change any black lines. Do not add extra marks. \\
\end{xltabular}
\endgroup

\subsection{L-series: semantic variation of rendered instructions}

\par\begingroup
\fontsize{8}{10}\selectfont
\captionsetup{hypcap=false,skip=4pt}
\setlength{\LTcapwidth}{\textwidth}
\begin{xltabular}{\textwidth}{llrZZr}
\caption{L-series evaluation summary. L0 is the canonical instruction listed in Table~\ref{tab:lseries-instructions} for comparison and is not counted as an evaluated perturbation row.}\label{tab:lseries-summary}\\
\toprule
Condition & Perturbation & Rows & Main observed outcome & Secondary observed outcome & Other \\

\midrule
\endfirsthead
\toprule
Condition & Perturbation & Rows & Main observed outcome & Secondary observed outcome & Other \\

\midrule
\endhead
\bottomrule
\endfoot
L1 & Same-meaning paraphrase & 44 & Target/original task followed: 36 & -- & 8 \\
L2 & Keyword substitution & 44 & Changed-not-original: 20 & Original task still followed: 12 & 12 \\
L3 & Role/order change & 44 & Changed-not-original: 34 & Original task still followed: 2 & 8 \\
L4 & Scrambled word order & 44 & Edited despite nonsense: 35 & -- & 9 \\
\end{xltabular}
\endgroup

\begingroup
\fontsize{8}{10}\selectfont
\setlength{\LTleft}{0pt}
\setlength{\LTright}{0pt}
\captionsetup{hypcap=false,skip=4pt}
\Needspace{5\baselineskip}
\setlength{\LTcapwidth}{\textwidth}
\begin{longtable}[l]{L{0.17\textwidth}L{0.16\textwidth}p{0.56\textwidth}}
\caption{Exact rendered instruction texts for the L-series semantic-variation probe. Tasks are ordered as in the core benchmark task table. L0 is the canonical rendered instruction shown for reference; the evaluated rows count uses the four perturbation conditions L1--L4.}\label{tab:lseries-instructions}\\
\toprule
Task & Variant & Rendered instruction text \\
\midrule
\endfirsthead
\toprule
Task & Variant & Rendered instruction text \\
\midrule
\endhead
\bottomrule
\endfoot
Line intersections & L0 canonical & You are given black line segments inside a light gray square frame. Place a small red dot at every intersection point of the black lines. Keep all existing black lines and printed text unchanged. Do not add any marks other than the requested edit. \\
Line intersections & L1 paraphrase & Mark every crossing point where black lines meet with one small red dot. Keep all existing black lines and printed text unchanged. Do not add any marks other than the requested edit. \\
Line intersections & L2 keyword swap & Put one small red dot at the midpoint of each black line segment. Keep all existing black lines and printed text unchanged. Do not add any marks other than the requested edit. \\
Line intersections & L3 role/order & Mark the black line segment that crosses another line, not the crossing point itself. Keep all existing black lines and printed text unchanged. Do not add any marks other than the requested edit. \\
Line intersections & L4 scrambled & red crossing point black lines small dot mark where meet. Keep all existing black lines and printed text unchanged. Do not add any marks other than the requested edit. \\
\addlinespace[1pt]
Line midpoint & L0 canonical & You are given a black line segment inside a light gray square frame. Place a small red dot at the midpoint of the line segment. Keep all existing black lines and printed text unchanged. Do not add any marks other than the requested edit. \\
Line midpoint & L1 paraphrase & Mark the center point of the black line segment with one small red dot. Keep all existing black lines and printed text unchanged. Do not add any marks other than the requested edit. \\
Line midpoint & L2 keyword swap & Put small red dots at the two endpoints of the black line segment. Keep all existing black lines and printed text unchanged. Do not add any marks other than the requested edit. \\
Line midpoint & L3 role/order & Mark the black line segment that contains the midpoint, not the midpoint itself. Keep all existing black lines and printed text unchanged. Do not add any marks other than the requested edit. \\
Line midpoint & L4 scrambled & red midpoint segment black line center point small dot. Keep all existing black lines and printed text unchanged. Do not add any marks other than the requested edit. \\
\addlinespace[1pt]
Circle every dot & L0 canonical & You are given black dots inside a light gray square frame. Draw a thin red circle around each black dot. Do not circle empty space and do not miss any dot. Keep all existing black lines and printed text unchanged. Do not add any marks other than the requested edit. \\
Circle every dot & L1 paraphrase & Draw a red ring around every black dot. Keep all existing black lines and printed text unchanged. Do not add any marks other than the requested edit. \\
Circle every dot & L2 keyword swap & Draw a red ring around only one black dot. Keep all existing black lines and printed text unchanged. Do not add any marks other than the requested edit. \\
Circle every dot & L3 role/order & Circle the square frame that contains the black dots, not each dot. Keep all existing black lines and printed text unchanged. Do not add any marks other than the requested edit. \\
Circle every dot & L4 scrambled & red ring every black dot around draw each circle. Keep all existing black lines and printed text unchanged. Do not add any marks other than the requested edit. \\
\addlinespace[1pt]
Overlapping shapes & L0 canonical & You are given multiple overlapping outlined shapes inside a light gray square frame. Place one small red dot at the center of each outlined shape. Keep all existing black lines and printed text unchanged. Do not add any marks other than the requested edit. \\
Overlapping shapes & L1 paraphrase & Place one small red dot at the center of each outlined shape. Keep all existing black lines and printed text unchanged. Do not add any marks other than the requested edit. \\
Overlapping shapes & L2 keyword swap & Place one small red circle at the center of each outlined shape. Keep all existing black lines and printed text unchanged. Do not add any marks other than the requested edit. \\
Overlapping shapes & L3 role/order & Place one red dot at the center of the whole group of shapes, not at each individual shape center. Keep all existing black lines and printed text unchanged. Do not add any marks other than the requested edit. \\
Overlapping shapes & L4 scrambled & red dot center each outlined shape small place. Keep all existing black lines and printed text unchanged. Do not add any marks other than the requested edit. \\
\addlinespace[1pt]
Fill shape & L0 canonical & You are given an outlined target shape inside a light gray square frame. Fill the outlined target shape solid red. Keep all existing black lines and printed text unchanged. Do not add any marks other than the requested edit. \\
Fill shape & L1 paraphrase & Color only the outlined target shape solid red. Keep all existing black lines and printed text unchanged. Do not add any marks other than the requested edit. \\
Fill shape & L2 keyword swap & Outline the target shape in red instead of filling it. Keep all existing black lines and printed text unchanged. Do not add any marks other than the requested edit. \\
Fill shape & L3 role/order & Color the region outside the outlined target shape red, not the shape itself. Keep all existing black lines and printed text unchanged. Do not add any marks other than the requested edit. \\
Fill shape & L4 scrambled & red target outlined shape solid fill only color. Keep all existing black lines and printed text unchanged. Do not add any marks other than the requested edit. \\
\addlinespace[1pt]
Circled-letter & L0 canonical & You are given a row of letters and exactly one letter is circled. Copy the circled letter into the answer box below using red ink. Keep all existing black lines and printed text unchanged. Do not add any marks other than the requested edit. \\
Circled-letter & L1 paraphrase & Copy the letter inside the circle into the answer box using red ink. Keep all existing black lines and printed text unchanged. Do not add any marks other than the requested edit. \\
Circled-letter & L2 keyword swap & Copy the uncircled letter immediately to the left of the circle into the answer box. Keep all existing black lines and printed text unchanged. Do not add any marks other than the requested edit. \\
Circled-letter & L3 role/order & Draw a red circle around the letter inside the answer box, rather than writing the circled row letter. Keep all existing black lines and printed text unchanged. Do not add any marks other than the requested edit. \\
Circled-letter & L4 scrambled & red answer box circled letter copy inside circle. Keep all existing black lines and printed text unchanged. Do not add any marks other than the requested edit. \\
\addlinespace[1pt]
All-A & L0 canonical & You are given a row of letters and exactly one letter is circled. Copy the circled letter into the answer box below using red ink. Keep all existing black lines and printed text unchanged. Do not add any marks other than the requested edit. \\
All-A & L1 paraphrase & Copy the letter inside the circle into the answer box using red ink. Keep all existing black lines and printed text unchanged. Do not add any marks other than the requested edit. \\
All-A & L2 keyword swap & Write the letter B in the answer box in red. Keep all existing black lines and printed text unchanged. Do not add any marks other than the requested edit. \\
All-A & L3 role/order & Draw a red circle around the answer box, rather than writing the circled A. Keep all existing black lines and printed text unchanged. Do not add any marks other than the requested edit. \\
All-A & L4 scrambled & red A circled answer box copy letter inside. Keep all existing black lines and printed text unchanged. Do not add any marks other than the requested edit. \\
\addlinespace[1pt]
Single letter & L0 canonical & You are given a single circled letter. Copy the circled letter into the answer box below using red ink. Keep all existing black lines and printed text unchanged. Do not add any marks other than the requested edit. \\
Single letter & L1 paraphrase & Write the single displayed letter in the answer box using red ink. Keep all existing black lines and printed text unchanged. Do not add any marks other than the requested edit. \\
Single letter & L2 keyword swap & Write the letter A in the answer box instead of the shown letter. Keep all existing black lines and printed text unchanged. Do not add any marks other than the requested edit. \\
Single letter & L3 role/order & Draw a red box around the shown letter, rather than copying it into the answer box. Keep all existing black lines and printed text unchanged. Do not add any marks other than the requested edit. \\
Single letter & L4 scrambled & red shown letter copy answer box write single. Keep all existing black lines and printed text unchanged. Do not add any marks other than the requested edit. \\
\addlinespace[1pt]
Count-to-digit & L0 canonical & You are given black dots inside a light gray square frame. Count the dots and write the correct digit from 1 to 9 in the answer box below using red ink. Keep all existing black lines and printed text unchanged. Do not add any marks other than the requested edit. \\
Count-to-digit & L1 paraphrase & Write in red the number of black dots inside the answer box. Keep all existing black lines and printed text unchanged. Do not add any marks other than the requested edit. \\
Count-to-digit & L2 keyword swap & Draw a red ring around every black dot instead of writing a number. Keep all existing black lines and printed text unchanged. Do not add any marks other than the requested edit. \\
Count-to-digit & L3 role/order & Write the number beside the dot group, not inside the answer box. Keep all existing black lines and printed text unchanged. Do not add any marks other than the requested edit. \\
Count-to-digit & L4 scrambled & red digit black dots count answer box number write. Keep all existing black lines and printed text unchanged. Do not add any marks other than the requested edit. \\
\addlinespace[1pt]
Blank response & L0 canonical & You are given a row of letters with a circle shown in the row. Leave the answer box completely blank and do not write any letter. Keep all existing black lines and printed text unchanged. Do not add any marks other than the requested edit. \\
Blank response & L1 paraphrase & Leave the answer box empty even though a circle is present; do not write any letter. Keep all existing black lines and printed text unchanged. Do not add any marks other than the requested edit. \\
Blank response & L2 keyword swap & Write the letter X in the answer box instead of leaving it blank. Keep all existing black lines and printed text unchanged. Do not add any marks other than the requested edit. \\
Blank response & L3 role/order & Circle the empty answer box with red ink, rather than leaving it blank. Keep all existing black lines and printed text unchanged. Do not add any marks other than the requested edit. \\
Blank response & L4 scrambled & blank circle present answer box leave empty no write. Keep all existing black lines and printed text unchanged. Do not add any marks other than the requested edit. \\
\addlinespace[1pt]
Empty circle & L0 canonical & You are given a row with an empty circle that contains no letter. Leave the answer box completely blank and do not write any letter. Keep all existing black lines and printed text unchanged. Do not add any marks other than the requested edit. \\
Empty circle & L1 paraphrase & Leave the answer box empty because the shown circle contains no letter. Keep all existing black lines and printed text unchanged. Do not add any marks other than the requested edit. \\
Empty circle & L2 keyword swap & Write the letter X inside the empty circle in red. Keep all existing black lines and printed text unchanged. Do not add any marks other than the requested edit. \\
Empty circle & L3 role/order & Circle the blank answer box, rather than treating the empty circle as no answer. Keep all existing black lines and printed text unchanged. Do not add any marks other than the requested edit. \\
Empty circle & L4 scrambled & empty circle no letter answer box leave blank. Keep all existing black lines and printed text unchanged. Do not add any marks other than the requested edit. \\
\addlinespace[1pt]
\end{longtable}
\endgroup

\subsection{P-series: color-channel probe}

\par\begingroup
\fontsize{8}{10}\selectfont
\captionsetup{hypcap=false,skip=4pt}
\setlength{\LTcapwidth}{\textwidth}
\begin{longtable}{lll}
\caption{P-series lexical/display-color design. The 240 evaluated outputs are 4 tasks $\times$ 10 base items $\times$ 6 conditions.}\label{tab:pseries-design}\\
\toprule
Probe ID & Lexical phrase & Display color of phrase \\

\midrule
\endfirsthead
\toprule
Probe ID & Lexical phrase & Display color of phrase \\

\midrule
\endhead
\bottomrule
\endfoot
P0 & red ink & gray \\
P1 & red ink & red \\
P2 & red ink & blue \\
P3 & the specified color & gray \\
P4 & the specified color & red \\
P5 & the specified color & blue \\
\end{longtable}
\endgroup

\par\begingroup
\fontsize{8}{10}\selectfont
\captionsetup{hypcap=false,skip=4pt}
\setlength{\LTcapwidth}{\textwidth}
\begin{xltabular}{\textwidth}{lZ}
\caption{P-series strong external prompts. The external prompt always requests the red answer channel; the rendered image header independently manipulates the lexical phrase and display color in Table~\ref{tab:pseries-design}.}\label{tab:pseries-external-prompts}\\
\toprule
Task & External prompt \\

\midrule
\endfirsthead
\toprule
Task & External prompt \\

\midrule
\endhead
\bottomrule
\endfoot
Fill shape & Fill the outlined target shape solid RED. Keep all existing black lines and any printed text unchanged. Do not add any new lines, shapes, or marks other than the required edit. \\
Midpoint & Place one small RED dot at the midpoint of the black line segment. Keep all existing black lines and any printed text unchanged. Do not add any new lines, shapes, or marks other than the required edit. \\
Circled-letter & Copy the circled letter into the answer box in RED. Keep all existing black lines and any printed text unchanged. Do not add any new lines, shapes, or marks other than the required edit. \\
Count-to-digit & Count the black dots and write the correct digit from 1 to 9 in the answer box in RED. Keep all existing black lines and any printed text unchanged. Do not add any new lines, shapes, or marks other than the required edit. \\
\end{xltabular}
\endgroup

\par\begingroup
\fontsize{8}{10}\selectfont
\captionsetup{hypcap=false,skip=4pt}
\setlength{\LTcapwidth}{\textwidth}
\begin{longtable}{lrrrr}
\caption{P-series automatic summary aggregated over four tasks. Target proxy is the mean of task-level target-proxy rates; the final column is the archived instruction-region red-signal flag, not evidence of newly generated leakage.}\label{tab:pseries-aggregate}\\
\toprule
Probe ID & Outputs & Mean target proxy & Min task target proxy & Mean red-signal flag \\

\midrule
\endfirsthead
\toprule
Probe ID & Outputs & Mean target proxy & Min task target proxy & Mean red-signal flag \\

\midrule
\endhead
\bottomrule
\endfoot
P0 & 40 & 0.975 & 0.900 & 0.000 \\
P1 & 40 & 0.975 & 0.900 & 1.000 \\
P2 & 40 & 0.975 & 0.900 & 0.000 \\
P3 & 40 & 0.975 & 0.900 & 0.000 \\
P4 & 40 & 0.975 & 0.900 & 1.000 \\
P5 & 40 & 1.000 & 1.000 & 0.000 \\
\end{longtable}
\endgroup

The color-probe metadata stores the lexical phrase and display color for every item, plus the rendered input image. Red instruction-region detections in P1 and P4 correspond to red pixels already present in the rendered input header, not to newly introduced answer edits.

\section{Reasoning-Stress Diagnostics}
The reasoning-stress suite contains 398 distinct outputs: 378 in Table~\ref{tab:reason-summary} and 20 Sudoku contradiction variants. It is not used for the main model ranking. All probes keep the answer convention in image space: the model must place the final answer as a red edit while preserving the worksheet.

\subsection{Summary table}
\par\begingroup

\captionsetup{hypcap=false,skip=4pt}
\setlength{\LTcapwidth}{\textwidth}
\begin{longtable}{lrrrr}
\caption{Reasoning-stress diagnostic scores for \nbp{}. Wason requires exact selection of the correct card set; other rows use archived image-based proxies. Score+Fmt. also requires native format.}\label{tab:reason-summary}\\
\toprule
Probe & $n$ & Task score (\%) & Format (\%) & Score+Fmt. (\%) \\

\midrule
\endfirsthead
\toprule
Probe & $n$ & Task score (\%) & Format (\%) & Score+Fmt. (\%) \\

\midrule
\endhead
\bottomrule
\endfoot
Neighbor, 1--10 operations & 100 & 94.0 & 100.0 & 94.0 \\
Neighbor, 10--300 operations & 98 & 91.8 & 99.0 & 91.8 \\
Public RAVEN & 70 & 22.9 & 100.0 & 22.9 \\
Wason selection & 100 & 49.0 & 97.0 & 49.0 \\
Sudoku $4\times4$ & 10 & 70.0 & 100.0 & 70.0 \\
\end{longtable}
\endgroup

\subsection{Neighbor-symbol operation chains}
The neighbor-symbol task renders a horizontal symbol sequence and a start marker. The instruction specifies a sequence of left/right moves, expressed in full words rather than ``L'' or ``R'' abbreviations. The required answer is the final symbol after sequentially applying all moves. Two schedules are used:
\begin{itemize}[leftmargin=1.2em,itemsep=1pt]
    \item a fine-grained short-chain schedule with 1, 2, \ldots, 10 operations and 10 items per level;
    \item a long-chain schedule with 10, 20, 30, 40, 50, 75, 100, 150, 200, and 300 operations and ten items per level except 200 operations, which has eight after removing two duplicate entries for a reused replacement output.
\end{itemize}
The aggregate proxy pass rate is 94.0\% for the 1--10 schedule and 90/98 (91.8\%) for the 10--300 schedule. The proxy rate is 70.0\% at 75 and 100 operations, but 100.0\% at 200 and 300. This small pilot does not show a monotonic decline with sequence length.

\subsection{Public RAVEN}
The RAVEN diagnostic samples 70 public matrix items, 10 from each of seven configurations: center-single, distribute-four, distribute-nine, left/right center-single, up/down center-single, in/out center-single, and in-distribute-four/out-center-single. The answer is an image-space mark selecting the correct option among eight choices. The image-overlap proxy passes 16/70 outputs (22.9\%). It does not enforce a unique option selection, so this is not exact-choice accuracy. The archived rule accepts red Dice of at least .25, or at least 12 overlapping pixels covering at least .25 of the ground-truth answer area after a 20-pixel dilation of the answer mask.

\subsection{Wason selection}
The Wason probe uses five surface variants of the same conditional-selection structure. The logically correct selection is always the $P$ and $\lnot Q$ cards. A permissive red-mask proxy scores 99.0\% of outputs as marking relevant card regions, but exact card-set success is 49.0\%, with 97.0\% native format validity. Table~\ref{tab:wason-variants} shows the surface dependence of exact card-set success.

\par\begingroup
\fontsize{8}{10}\selectfont
\captionsetup{hypcap=false,skip=4pt}
\setlength{\LTcapwidth}{\textwidth}
\begin{longtable}{lrrr}
\caption{Wason selection results by surface variant. Cardwise success requires selecting both and only the falsifying $P$ and $\lnot Q$ cards.}\label{tab:wason-variants}\\
\toprule
Surface variant & $n$ & Red-mask proxy (\%) & Exact card-set success (\%) \\

\midrule
\endfirsthead
\toprule
Surface variant & $n$ & Red-mask proxy (\%) & Exact card-set success (\%) \\

\midrule
\endhead
\bottomrule
\endfoot
Vowel/even & 28 & 100.0 & 67.9 \\
Blue/triangle & 18 & 94.4 & 38.9 \\
Cat/milk & 18 & 100.0 & 38.9 \\
Early-letter/odd & 18 & 100.0 & 55.6 \\
Dax/star & 18 & 100.0 & 33.3 \\
\end{longtable}
\endgroup

\subsection{Sudoku and contradiction variants}
The Sudoku diagnostic uses $4\times4$ puzzles with one shaded cell. The model must write the valid digit for the shaded cell in red. The scored Sudoku set contains 10 items and reaches a 70.0\% proxy pass rate.

The contradiction variant tests no-solution handling. When the rendered instruction explicitly states that the model should write ``NONE'' if no digit satisfies the shaded cell, \nbp{} writes ``NONE'' in all 10 outputs; however, two outputs render it in black rather than in the required red answer channel, giving 80.0\% success on the answer-plus-red-color requirement. This is not a native-canvas format rate. When the same type of unsatisfiable puzzle is shown without mentioning a no-solution response, the model never writes ``NONE''; it writes digits in 9 of 10 outputs and leaves the answer box blank in 1 of 10 outputs.

\raggedbottom
\section{Reproducibility and Data Access}
The public repository provides the evaluator, development renderer, configurations, and release instructions: \url{https://github.com/misora-sugiyama/WISP}. The frozen shared data contain 2,200 records with input images and ground truths; six frozen output sets contain 13,200 outputs in total. The evaluator derives scoring masks from the images. Development examples produced by rerunning the renderer are separate from this fixed dataset.

Versioned release assets provide the shared data and outputs, additional-experiment summaries, archived scores and aggregates for the large FLUX sweep, and pooled human-evaluation summaries. The additional-experiment package does not include its raw inputs, outputs, reference images, or inference notebooks; the FLUX sweep package does not include all 35,200 candidate images. Release \texttt{data-v1.0.2} contains the additional experiment summaries and protocols. Its V4--V7, CREF, and O1/O2 pass rates were produced by the archived evaluator. We exclude those rates here because of target-matching inconsistencies; they are not results from the final shared scorer.

Release \texttt{data-v1.0.1} contains the corrected human-evaluation summaries. The human package includes the study protocols, rating codebooks, pooled tables, and four selected majority-labeled examples. It excludes participant-level solve-only data, individual rater annotations, identifying information, and correspondence. Dataset and code license terms are provided with their respective release materials.

Use the repository's documented download and verification procedure to obtain the frozen data and verify checksums. The public scoring entry point remains \path{evaluator/score_wisrd.py}; this legacy filename does not change the benchmark name WISP. Ground-truth-as-candidate checks are execution and schema checks, not model evaluations.

\section{Known Limitations of the Supplementary Diagnostics}
The benchmark is synthetic, and the reasoning-stress probes are small diagnostic suites. Symbol and digit scoring uses red-mask proxies without recognizing glyph identity. The earlier FLUX failure audit uses one rater; the separate three-rater study validates automatic decisions on 440 outputs from five editors. Human solve-only responses use clicks or typing and are not an image-editing baseline. CREF lacks manual source-following labels. Public RAVEN, Wason, and Sudoku may involve familiarity or template effects, and the probes do not establish open-ended rule induction or transfer to unfamiliar rules. Task-wise validation, held-out rule families, unfamiliar symbols, and counterfactual layouts remain needed.

\section{Agreement and disagreement examples}
\label{sec:supp_proxy_examples}
The examples illustrate agreement and disagreement between automatic proxies and the human majority. They are selected illustrations, not a prevalence estimate. The same pass/fail relationships hold for Strict and Loose in all four examples.
\begin{figure}[H]
\centering
\exampleimagestall{a}
\caption{\textbf{Agreement on a pass (true positive).} Nano Banana Pro, Target-Shape Filling, V0. The task requires filling the target shape in red while preserving its outline and the worksheet. Both the automatic evaluator and the human majority accept this output under Strict and Loose.}
\label{fig:supp_proxy_tp}
\end{figure}

\begin{figure}[H]
\centering
\exampleimages{b}
\caption{\textbf{Agreement on a failure (true negative).} Nano Banana Pro, Line-Intersection Marking, V3. The expected edit is a red dot at the intersection. The model redraws the scene and omits the requested mark. Both the automatic evaluator and the human majority reject this output under Strict and Loose.}
\label{fig:supp_proxy_tn}
\end{figure}

\begin{figure}[H]
\centering
\exampleimages{c}
\caption{\textbf{Automatic false positive relative to the human majority.} FLUX.2 Klein 4B API, Dot Counting to Digit, V0. There are five dots, but the output does not show the required digit 5. Both automatic proxies pass the output, whereas the human majority rejects it. The proxies assess red-mask properties and preservation; passing them does not guarantee the correct digit identity.}
\label{fig:supp_proxy_fp}
\end{figure}

\begin{figure}[H]
\centering
\exampleimages{d}
\caption{\textbf{Automatic false negative relative to the human majority.} Nano Banana Pro, Circled-Letter Copying, V0. The majority accepts the copied letter I under Strict and Loose, but the proxies reject it. The candidate has 5,030 red pixels outside the padded ground-truth glyph region, exceeding the Strict and Loose limits of 250 and 1,000. This region is defined around the ground-truth glyph, not by the answer-box boundary: the candidate letter remains inside the answer box. The example illustrates sensitivity to glyph shape and size.}
\label{fig:supp_proxy_fn}
\end{figure}